\pdfoutput=1
\documentclass[3p]{./elsarticle}
\biboptions{sort&compress, square}
\journal{Robotics and Autonomous Systems}

\def\mydoctitle{Stochastic Triangular Mesh Mapping:\\ {\smaller A Terrain Mapping Technique for Autonomous Mobile Robots}}
\def\mydocauthor{Clint D. Lombard, Corn\'e E. van Daalen}

\def\mydockeywords{robotics, dense mapping, submapping, perception, triangular mesh, probabilistic graphical model, stereo cameras, LiDAR}

\pdfoutput=1
\usepackage{ifpdf}

\usepackage{mathtools}
\usepackage{commath}
\usepackage{amssymb}
\usepackage{amsmath}
\usepackage{stmaryrd}
\usepackage{stackrel}
\usepackage{scalerel}
\usepackage{bm} 

\usepackage{algorithm}
\usepackage{algpseudocode}
\algnewcommand{\LineComment}[2]{\hspace*{#1em}$\triangleright$ \textit{#2}}
\usepackage{relsize}

\newcommand{\mvec}[1]{\bm{#1}}
\newcommand{\mmat}[1]{\uppercase{\bm{#1}}}
\newcommand{\mset}[1]{\uppercase{\mathit{#1}}}
\newcommand{\given}[1]{\mathtt{#1}}
\newcommand{\bigO}[1]{\mathcal{O}(#1)}
\newcommand{\bsmat}[1]{\begin{bsmallmatrix} #1 \end{bsmallmatrix}}
\newcommand{\bmat}[1]{\begin{bmatrix} #1 \end{bmatrix}}
\newcommand{\T}[0]{\intercal}
\newcommand{\approxhat}[1]{\expandafter\tilde#1}
\newcommand{\belief}[0]{{\mathcal{B}}}
\newcommand{\approxBel}[0]{\approxhat{\belief}}
\newcommand{\GM}[0]{\mathcal{N}_m} 
\newcommand{\GC}[0]{\mathcal{N}_c} 
\newcommand{\KL}[2]{D_\text{KL}\left(#1 \,\middle\|\, #2 \right)}

\newcommand{\inarrow}[2]{_{#1 \shortleftarrow #2}}
\newcommand{\outarrow}[2]{_{#1 \shortrightarrow #2}}
\newcommand{\inmsg}[2]{ \delta{\inarrow{#1}{#2}} }
\newcommand{\outmsg}[2]{ \delta{\outarrow{#1}{#2}} }
\newcommand{\approxinmsg}[2]{\approxhat{\inmsg{#1}{#2}}}
\newcommand{\approxoutmsg}[2]{\approxhat{\outmsg{#1}{#2}}}

\usepackage{suffix}
\WithSuffix\newcommand\approxinmsg*[2]{\approxhat{ \delta{^*\inarrow{#1}{#2}} }}
\WithSuffix\newcommand\approxoutmsg*[2]{\approxhat{ \delta{^*\outarrow{#1}{#2}} }}

\newcommand{\cp}[0]{\phi}
\newcommand{\approxcp}[0]{\approxhat{\phi}}

\newcommand{\sepset}[2]{\mathcal{S}_{#1,#2}}

\makeatletter
\renewcommand*\env@matrix[1][*\c@MaxMatrixCols c]{%
  \hskip -\arraycolsep
  \let\@ifnextchar\new@ifnextchar
  \array{#1}}
\makeatother

\newcommand{\appropto}{\mathrel{\vcenter{
      \offinterlineskip\halign{\hfil$##$\cr
        \propto\cr\noalign{\kern2pt}\sim\cr\noalign{\kern-2pt}}}}}

\DeclareMathOperator*{\argmin}{arg\,min}
\DeclareMathOperator*{\diag}{diag}
\newdefinition{caveat}{Caveat}
\newdefinition{assume}{Assumption}

\makeatletter
\newsavebox\myboxA
\newsavebox\myboxB
\newlength\mylenA

\newcommand*\xoverline[2][0.75]{%
    \sbox{\myboxA}{$\m@th#2$}%
    \setbox\myboxB\null
    \ht\myboxB=\ht\myboxA%
    \dp\myboxB=\dp\myboxA%
    \wd\myboxB=#1\wd\myboxA
    \sbox\myboxB{$\m@th\overline{\copy\myboxB}$}
    \setlength\mylenA{\the\wd\myboxA}
    \addtolength\mylenA{-\the\wd\myboxB}%
    \ifdim\wd\myboxB<\wd\myboxA%
       \rlap{\hskip 0.5\mylenA\usebox\myboxB}{\usebox\myboxA}%
    \else
        \hskip -0.5\mylenA\rlap{\usebox\myboxA}{\hskip 0.5\mylenA\usebox\myboxB}%
    \fi}
\makeatother

\usepackage{siunitx}
\DeclareSIUnit\px{P}

\usepackage{textcomp}

\usepackage{booktabs}
\usepackage{array}

\newcolumntype{+}{>{\global\let\currentrowstyle\relax}}
\newcolumntype{^}{>{\currentrowstyle}}
\newcommand{\rowstyle}[1]{\gdef\currentrowstyle{#1}%
#1\ignorespaces
}
\usepackage{tabu}
\usepackage[table,xcdraw]{xcolor}

\usepackage{pifont}
\newcommand{\cmark}{\ding{51}}%
\newcommand{\xmark}{\ding{55}}%

\usepackage{soul} 

\usepackage{units}

\usepackage{acro}
\newcommand{\acfullr}[1]{\acs{#1} (\acl{#1})}

\ifpdf
\usepackage[pdftex]{hyperref}
\hypersetup{
	pdftitle=\mydoctitle,
	pdfauthor=\mydocauthor,
	pdfkeywords=\mydockeywords,
	plainpages=false,
	bookmarksopen=true,
	bookmarksopenlevel=1,
	bookmarksnumbered,
	final,
	pdfstartview=FitH,
  hidelinks,
	colorlinks=false}
\fi

\usepackage{cleveref}

\usepackage[textwidth=1.9cm]{todonotes}

\usepackage[inkscape=false]{./config/svg}
\setsvg{inkscapearea=page, extract=false}
\svgpath{./figs/}
\usepackage{dblfloatfix}

\usepackage{afterpage}

\usepackage{graphicx}
\usepackage[]{subfig}
\graphicspath{{./figs/}}
\DeclareGraphicsExtensions{.pdf,.jpeg,.png,.jpg}

\DeclareAcronym{SLAM}{ short=SLAM, long={simultaneous localisation and mapping}}
\DeclareAcronym{SPLAM}{ short=SPLAM, long={simultaneous planning, localisation and mapping}}
\DeclareAcronym{LiDAR}{ short=LiDAR, long={light detection and ranging}}
\DeclareAcronym{IRF}{ short=IRF, long={inertial reference frame}}
\DeclareAcronym{BRF}{ short=BRF, long={body reference frame}}
\DeclareAcronym{PGM}{ short=PGM, long={probabilistic graphical model}}
\DeclareAcronym{BP}{ short=BP, long={belief propagation}}
\DeclareAcronym{LBP}{ short=LBP, long={loopy belief propagation}}
\DeclareAcronym{LBU}{ short=LBU, long={loopy belief update}}
\DeclareAcronym{surfel}{ short=surfel, long={surface element}}
\DeclareAcronym{voxel}{ short=voxel, long={volumetric element}}
\DeclareAcronym{pose}{ short=pose, long={position and orientation}}
\DeclareAcronym{NDT}{ short=NDT, long={normal distributions transform}}
\DeclareAcronym{GP}{ short=GP, long={Gaussian process}, long-plural=es}
\DeclareAcronym{SDF}{ short=SDF, long={signed distance function}}
\DeclareAcronym{TSDF}{ short=TSDF, long={truncated signed distance function}}
\DeclareAcronym{MLS}{ short=MLS, long={multi-level surface}}
\DeclareAcronym{PML}{ short=PML, long={probabilistic multi-level}}
\DeclareAcronym{ROC}{ short=ROC, long={receiver operating characteristic}}
\DeclareAcronym{BCM}{ short=BCM, long={Bayesian committee machine}}
\DeclareAcronym{UT}{ short=UT, long={unscented transform}}
\DeclareAcronym{RKHS}{ short=RKHS, long={reproducing kernel Hilbert space}}
\DeclareAcronym{KL}{ short=KL, long={Kullback-Leibler}}
\DeclareAcronym{EP}{ short=EP, long={expectation propagation}}
\DeclareAcronym{VMP}{ short=VMP, long={variational message passing}}
\DeclareAcronym{IID}{ short=IID, long={independently and identically distributed}}
\DeclareAcronym{STM}{
  short=STM,
  long={stochastic triangular mesh},
  short-indefinite={an},
  long-indefinite={a} }
\DeclareAcronym{HYMM}{ short=HYMM, long={hybrid metric map}}
\DeclareAcronym{MCMC}{ short=MCMC, long={Markov chain Monte Carlo}}
\DeclareAcronym{MAP}{ short=MAP, long={maximum a posteriori}}
\DeclareAcronym{MSE}{
  short=MSE,
  long={mean squared error},
  short-indefinite={an},
  long-indefinite={a} }
\DeclareAcronym{DGPS}{ short=DGPS, long={differential global positioning system}}
\DeclareAcronym{LIBELAS}{ short=LIBELAS, long={library for efficient large-scale stereo matching}}
\DeclareAcronym{MRF}{ short=MRF, long={Markov random field}}
\DeclareAcronym{DAG}{ short=DAG, long={directed acyclic graph}}
\DeclareAcronym{KF}{ short=KF, long={Kalman filter}}
\DeclareAcronym{EKF}{ short=EKF, long={extended Kalman filter}}
\DeclareAcronym{UKF}{ short=UKF, long={unscented Kalman filter}}
\DeclareAcronym{LM}{ short=LM, long={Levenberg-Marquardt}}
\DeclareAcronym{PnP}{ short=PnP, long={Perspective-n-Point}}
\DeclareAcronym{IMU}{ short=IMU, long={inertial measurement unit}}
\DeclareAcronym{EMDW}{ short=EMDW, long={elementary, my dear Watson}}

\begin{document}

\newgeometry{left=25mm,right=25mm,top=25mm,bottom=20mm}
\begin{frontmatter}
  \title{\mydoctitle}
  \author[1]{Clint D. Lombard\corref{cor1}}
  \ead{clint dot lom at gmail dot com}
  \author[1]{Corn\'e E. van Daalen}
  \ead{cvdaalen at sun dot ac dot za}

  \cortext[cor1]{Corresponding author}

  \address[1]{Department of Electrical and Electronic Engineering, Stellenbosch
    University, Stellenbosch, South Africa}

  \begin{abstract}
  For mobile robots to operate autonomously in general environments, perception
  is required in the form of a dense metric map. For this purpose, we present
  the \acf*{STM} mapping technique: a 2.5-D representation of the surface of the
  environment using a continuous mesh of triangular surface elements, where each
  surface element models the mean plane and roughness of the underlying surface.
  In contrast to existing mapping techniques, \iacs*{STM} map models the
  structure of the environment by ensuring a continuous model, while also being
  able to be incrementally updated with linear computational cost in the number
  of measurements. We reduce the effect of uncertainty in the robot
  \acfullr{pose} by using landmark-relative submaps. The uncertainty in the
  measurements and robot pose are accounted for by the use of Bayesian inference
  techniques during the map update. We demonstrate that \iacs*{STM} map can be
  used with sensors that generate point measurements, such as \acf*{LiDAR}
  sensors and stereo cameras. We show that \iacs*{STM} map is a more accurate
  model than the only comparable online surface mapping technique---a standard
  elevation map---and we also provide qualitative results on practical datasets.

  \vspace{-1em}
  \section*{Graphical Abstract}
  {\center \includegraphics[width=\textwidth]{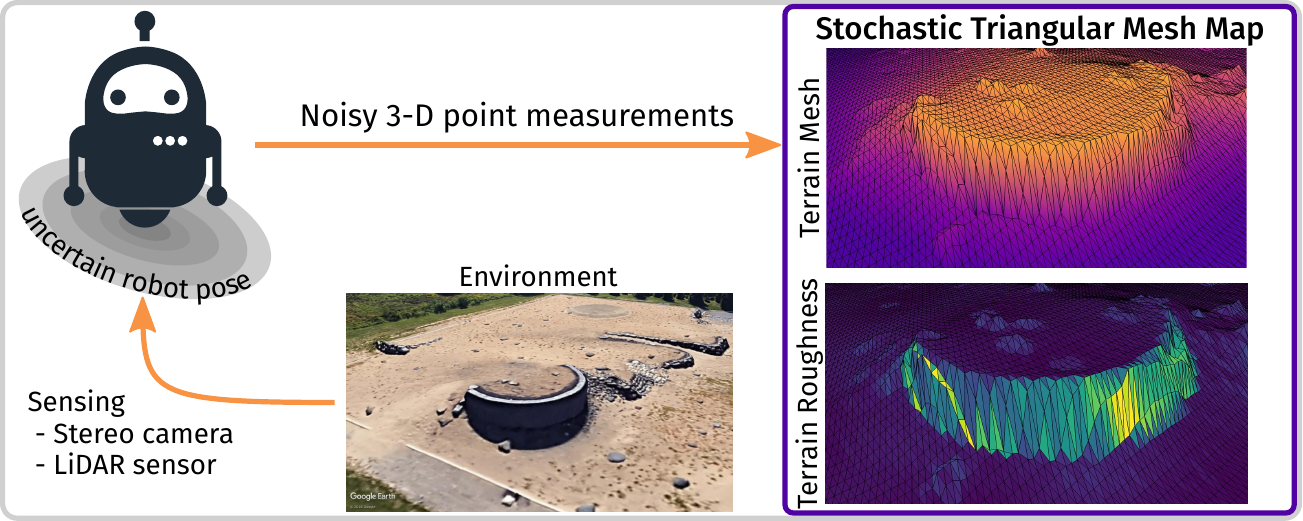}} 

  \begin{minipage}[t]{.48\textwidth}
    \raggedright
    \section*{Highlights}
    \begin{itemize}
      \setlength\itemsep{0.1em}
    \item A novel dense mapping technique that uses a stochastic triangular mesh is
      presented.
    \item The technique handles uncertainty from multiple sources in a principled
      manner.
    \end{itemize}
  \end{minipage}%
  \begin{minipage}[t]{.48\textwidth}
    \raggedleft
    \begin{itemize}
      \setlength\itemsep{0.1em}
    \item The map models the structure in the environment while remaining tractable
      to update.
    \item The inference algorithm employs a novel combination of message passing
      techniques.
    \item The mapping technique is tested on a large-scale practical dataset.
    \end{itemize}
  \end{minipage}%
\end{abstract}

\begin{keyword}
  dense mapping \sep
  submapping \sep
  perception \sep
  triangular mesh \sep
  probabilistic graphical model \sep
  stereo cameras \sep
  LiDAR
\end{keyword}


\end{frontmatter}

\newgeometry{left=25mm,right=25mm,top=25mm,bottom=25mm}
\section{Introduction}
\label{sec:intro}

For an autonomous mobile robot to operate effectively, it needs the ability to
perceive its environment; this problem of perception is commonly referred to as
\emph{mapping}. More precisely, mapping can be defined as representing a robot's
belief over the environment, where ``belief'' refers to the knowledge about a
state, given all measurements and any prior information. In the general case of
a robot operating in an initially unknown environment, the map needs to be
incrementally built online using the robot's measurements of the environment,
such that the robot can make decisions in real time. These measurements are
obtained using exteroceptive sensors---typically a combination of \ac{LiDAR}
sensors, stereo cameras, and depth-cameras. In order to perform complex
tasks---such as path planning, collision detection, or object manipulation---a
3-D dense metric map is required. \citet{Guizilini2018} argue that, for a dense
metric map to be an effective representation, it should have the following
attributes:
\begin{itemize}
\item \emph{Reason under uncertainty.} Measurements from all robot sensors
  contain some degree of uncertainty, which needs to be accounted for when
  updating the map.

\item \emph{Incremental updates.} Due to the online nature of the problem, all
  the measurements are not available at once and, as the robot needs to
  continuously query the map, the map needs to be updated incrementally.

\item \emph{Update and query efficiently.} Exteroceptive sensors generate vast
  amounts of data---typically in the form of dense point clouds---that need to
  be incorporated quickly into the map. Additionally, the resulting map needs to
  be accessed quickly when required.

\item \emph{Represent the structure in the environment.} The environment is
  inherently structured according to spatial relationships on various length
  scales. By exploiting this structure, a map can interpolate occluded regions,
  or better estimate regions with few measurements.
\end{itemize}

In this paper, we present \acf{STM} mapping\footnote{Source code can be found at
  \url{https://github.com/clintlombard/stm-mapping}.}, an online dense mapping
technique designed around these four attributes. In particular, we focus on two
aspects:

\paragraph{Robot pose belief}
A source of uncertainty influencing dense mapping is \emph{localisation}; to
build an accurate model of the environment, one must also consider the belief
over the robot \ac{pose}. Estimating the robot \ac{pose} belief is typically
performed using the process of \ac{SLAM}, whereby the belief over a sparse
map---consisting of re-identifiable landmarks---is estimated in conjunction with
the robot \ac{pose}. The sparsity of this landmark map, however, prohibits it
being used for complex planning tasks, and it should therefore rather be
considered as a localisation aid. When performing \ac{SLAM}, the beliefs over
the robot \ac{pose} at any time step can change due to new information, as is
also the case when a robot revisits a part of the environment---that is, when
loop closure is performed. This can cause significant changes in the beliefs
over these previous poses, which present a problem, as measurements obtained at
previous poses would have been incorporated into the dense map using the
previous \ac{pose} beliefs. To account for this, measurements would need to be
removed and reintegrated into the map using the retroactively updated \ac{pose}
beliefs---an expensive undertaking.

\paragraph{Statistical dependencies}
An aspect of representing the structure of the environment in a map is to model
the continuity of surfaces. However, most dense mapping techniques are made
tractable through the assumption that neighbouring map elements are
statistically independent (as we will discuss in \Cref{sec:lit}). This allows
each element to be updated independently, but at the expense of modelling
continuity. As a consequence of this assumption, the resulting models contain
numerous discontinuities. This is problematic for a task like collision
prediction, for which it is desirable to have a continuous representation of the
environment in order for it to function accurately. The alternative approach of
modelling dependencies between map elements will lead to more accurate map
beliefs, but at a higher computational cost.

In order to address the issue of the uncertainty in the robot pose belief, we
develop the \ac{STM} mapping technique using triangular submapping regions,
demarcated by landmarks extracted using \acf*{SLAM}---the 2-D \acf*{HYMM}
framework \citep{Guivant2004, Nieto2006}. In order to motivate this decision, we
briefly explain the HYMM framework, and provide a 3-D extension to it
(\Cref{sec:irfs}). Using this submapping framework, we then develop our \ac{STM}
mapping technique, representing the structure of the environment using a
continuous surface (\Crefrange{sec:mapmodel}{sec:model-inference}). Finally, we
present some experimental results, building \ac{STM} maps using both simulated
and practical data (\Cref{sec:results}).


\section{Existing Approaches to Dense Mapping}
\label{sec:lit}

To motivate our proposal for a new mapping technique, we now review and evaluate
the existing approaches to dense mapping, and specifically consider the desired
attributes a dense map should have.

\subsection{Polygonal Mesh Maps}
Polygonal meshes are a well-known method of spatial modelling used in computer
graphics. A mesh is constructed from a point cloud by linking points together
with edges, forming polygons. In an early application, \citet{Thrun2004}
incrementally built sections of a triangular mesh of indoor environments,
subsequently simplifying areas of the mesh by fitting rectangular planes using
an online implementation of expectation maximisation. More recently,
\citet{Wiemann2016} used a $k$-nearest-neighbours approach to estimate the
normals of a triangular mesh, creating a consistent representation of planar
surfaces. The resulting mesh was simplified by fusing local polygons with
similar normal vectors; however, this process was calculated in post-processing.
\citet{Zienkiewicz2016} modelled small-scale environments with a fixed-topology
triangular mesh, and formulated the update procedure probabilistically. By
performing weighted optimisation, they were able to iteratively and
incrementally obtain solutions to the mesh. Although their approach is formatted
probabilistically, it does not capture the uncertainty in the model. Polygonal
meshes are not designed to explicitly represent any uncertainty, and therefore
cannot incorporate the robot pose belief or sensor uncertainty in a principled
manner. They are, however, a popular method for visualising other probabilistic
techniques.

\subsection{Occupancy Grid Maps}
Occupancy grid mapping is one of the most widely used mapping techniques and is
considered the de facto representation for dense mapping. Originally developed
in 2-D by \citet{Moravec1985}, this technique discretises the map into a
volumetric grid, where each \ac{voxel} stores the probability that the
associated region is occupied. As with all volumetric representations in 3-D,
spanning the mapping space with a grid becomes prohibitively expensive, despite
the majority of the mapped space being either unoccupied or unobserved. To
address this, \citet{Hornung2013} developed the OctoMap framework---this is
arguably the most used dense mapping framework. Instead of a fixed voxel size,
they used octrees to recursively adjust the voxel resolution. Unimportant
information, namely the unoccupied or unobserved space, is compressed into
coarse voxels, while a finer resolution is maintained for occupied regions.
Improving on this, \citet{Einhorn2011} developed an adaptive online method of
determining the division depth in local regions of the map using the more
general k-d tree. \citet{Khan2015} used rectangular voxels to compress
unoccupied cubic voxels. In recent work, \citet{Droeschel2017} used an
allocentric (robot-centric) grid, where the resolution increases with the
distance from the robot. Their approach was also implemented practically, and
tested at the DARPA Robotics Challenge. In order to handle the uncertainty of
the robot pose belief, \citet{Joubert2014} incorporated a beam sensor model into
occupancy map updates using Monte Carlo integration, although this approach was
only demonstrated in 2-D. In an attempt to incorporate statistical dependency
between 2-D voxels, \citet{Thrun2003a} used a forward measurement model and, in
an offline post-processing procedure, optimised for the maximum likelihood map.

Despite their popularity, occupancy grid maps fail to incorporate the
dependencies between map elements in an online manner.

\subsection{Elevation Maps}
The most general way of representing an environment is to use a 3-D dense map;
however, some environments are adequately represented as a 2.5-D map---that is,
a single axis\footnote{Typically the vertical axis.} is constrained to only one
value per map element. These elevation maps reduce the dimensionality of the map
by associating a single height to each element of a 2-D grid. Early work by
\citet{Hebert1989} used an elevation map to perform localisation, as well as to
identify footholds for a legged robot. A drawback of elevation maps is their
inability to represent overhangs, such as bridges. \citet{Triebel2006} dealt
with this problem in their method of \ac{MLS} mapping by clustering the
measurements in each map element, with the clusters across neighbouring map
elements being segmented into elevation classes. Recent work by
\citet{Fankhauser2018} incorporated the uncertainty in robot pose belief into an
allocentric elevation map. They achieved this by maintaining a distribution over
the spatial uncertainty of each cell, which is updated over time based on the
uncertainty in transformations between the allocentric reference frames. The
resultant map is constructed from the weighted average of neighbouring cells,
based on each cell's spatial uncertainty. Their approach was demonstrated for a
legged robot. In gamma-SLAM, \citet{Marks2009} represent the environment using a
2.5-D precision map, where measurements in each map element are modelled as
samples from a Gaussian distribution over the elevation, with an unknown mean
and precision. By considering only the precision---marginalising out the
mean---their method generates accurate maps for localisation, while also
providing a pseudo-metric for traversability. However, their model fails to
account for the case where the sensor uncertainty varies at different ranges,
and the model disregards height.

Elevation maps are an efficient and effective method of representing
environments in which a full 3-D representation is unnecessary. However,
elevation maps fail to incorporate the statistical dependencies between map
elements.

\subsection{Signed Distance Function Maps}
Over the last decade, advances in depth-camera technology have created
affordable and relatively accurate depth sensors. This has given rise to a
non-parametric surface representation of the environment using the \ac{SDF},
commonly used in computer graphics. The SDF calculates the Euclidean distance to
the nearest surface, defining positive values to indicate free space and
negative values to indicate occluded regions. Consequently, the surface is
implicitly described at the zero-crossings. In KinectFusion, a technique
pioneered by \citet{Newcombe2011}, measurements are fused into a regular 3-D
grid of voxels, storing truncated SDF values. Due to memory constraints, this
method is only suitable to medium-scale environments. \citet{Whelan2015}
expanded this to handle large-scale environments by using a sliding window to
maintain a local section of the environment as an SDF map, while incrementally
converting the regions exiting this window into a triangular mesh. A major
advantage of this method is its ability to adjust the map upon a loop closure.
This is achieved through an optimisation procedure over the sensor poses and
vertices of the meshed map---creating a globally consistent map. BundleFusion, a
method devised by \citet{Dai2017}, performs bundle adjustment to optimise over
the sensor poses to create accurate SDF maps. Similarly to KinectFusion, this
method suffers from memory constraints due to maintaining the complete map in a
grid, and is therefore limited to small-scale environments. The standard SDF
representation is not probabilistic, and consequently neither sensor uncertainty
nor pose uncertainty can be incorporated when updating the map. To address this,
\citet{Dietrich2016} modelled each distance as Gaussian distributed. However,
this choice of distribution can yield negative estimates for a positive-only
quantity (distance).

Although SDF methods have been shown to generate highly detailed maps in
real time, they have not been used widely in the robotics community due to
several drawbacks. Firstly, the SDF can only implicitly describe the surfaces in
the environment; when an explicit surface map is required, the SDF map needs to
be converted to a mesh. Secondly, as the standard SDF representation is not
probabilistic, it cannot model the statistical dependencies between mapping
elements. Finally, SDF methods are limited to a single sensor type, namely depth
cameras.

\subsection{Normal Distribution Transform Maps}
The \ac{NDT} mapping model was originally developed by \citet{Biber2003} as a
method for 2-D scan matching, and then independently expanded to 3-D by
\citet{Takeuchi2006}, and \citet{Magnusson2007}. The resulting volumetric
representation maintains a 3-D Gaussian distribution in each voxel.
Consequently, measurements falling within a voxel are assumed to be \ac{IID}
samples drawn from a Gaussian distribution, where the sufficient statistics,
namely the mean and covariance, can be calculated incrementally.
\citet{Stoyanov2013} showed that the resulting maps are accurate spatial
representations when compared with the methods of occupancy grid mapping and
polygonal meshes. They analysed all the methods on simulated and real-world
data, comparing \ac{ROC} curves, accuracy, and runtime. \ac{NDT} mapping was
augmented by \citet{Saarinen2013} to include the dimension of occupancy, thus
creating the \ac{NDT} occupancy map (NDT-OM). Their representation allows
multi-resolution support for the \ac{NDT}, while also introducing a temporal measure
to handle dynamic environments.

Due to the IID assumption, a drawback of the \ac{NDT} is that it cannot account for
heterogeneous uncertainty, which is present in the sensor models and the robot
pose belief. The map elements are also assumed to be statistically independent.

\subsection{Gaussian Process Maps}
A popular method of incorporating map element dependencies uses \ac{GP}
regression \citep{Rasmussen2006}---a method of regression using a non-parametric
stochastic model. A set of input training points (measurements of the
environment) are used to predict the output at some desired query points. In
order to perform this regression, the correlation between points is described by
a kernel (covariance), which is parameterised by a set of hyperparameters.
\acp{GP} can only perform regression on a 1-D output; due to this, \acp{GP} were
initially applied to 2.5-D elevation maps. \citet{Lang2007} used an iterative,
locally adaptive non-stationary kernel, which was able to handle sharp
discontinuities without severe smoothing. Extending this, \citet{Plagemann2009}
used a separate \ac{GP} to estimate the hyperparameters of each kernel. An
ensemble of overlapping \acp{GP} was also tiled to cover the mapping region.
This method was practically applied to foothold detection of a quadruped robot.
\citet{Vasudevan2012} implemented a dependent \ac{GP} to perform offline
large-scale terrain modelling on scans of an open mine. \citet{Hadsell2010} used
a non-stationary kernel of which the hyperparameters were a function of the
uncertainty of the measurement range.

In order to use \ac{GP} regression in full 3-D space, the environment needs to
be described implicitly; one method of doing this uses occupancy. Occupancy maps
in the context of \ac{GP} mapping were first proposed by \citet{OCallaghan2012},
who developed a framework for constructing 2-D \ac{GP} occupancy maps (GPOMs).
To determine the probability of occupancy, an additional probabilistic
least-squares classifier was used. This method was able to incorporate
uncertainty in the sensor measurements and the robot pose into the map using the
unscented transform and Gauss-Hermite quadrature. \citet{Jadidi2017} improved on
GPOMs by considering the case in which the pose uncertainty is significant, and
incorporated this uncertainty into a warped GPOM representation using
Gauss-Hermite quadrature and Monte Carlo integration. Both these methods,
however, were offline post-processing procedures.

The standard \ac{GP} formulation cannot be applied to online mapping due to two
main issues. Firstly, it is required that all the training data be available at
once. Secondly, performing online \ac{GP} regression is intractable due the
cubic computational complexity in the amount of training data. An approximate
method of addressing both of these issues is to partition the training data
using a \ac{BCM} \citep{Tresp2000}, which is a method of combining estimators
trained on different data by assuming conditional independence. \ac{BCM}s have a
linear computational complexity in the partition size and have been shown to
decrease computational time---matching that of even the OctoMap framework
\citep{Kim2014, Wang2016}. This method, however, combines multiple independently
trained estimators, which is an inaccurate representation of the map belief.

Although \acp{GP} do not assume statistical independence between map elements,
this comes at a prohibitive computational cost and, for this reason, \acp{GP}
have primarily been limited to offline mapping. Their performance is also
largely dependent on suitable hyperparameters, which can require optimisation
throughout the mapping process. The resulting map is also a jointly Gaussian
distribution over all the query points, which can become prohibitively expensive
to maintain.

\subsection{Hilbert Maps}
A more recent method of incorporating the dependencies between map elements was
introduced by \citet{Ramos2016}. Their approach, Hilbert maps, aims to generate
a continuous occupancy representation of the environment. To achieve this,
spatial measurements are projected onto a high-dimensional feature space, in
which a simple linear classifier, namely logistic regression, is incrementally
trained using stochastic gradient descent. The resulting representation is a
parametric occupancy map of the environment. The robot pose belief and sensor
uncertainties are also incorporated into the features using numerical
integration. To improve efficiency, \citet{Doherty2016} used the approach of
fusing local maps, whereby multiple logistic regression classifiers could be
combined incrementally. \citet{Guizilini2018} proposed an efficient Hilbert maps
approach by clustering measurements to decrease the dimensionality of the
feature vectors. This efficient extension to the Hilbert maps framework was able
to achieve training speeds rivalling that of the OctoMap framework. However, as
a grid is not specified during training, this results in significantly poorer
performance when querying the map in comparison to OctoMap.

Hilbert maps do not represent the uncertainty in the resulting map and therefore
cannot truly be considered a probabilistic representation of the environment.
The Hilbert maps framework has also only been demonstrated using accurate LiDAR
sensors, and whether the approximations made will expand for less accurate
sensors---like stereo cameras---is unclear.

\subsection{Evaluation}
\label{sec:evaluation}
The main attributes of the existing mapping techniques are summarised in
\Cref{tab:mapcompare}. We also include our proposed mapping
technique---\acf{STM} mapping. From this we highlight a few important points
related to the existing techniques:

\begin{table*}[h!]
\small
\centering
\caption{Comparison of the different dense mapping techniques across different
  attributes. We also show our \acf{STM} mapping technique for comparison.}
\label{tab:mapcompare}
\renewcommand*{\arraystretch}{1.4}
\rowcolors{2}{black!10}{white}
\begin{tabular}{
  @{}
  +>{\centering\arraybackslash}m{0.16\textwidth}
  ^>{\centering\arraybackslash}m{0.2\textwidth}
  ^>{\centering\arraybackslash}m{0.12\textwidth}
  ^>{\centering\arraybackslash}m{0.11\textwidth}
  ^>{\centering\arraybackslash}m{0.16\textwidth}
  ^>{\centering\arraybackslash}m{0.1\textwidth}
  @{}}
  \toprule
  \rowstyle{\bfseries}
  Mapping Technique & Incorporate \linebreak Uncertainty & Incremental \linebreak Updates &
  Update \linebreak Efficiency & Statistical \linebreak Dependencies & Explicit
  \linebreak Surface\\
  \midrule
  Polygonal Mesh & \xmark & \xmark & $\bigO{N}$ &  \xmark & \cmark\\
  Occupancy Grid & \cmark & \cmark & $\bigO{N}$ & \xmark & \xmark\\
  Elevation & \cmark & \cmark & $\bigO{N}$ & \xmark & \cmark\\
  \acs{SDF} & \xmark & \cmark & $\bigO{N}$ & \xmark & \xmark\\
  \acs{NDT}  & Not in measurements & \cmark & $\bigO{N}$ & \xmark & \cmark\\
  \acs{GP}  & \cmark & \xmark & $\bigO{N^3}$ & \cmark & \cmark\\
  Hilbert & Not in the map & \cmark & $\bigO{N}$ & \cmark & \xmark\\
  \acs{STM} & \cmark & \cmark & $\bigO{N}$ & \cmark & \cmark
\end{tabular}
\end{table*}

\paragraph{Robot pose belief uncertainty}
The majority of probabilistic mapping techniques---bar \ac{NDT} mapping---have the
ability to account for the uncertainty in the robot pose belief by marginalising
it out using numerical integration\footnote{Another method of considering the
  robot pose belief---applicable to all mapping techniques---would be using a
  Rao-Blackwellised particle filter \citep[e.g.][]{Grisetti2007}. Here a set of
  random samples (particles) of the robot pose belief each maintain a map.
  However, this approach does not scale well in 3-D, as it requires maintaining
  multiple maps, which is already expensive for a single dense map.}; this
increases the uncertainty in the measurements, consequently increasing the map
uncertainty. Although this represents the belief over the environment more
accurately, if the uncertainty in the robot pose belief is significant, this
could render the map essentially useless. Additionally, marginalising out the
robot pose belief removes the dependencies between poses. This is again a
problem when there are significant changes in the beliefs over previous poses,
as the affected areas of the map need to be recalculated. A popular approach to
alleviate these issues is to perform submapping, which we discuss in depth later
(\Cref{sec:irfs}).

\paragraph{Statistical dependencies}
Most existing mapping techniques cannot incorporate statistical dependencies
between map elements. Of the two main techniques that can, \ac{GP} mapping comes at an
intractable computational cost, and Hilbert maps do not capture any uncertainty
in the resulting map, which does not accurately model the belief over the
environment. In \ac{STM} mapping, we enforce continuity in the model, which
causes statistical dependencies between map elements; however, this only comes
at a linear computational cost and still maintains a probabilistic map
(\Cref{sec:mapmodel}).

\paragraph{Explicit surface models}
The existing dense mapping techniques either model the underlying surfaces in
the environment explicitly or implicitly. An explicit surface representation is
arguably more useful, because some methods of collision prediction
\citep{Jones2006,VanDaalen2009} require an explicit surface representation. An
explicit surface can be extracted from an implicit surface representation;
however, this requires an extra step of computation, and results in a model
which does not represent any surface uncertainty. For example, an occupancy grid
map could be converted to an explicit surface by thresholding the probability of
occupancy. We therefore opt for an explicit surface representation of the
environment (\Cref{sec:mapmodel}).

\paragraph{Incremental updates}
Being able to incrementally update the map is critical for online operation.
However, to produce an accurate map belief, it is just as important to use
Bayesian reasoning when updating the map incrementally. Specifically, the map's
current state---based on prior measurements---should be considered as context
when fusing new measurements into the map; that is, we should use Bayes'
theorem. The probabilistic techniques that use incremental Bayesian
updates---occupancy grid, elevation, and \ac{NDT} maps---do not incorporate
statistical dependencies. Despite being a Bayesian technique, \ac{GP} maps
require the \ac{BCM} approximation to perform incremental updates, and this
approximation neglects the map's current state when updating. Additionally, the
Hilbert maps method cannot represent the uncertainty in its estimates of the
map, as it follows a frequentist approach. In \ac{STM} mapping, we incorporate
incremental Bayesian updates into a continuous representation of the environment
(\Cref{sec:mapmodel}).

Given our evaluation of the existing dense mapping techniques and the comparison
to our proposed technique (\Cref{tab:mapcompare}), we believe that the
\acf*{STM} mapping technique constitutes a valuable contribution to the field of
dense mapping for mobile robots. In the remainder of this paper, we develop and
test the \ac{STM} mapping technique.


\section{Inertial Reference Frames in Dense Mapping}
\label{sec:irfs}

The uncertainty in the robot pose belief is ever present when performing online
dense mapping. If the robot pose belief contains significant uncertainty, it
creates problems when incorporating measurements into the dense map. In this
chapter, we explore submapping---an approach that reduces the effect of the
uncertainty in the robot pose belief---and motivate our choice of submapping
framework in the \ac{STM} mapping technique.

All robotic systems are described in terms of \acp{IRF}. When discussing the
issue of the uncertainty in the robot pose belief in dense mapping, it is
important to understand the effect that the choice of \ac{IRF} has on the
resulting map. Some mapping techniques have used an allocentric \ac{IRF} to
ensure that the region of the map currently surrounding the robot has lower
uncertainty \citep{Fankhauser2018} or a finer grid resolution
\citep{Droeschel2017}. In contrast, most mapping techniques focus on building
globally consistent maps in a fixed, single-privileged \ac{IRF}---often referred
to as a global \ac{IRF}. Due to the compounding effect of uncertain motion on
the robot pose belief, regions of the map further away from the origin of the
chosen \ac{IRF} are increasingly uncertain; in the absence of absolute position
sensing, this uncertainty is unbounded. Furthermore, when performing \ac{SLAM}
in a global \ac{IRF} and then performing loop closure, large changes in beliefs
over previous robot poses can be induced. Despite this, the \emph{relative}
uncertainty between successive robot poses is bounded by the---typically
confident---motion dynamics. The submapping approach takes advantage of this by
segmenting the environment into several submaps, each with its own \ac{IRF}.
This approach is used in some of the recent state-of-the-art vision-based
\ac{SLAM} techniques, namely LSD-SLAM by \citet{Engel2014}, and ORB-SLAM by
\citet{Mur2017}; both of these techniques use keyframes as intermediate submaps
during \ac{SLAM}. In the case of dense mapping in indoor environments, the map
can be semantically segmented by room \citep{Friedman2007}. In general
environments, a typical method of defining submaps uses overlapping rectangular
cuboids to span the mapping space \citep{Konolige2011,Schmuck2016}. However,
this causes two issues: the map contains redundant regions, and it is
non-trivial to accurately describe new measurements in terms of the \acp{IRF} of
previously observed submaps. \Acp{HYMM} \citep{Guivant2004, Nieto2006} is a 2-D
submapping technique that addresses both of these issues by using relative
\acp{IRF} based on the sparse landmark map obtained from \ac{SLAM}. In
\Cref{sec:relat-inert-refer}, we review the HYMM framework and present a 3-D
extension to it. Utilising this framework, we present a principled method of
decoupling measurements of the environment from the robot pose belief
(\Cref{sec:decouple}).

\paragraph{Notational remark} A point in the global \ac{IRF}, $\mvec{x}$, is
expressed in the \ac{BRF} of the robot as $\mvec{x}^B$, and in a relative \ac{IRF} as
$\mvec{x}^R$.

\subsection{Relative Inertial Reference Frames}
\label{sec:relat-inert-refer}

The HYMM submapping technique segments the environment by constructing a
triangular mesh between selected landmarks (\Cref{fig:lm_tri}). A relative \ac{IRF}
is associated with each triangular submap, defined in terms of the three
landmarks at the vertices of the triangular region. To elucidate this
definition, consider a submap demarcated by the convex hull of the landmarks
$\mvec{l}^{}_0$, $\mvec{l}^{}_{\alpha}$ and $\mvec{l}^{}_{\beta}$. A 2-D point,
$\mvec{m}^{}$, in the global \ac{IRF} can be described by
\begin{equation}
  \label{eq:2drel}
  \begin{split}
    \mvec{m}^{}
    &= \alpha (\mvec{l}^{}_{\alpha} - \mvec{l}^{}_0) +
    \beta  (\mvec{l}^{}_{\beta} - \mvec{l}^{}_0) +\mvec{l}^{}_0 \\
    &= \alpha  \mvec{a} +\beta  \mvec{b} + \mvec{l}^{}_0.
  \end{split}
\end{equation}
For $\mvec{m}$ to be within the submap, $\alpha$ and $\beta$ must
satisfy
\begin{equation}
  \label{eq:constraints}
  \alpha+\beta \leq 1 \quad \text{and} \quad \alpha,\beta \in [0,1].
\end{equation}
The 2-D relative \ac{IRF} axes are defined by $\mvec{a}$ and
$\mvec{b}$---$\mvec{l}^{}_0$ is an offset---and $(\alpha, \beta)$ are the
coordinates of a 2-D point $\mvec{m}$ in the relative \ac{IRF}.

To extend the 2-D relative \acp{IRF} of the HYMMs to 3-D, a third axis is required.
One could project the 3-D landmarks onto the horizontal plane and use the
vertical axis of the global \ac{IRF}. However, this approach couples the relative \ac{IRF}
to the global \ac{IRF}, negating the true benefits of a relative \ac{IRF}.
\citet{Guivant2004} suggested---but did not develop---two ideas of extending
HYMMs to 3-D. The relative \ac{IRF} axes could be constructed using four landmarks,
forming tetrahedral submapping regions; however, it could be impractical to
adequately span the mapping space. Alternatively, the 2-D approach can be
adapted to 3-D by defining the submapping region being normal to the triangular
faces of the 3-D landmark mesh. We develop the latter approach, extending the
relative \ac{IRF} to 3-D by introducing the third axis as
\begin{equation}
  \mvec{n} = \frac{\mvec{a} \times \mvec{b}}{\norm{\mvec{a} \times \mvec{b}}}.
\end{equation}
Using this definition, a 3-D point, $\mvec{m}$, can be described as
\begin{equation}
  \label{eq:3drel}
  \begin{split}
    \mvec{m}^{}
    &= \alpha \mvec{a} +\beta \mvec{b} + \gamma \mvec{n} +\mvec{l}^{}_0\\
    &=
    \begin{bmatrix}
      \mvec{a} & \mvec{b} & \mvec{n}
    \end{bmatrix}
    \mvec{m}^R + \mvec{l}^{}_0,
  \end{split}
\end{equation}
where $\mvec{m}^R=\begin{bmatrix}\alpha & \beta & \gamma\end{bmatrix}^\intercal$
is a point in the 3-D relative \ac{IRF} (\Cref{fig:rel_pt}). The constraints in
\Cref{eq:constraints} also hold for $\mvec{m}$ to fall within the submap.

\begin{figure*}[h!]
  \begin{minipage}{\textwidth}
    \renewcommand\thempfootnote{$\dagger$}
    \centering
    \subfloat[]{
      \includesvg[width=0.48\textwidth]{lm_triangulation}
      \label{fig:lm_tri}}
    \subfloat[]{
      \includesvg[width=0.45\textwidth]{3d_pt}
      \label{fig:rel_pt}}
    \caption{(a) The \acf{HYMM} submapping method \citep{Guivant2004,Nieto2006}
      segments the environment using landmarks\protect\footnote{The robot icon was
        adapted from an original by SimpleIcon from Flaticon.}. (b) In our 3-D
      extension to the HYMM representation, a point $\mvec{m}$ within a submap can
      be fully described relative to these landmarks.}
  \end{minipage}
\end{figure*}

There is, however, a caveat to using this relative \ac{IRF} description. As the
definition of the relative \ac{IRF} relies on landmarks, it is necessary for the
chosen landmarks to be robustly and persistently identifiable. While this is
still an open problem for general environments, given the progress that has been
made in the past decade in feature and place recognition \citep{Lowry2016}, we
believe that this will be possible in the future. Despite this, there are
practical applications, in more controlled environments, where easily
identifiable landmarks could be placed manually in the environment---for example
in land surveying, 3-D object reconstruction, or factory operation. We
practically demonstrate an example of such an application in
\Cref{sec:relative-irf-experiment}.

\subsection{Decoupling Dense Measurements from the Robot Pose Belief}
\label{sec:decouple}

With the relative \ac{IRF} definition in hand, we now present a method of decoupling
measurements of the surface of the environment from their associated robot pose
beliefs. Consequently, this method decouples the process of localisation from
that of dense mapping. We achieve this decoupling by transforming surface
measurements from the \ac{BRF} of the robot to the relative \ac{IRF} of the
associated submap---as opposed to transforming to the global \ac{IRF}.

We model the process of observing a point on the surface of the environment (in
the \ac{BRF} of the robot) using a sensor beam model, whereby a noisy
measurement, $\mvec{z}^B$, is generated by a point, $\mvec{m}^B$, on the surface
of the environment. Most sensors observe multiple environment surface points at
a single time step. We denote the sequence of measurements as $\mset{Z}^B =
(\mvec{z}^B_i)$, and the associated environment surface points as $\mset{M}^B =
(\mvec{m}^B_i)$. Assuming an uninformative prior over $\mset{M}^B$, we can
calculate the belief over $\mset{M}^B$ as
\begin{equation}
  \label{eq:bel_m_body}
  \belief(\mset{M}^B)
  \triangleq
  p(\mset{M}^B|\mset{Z}^B)
  \propto \prod_{i} p(\mvec{z}^B_i|\mvec{m}^B_i),
\end{equation}
where the belief distribution, denoted by $\belief(\cdot)$, given all evidence,
is the posterior distribution over some random variables. The conditional
distribution, $p(\mvec{z}^B_i|\mvec{m}^B_i)$, describes the measurement model,
which is known beforehand for each sensor. Following from this, the belief over
$\mset{M}^B$ and the \ac{SLAM} states---the current robot pose, $\mvec{x}^{}$,
and the sparse landmark map, $\mset{L}^{}$, in the global \ac{IRF}---can be
factorised as
\begin{equation}
  \begin{split}
    \belief(\mvec{x}^{},\mset{L}^{},\mset{M}^B)
    &\triangleq
    p(\mvec{x}^{},\mset{L}^{},\mset{M}^B|\mset{U}^{},\mset{Y},\mset{Z}^B)\\
    &\approx
    \underbrace{
      p(\mvec{x}^{},\mset{L}^{}|\mset{U}^{},\mset{Y})
    }_{\belief(\mvec{x}^{},\mset{L}^{})}
    \underbrace{
      p(\mset{M}^B | \mset{Z}^B)
    }_{\belief(\mset{M}^B)}
    ,
  \end{split}
\end{equation}
where $\mset{U}$ is the sequence of robot control commands, and $\mset{Y}$ is
the sequence of landmark measurements. This factorisation makes the assumption
that $\mset{M}^B$ is conditionally independent of the \ac{SLAM} states given
$\mset{Z}^B$. It should also be noted that $\belief(\mvec{x}^{},\mset{L}^{})$ is
the result of performing \ac{SLAM}. Although some \ac{SLAM} algorithms only
calculate the \ac{MAP} estimate of this belief, we only consider \ac{SLAM}
algorithms that calculate the full belief distribution.

Going forward, we assume $\belief(\mvec{x}^{},\mset{L}^{},\mset{M}^B)$ to be
Gaussian distributed. This does not necessarily limit the \ac{SLAM} algorithm to one
that makes the Gaussian assumption, but simply requires an algorithm where the
resulting belief distribution can be projected onto the Gaussian family of
distributions---for example by using moment matching.

To transform $\mset{M}^B$ to the relative \ac{IRF}, it must first be transformed to
the same \ac{IRF} as the \ac{SLAM} states---the global \ac{IRF}. We use the \ac{UT}
\citep{Julier1997, Julier2002} to transform the environment surface points from
the body RF to the global \ac{IRF}, $\mset{M}^B \rightarrow \mset{M}^{}$, and then
from the global to the relative \ac{IRF}, $\mset{M}^{} \rightarrow \mset{M}^R$.
The transformation $\mset{M}^B \rightarrow \mset{M}^{}$ is determined by the
robot pose in the global \ac{IRF}, $\mvec{x}$. As a consequence of this,
$\mset{M}^{}$ and $\mvec{x}$ are statistically dependent, and therefore, in
general, the belief distribution over all the states in the global \ac{IRF},
$\belief(\mvec{x}^{},\mset{L}^{},\mset{M}^{})$, cannot be factorised. If dense
mapping was performed in the global \ac{IRF}, then the global \ac{IRF}
measurement belief, $\belief(\mset{M})$, would be incorporated into the map. In
order for this mapping process to be tractable, the environment points are
assumed to be statistically independent of one another; however, this is almost
never the case; because of the transformation to the global \ac{IRF}, they are
usually highly correlated with one another (as we will show subsequently).

\begin{figure*}[h!]
  \centering
  \subfloat[$\rho=0$]{
    \includegraphics[width=0.22\textwidth]{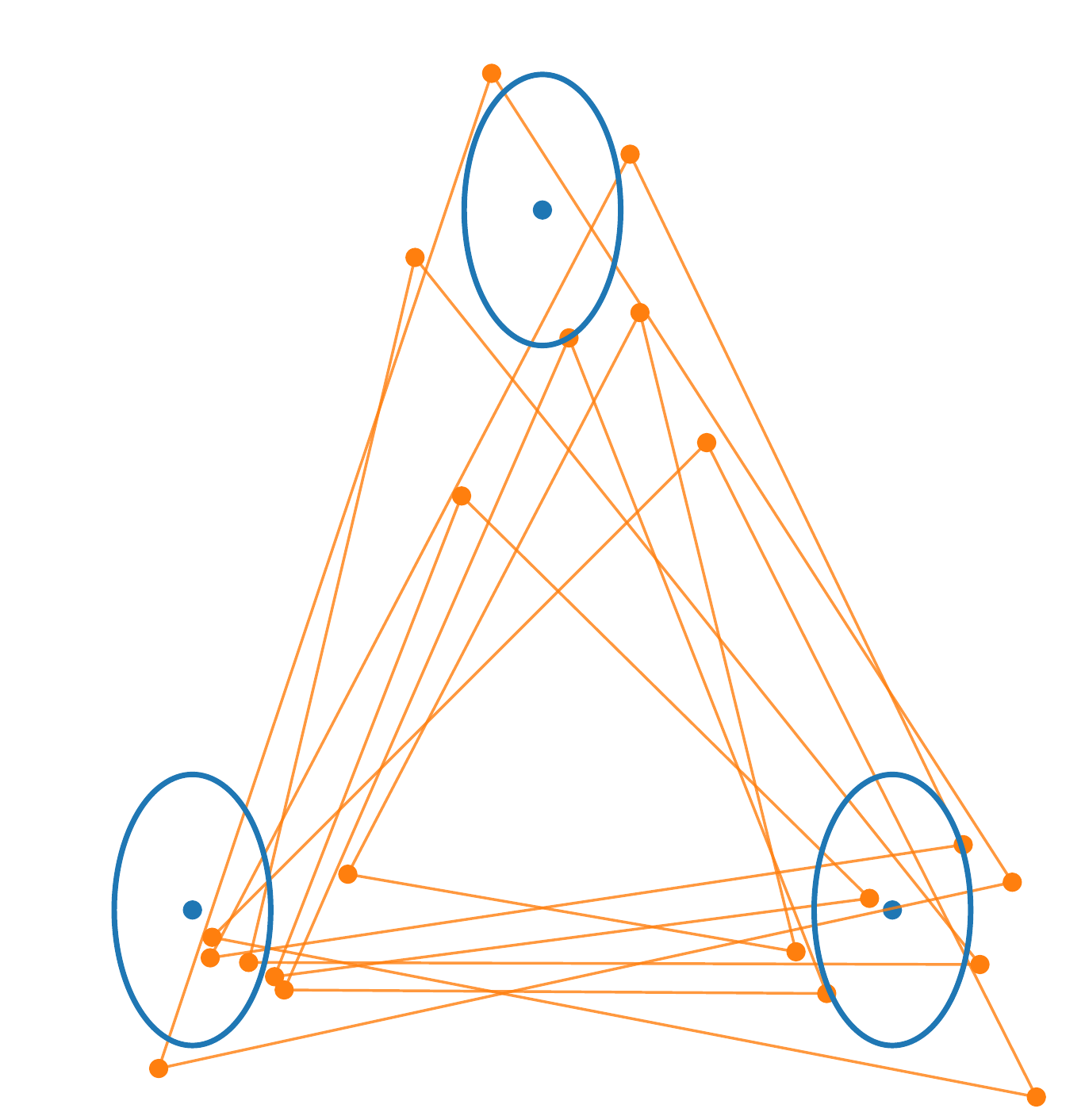}
  }
  \subfloat[$\rho=0.5$]{
    \includegraphics[width=0.22\textwidth]{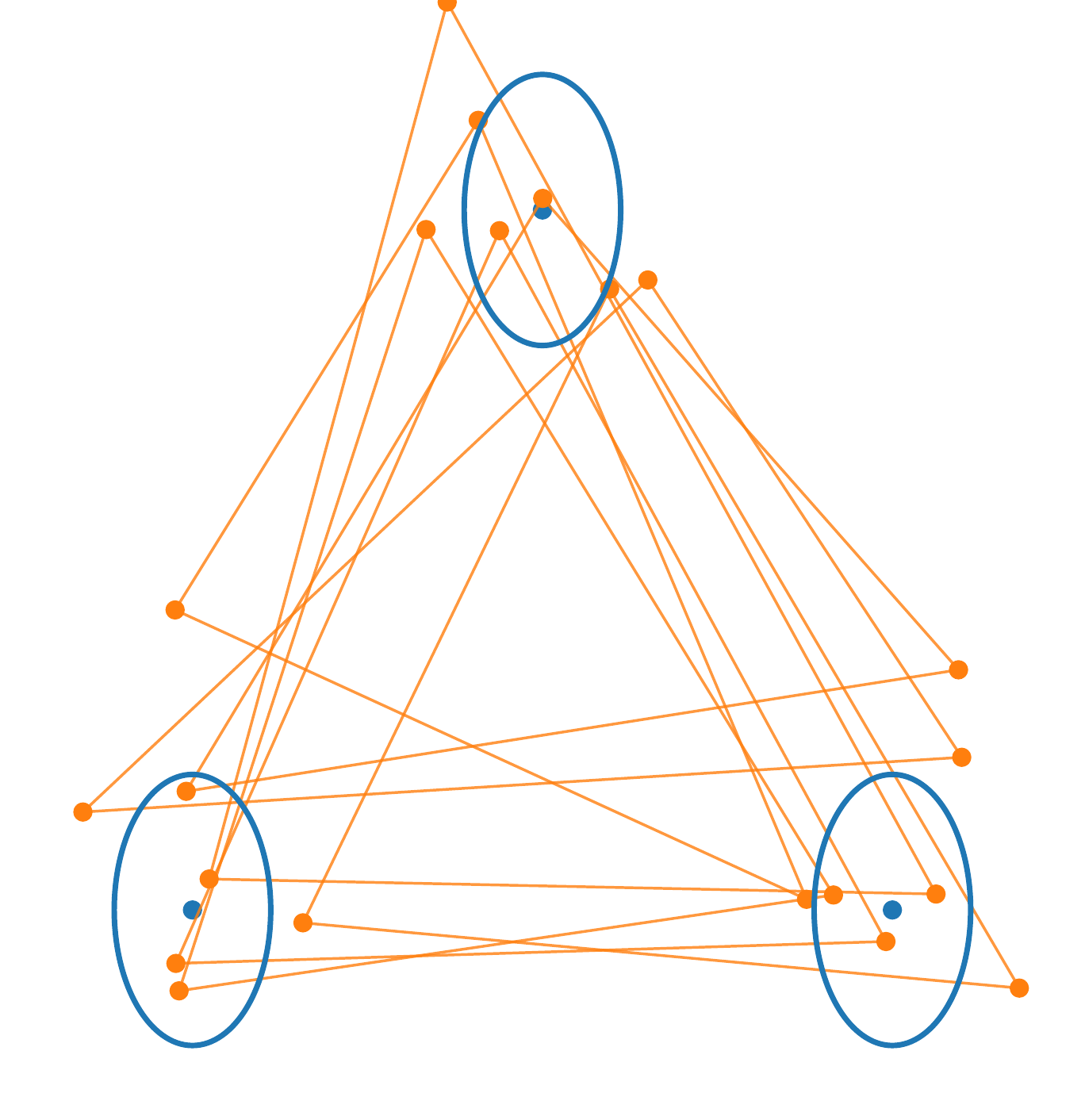}
  }
  \subfloat[$\rho=0.95$]{
    \includegraphics[width=0.22\textwidth]{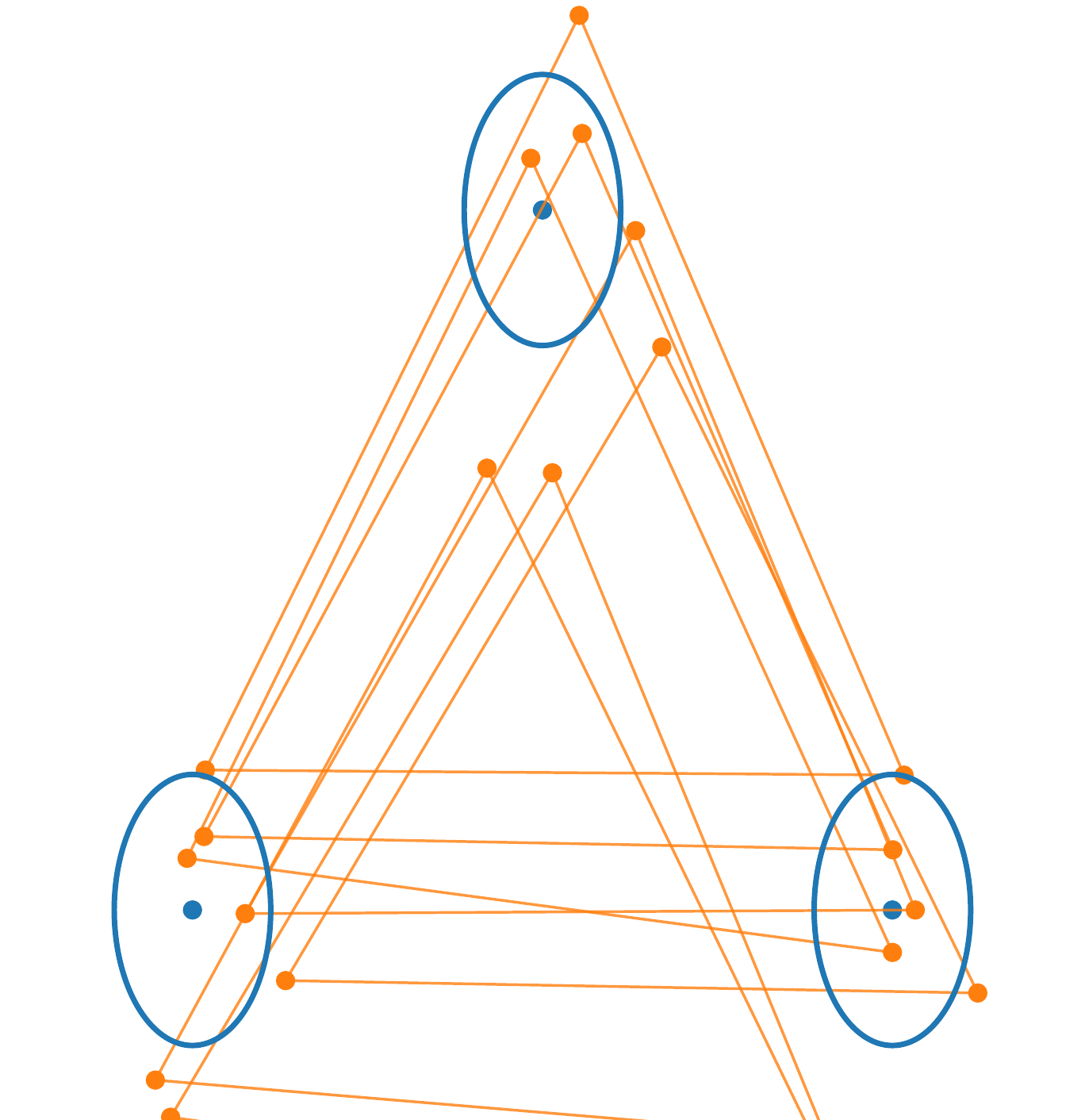}
  }
  \subfloat[$\rho=1$]{
    \includegraphics[width=0.22\textwidth]{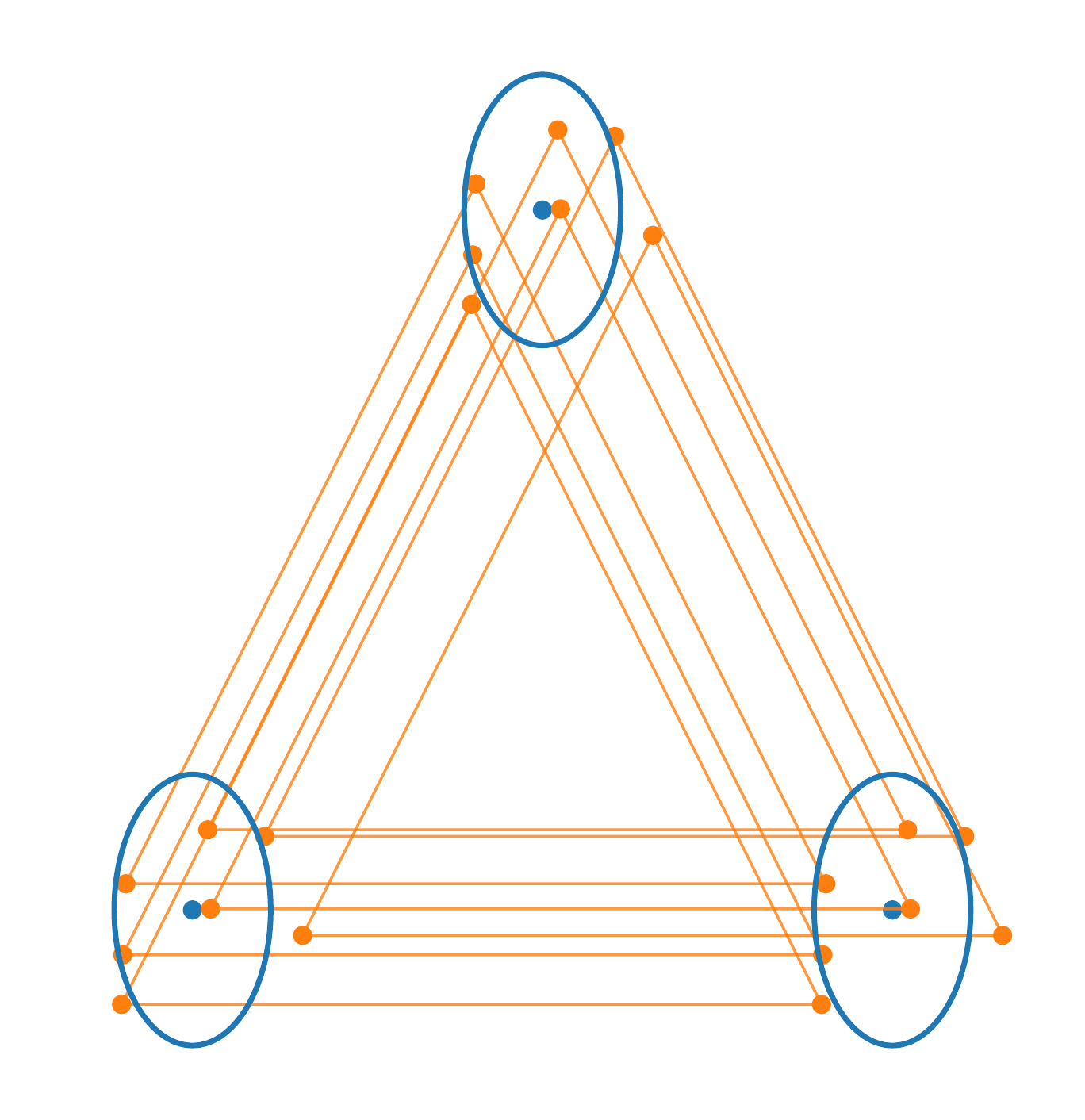}
  }
  \caption{The effect of the degree of linear correlation on the relative
    positions of three 2-D landmarks. The correlation is quantified using the
    Pearson correlation coefficient, $\rho$, for a synthetic Gaussian belief
    over the landmarks. Shown in blue are the marginal distributions of each
    landmark---the ellipses indicate the first standard deviations. Each orange
    triangulation of points represents a single 6-D sample from the joint
    distribution over all three landmarks.}
  \label{fig:lm_corr}
\end{figure*}

In order to decouple the process of dense mapping from the robot pose, we need
to remove the statistical dependency between $\mset{M}^{}$ and $\mvec{x}$.
Transforming $\mset{M}^{}$ to the relative \ac{IRF} achieves this through an
important characteristic of \ac{SLAM}: as clusters of neighbouring landmarks are
often observed from a single robot pose---or from successive robot poses---these
landmarks are highly correlated.\footnote{The degree of statistical dependency
  can be quantified using linear correlation---specifically, using the Pearson
  correlation coefficient, $\rho \in [-1, 1]$. In the Gaussian case, linear
  correlation equates to statistical dependency; that is, $\rho = 0$ equates to
  statistical independence, and $\rho = \pm 1$ equates to perfect linear
  dependence.} Therefore, although the marginal belief over each landmark in the
global \ac{IRF} can contain significant uncertainty, the relative positions
between neighbouring landmarks are known with little uncertainty. More
importantly, the landmark correlation increases monotonically with the number of
measurements \citep{Durrant-Whyte2006} and, in the limit, becomes perfectly
correlated for the linear Gaussian case \citep{Dissanayake2001}. In the case of
perfect correlation, there is no uncertainty between the relative landmark
positions. We illustrate the effect of various degrees of correlation on the
relative landmark positions for a synthetic Gaussian belief over three 2-D
landmarks (\Cref{fig:lm_corr}). For perfect linear correlation, the triangles
formed by each 6-D sample are identical in shape, whereas the shapes are
inconsistent if there is no correlation. In practice, only high correlations
between landmarks are achievable, although \citet{Nieto2006} show that the
relative uncertainty between landmarks is sufficiently low for relative
\acp{IRF} to be used in mapping.

Similarly, the robot pose will be highly correlated with nearby landmarks, and
consequently the relative positions of the robot pose and landmarks will also be
known accurately. In the ideal case, in which the robot and landmarks' positions in
the global \ac{IRF} are perfectly correlated, their relative positions would be
perfectly known. If we then describe $\mset{M}^{}$ with respect to the landmarks
(that is, transforming $\mset{M}^{} \rightarrow \mset{M}^R$), $\mset{M}^{R}$
would be statistically independent of both the landmarks and the robot pose.
This is because the relative positions of the robot and landmarks would be known
perfectly, and therefore the uncertainty in the belief over $\mset{M}^R$ would
only stem from measurement uncertainty. Although the relative positions are
never perfectly known, they are usually accurately known and it is reasonable to
approximate the belief over the \ac{SLAM} states and $\mset{M}^R$ as
statistically independent
\begin{equation}
  \label{eq:bel_m_rel}
  \belief(\mvec{x}^{},\mset{L}^{},\mset{M}^R)
  \approx
  \belief(\mvec{x}^{},\mset{L}^{})\belief(\mset{M}^R),
\end{equation}
where
\begin{equation}
  \label{eq:measrelindep}
  \belief(\mset{M}^R) \propto \prod_{i} p(\mvec{z}^R_i|\mvec{m}^R_i).
\end{equation}

\begin{figure*}[h!]
  \begin{minipage}{\textwidth}
    \renewcommand\thempfootnote{$\dagger$}
    \centering
    \subfloat[]{
      \includegraphics[height=7cm]{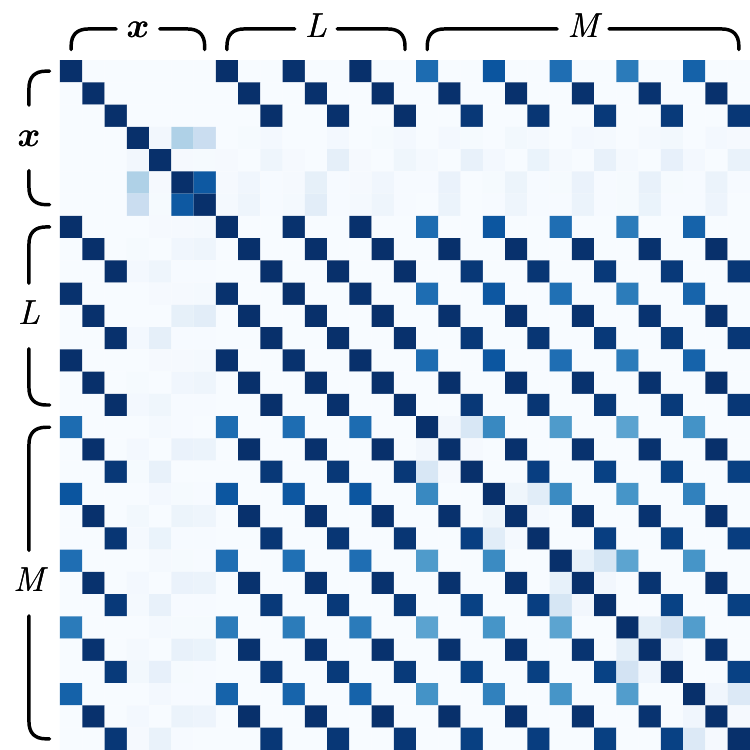}%
      \label{fig:corr_abs}
    }
    \subfloat[]{
      \includegraphics[height=7cm]{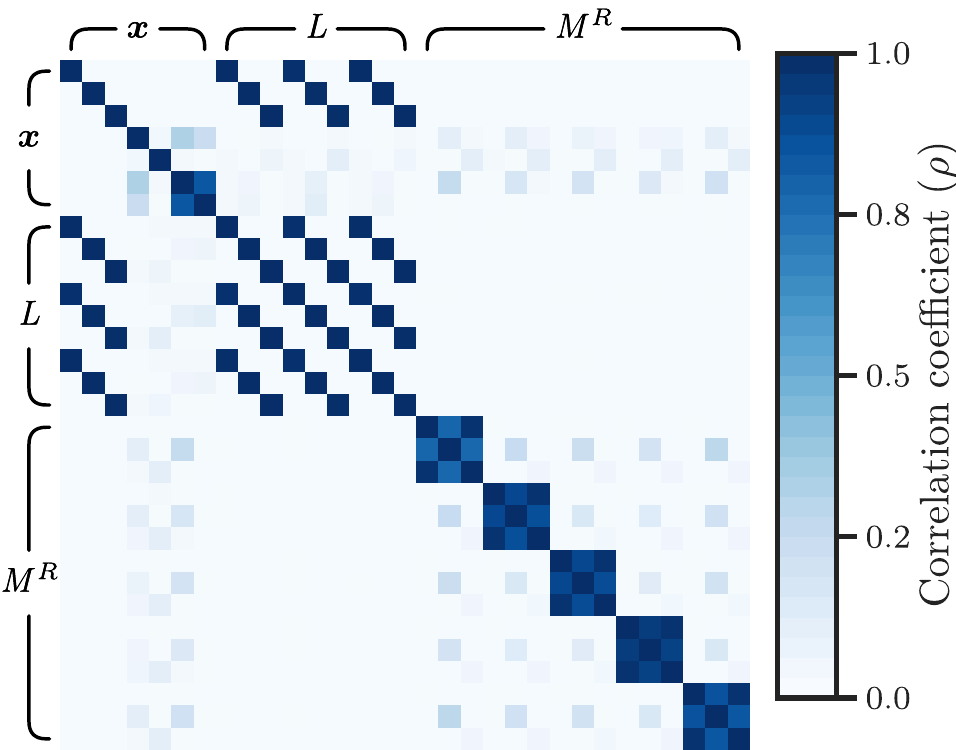}%
      \label{fig:corr_rel}
    }

    \subfloat[]{
      \includegraphics[width=0.55\textwidth]{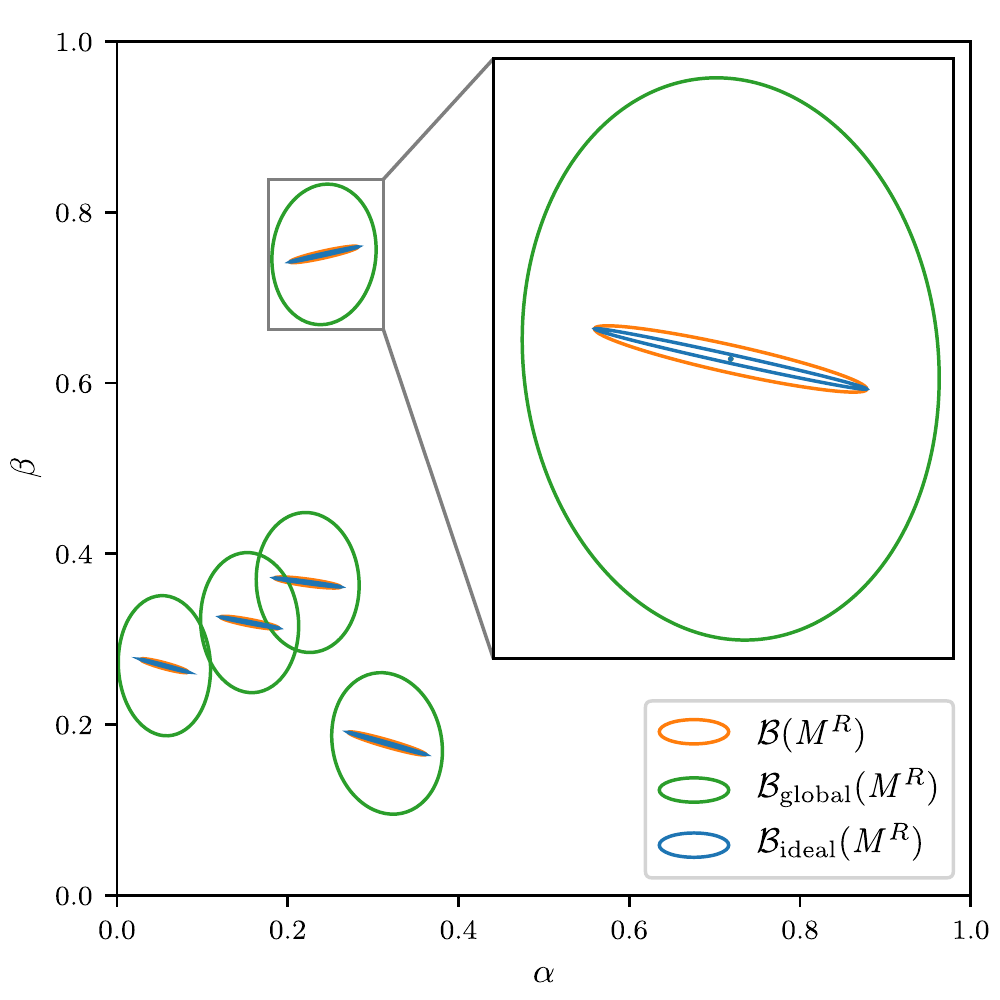}%
      \label{fig:corr_uncert}
    }
    \caption{A visualisation of the absolute value of the Pearson correlation
      coefficient matrix (the darker, the more correlated) for the Gaussian
      belief distributions\protect\footnote{We denote the current robot pose and
        landmark map in the global \ac{IRF}, $\mvec{x}^{}$ and $\mset{L}^{}$
        respectively, and the environment surface points in the global \ac{IRF},
        $\mset{M}^{}$, and the relative \ac{IRF}, $\mset{M}^R$.}
      $\belief(\mvec{x}^{},\mset{L}^{},\mset{M}^{})$ and (b)
      $\belief(\mvec{x}^{},\mset{L}^{},\mset{M}^R)$. In (b) the $3\times3$ block
      diagonals are from the individual 3-D environment surface points. (c) A
      comparison of the uncertainty in the beliefs over the environment surface
      measurements in a relative and global \acf{IRF}. Our approximate belief,
      $\belief(\mset{M}^R)$, is compared to $\belief_{\text{ideal}}(\mset{M}^R)$
      and $\belief_{\text{global}}(\mset{M}^R)$ (see the text for further
      details of the belief definitions). The ellipses outline the first
      standard deviations of the respective belief distributions. The data in
      this visualisation was taken from a practical stereo vision dataset.}
    \label{fig:corr_exp}
  \end{minipage}
\end{figure*}

To illustrate the validity of this approximation, we visualise the absolute
value of the Pearson correlation coefficient matrix for the Gaussian belief
distributions $\belief(\mvec{x}^{},\mset{L}^{},\mset{M}^{})$ and
$\belief(\mvec{x}^{},\mset{L}^{},\mset{M}^R)$ of a practical stereo vision
dataset (\Cref{fig:corr_abs,fig:corr_rel}). In
$\belief(\mvec{x}^{},\mset{L}^{},\mset{M}^{})$, we see that there is indeed a
high correlation between $\mvec{x}$, $\mset{L}$ and $\mset{M}$. Upon
transforming $\mset{M} \rightarrow \mset{M}^R$, the resulting belief
$\belief(\mvec{x}^{},\mset{L}^{},\mset{M}^R)$ has negligible correlations
between the \ac{SLAM} states and $\mset{M}^R$. The correlations are almost zero,
which equates to \ac{SLAM} states and $\mset{M}^R$ being approximately
statistically independent.\footnote{Jointly Gaussian-distributed variables are
  statistically independent if and only if they are uncorrelated.} Therefore,
this process of transforming to a relative \ac{IRF} decouples the process of
dense mapping from the robot pose.

The main advantage of this decoupling process is highlighted in the scenario in
which loop closure is performed. The belief over the environment surface points
in the relative \ac{IRF} is decoupled from the \ac{SLAM} states; hence, when the
belief over the \ac{SLAM} states changes due to loop closure, the environment
surface points no longer need to be reintegrated into the map---as would be
required in a global \ac{IRF}. Instead, the burden is placed on the \ac{SLAM}
algorithm to maintain (or allow the extraction of) marginal distributions over
the landmarks demarcating the submaps, which is a far more tractable approach.

In order to evaluate the efficacy of performing dense mapping in a relative
IRF---as opposed to a global \ac{IRF}---we compare the uncertainty in the
beliefs over the environment surface points in a relative and global \ac{IRF}.
For the case of mapping in a relative \ac{IRF}, we consider our approximate
belief $\belief(\mset{M}^R)$ (\Cref{eq:measrelindep}). We also consider the
ideal case, in which our statistical independence approximation is exact---that
is, the robot pose and landmarks are perfectly correlated or, in other words,
the relative positions between $\mvec{x}$ and $\mset{L}$ are perfectly known.
The uncertainty in this ideal belief, $\belief_{\text{ideal}}(\mset{M}^R)$, is
solely dependent on the measurement uncertainty. For the case of mapping in a
global \ac{IRF}, we consider the belief over the environment surface points in a
global \ac{IRF}, $\belief(\mset{M})$. To compare the beliefs in the relative
\ac{IRF} with that of the global \ac{IRF}, we need to evaluate all the beliefs
using the same units. We therefore deterministically transform
$\belief(\mset{M})$ to the relative \ac{IRF} using the maximum likelihood value
of the landmarks; we denote the resulting belief
$\belief_{\text{global}}(\mset{M}^R)$. The resulting comparison of all the
beliefs is shown in \Cref{fig:corr_uncert}. From this we can conclude that the
uncertainty in belief over the environment surface point used for dense mapping
can be significantly less when using a relative \ac{IRF} instead of a global
IRF. Additionally, the uncertainty in the belief becomes almost solely dependent
on the measurement uncertainty. \citet{Nieto2006} noted similar effects when
representing other landmarks in a relative \ac{IRF}; that is, a dramatic
decrease in the correlation between landmarks in the relative and global
\acp{IRF}, and a decrease in the marginal landmark belief uncertainty in the
relative \ac{IRF}.

In this section, we presented a method of decoupling the robot pose from the
process of dense mapping. By transforming the information from the measured
environment surface points to a relative \ac{IRF}, we have approximately removed
their statistical dependence on the robot pose. Next we look at incorporating
this information into our model of the environment.



\section{The Stochastic Triangular Mesh Map}
\label{sec:mapmodel}

\begin{figure*}[h!]
  \begin{minipage}{\textwidth}
    \renewcommand\thempfootnote{$\dagger$}
    \centering
    \subfloat[]{
      \includegraphics[width=0.49\textwidth]{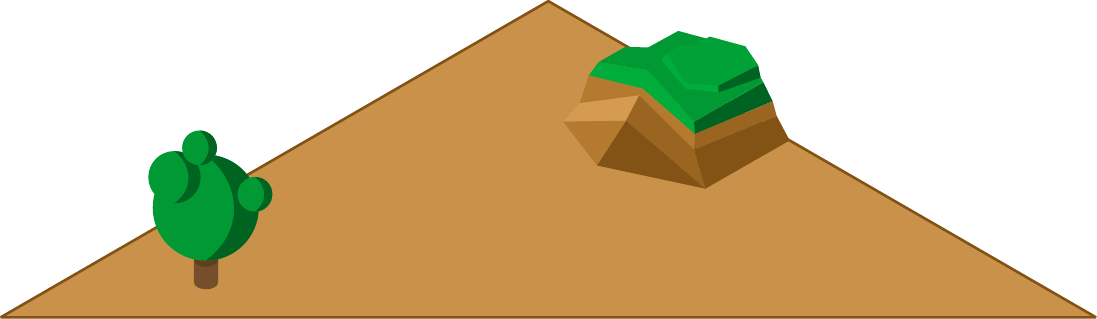}
    }
    \subfloat[]{
      \includegraphics[width=0.49\textwidth]{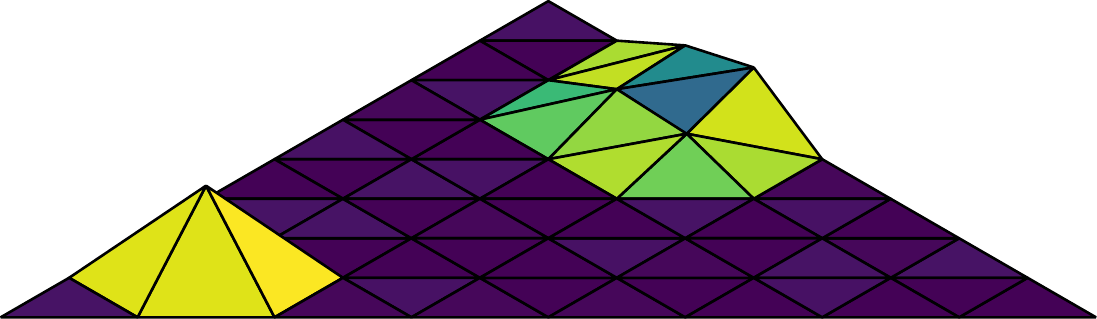}
    }
    \caption{(a) A hypothetical environment within a triangular
      submap\protect\footnote{The hill and tree icons were adapted from an
        original by sceneit from Vecteezy.com.}, and (b) a depiction of the
      corresponding \acf*{STM} map. Each triangular \acf*{surfel} in the mesh is
      coloured according to its planar deviation---the deviation of the actual
      surface of the environment from the triangular planes (the lighter, the
      higher the deviation).}
    \label{fig:example}
  \end{minipage}
\end{figure*}

Based on the highlighted drawbacks of the existing mapping techniques
(\Cref{sec:evaluation}), we propose a mapping technique that explicitly models
the surface of the environment using \iacf{STM}; that is, a triangular mesh
consisting of stochastic \acfp{surfel}, forming a continuous representation of
the surface of the environment (\Cref{fig:example}).

Most of the existing explicit surface-mapping techniques operate under the
approximation that the map \emph{can} model the surface of the environment
exactly. \Iac{STM} map, on the other hand, accounts for the fact that the
surface of the environment cannot be modelled exactly with deterministic map
elements; instead, the surface of the environment is treated as a
\emph{stochastic process}. Additionally, even if it is possible to represent the
environment exactly, the representation would be very wasteful; we rather
summarise the surface information relevant to the application. In contrast to
most mapping techniques, \iac{STM} map also captures the structure of the
environment by enforcing continuity in the model. Although the principle is
similar to Gaussian process mapping, we are able to update the model
incrementally and in a scalable fashion.

\begin{figure}[h!]
  \centering
  \includesvg[width=0.49\textwidth,pretex=\relscale{0.6}]{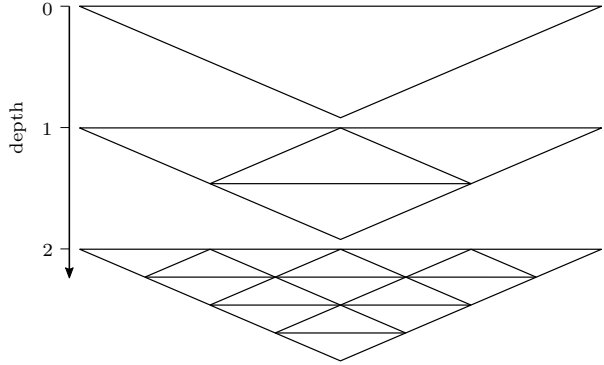}
  \caption{The recursive division of a triangular submap into equisized
    triangular grid elements.}
  \label{fig:subdiv}
\end{figure}

Following the aforementioned submapping method (\Cref{sec:irfs}), we develop the
\ac{STM} map within a triangular submap. We partition the submap into a regular
triangular grid by recursively subdividing the horizontal ($\alpha$-$\beta$)
plane into equisized triangular grid elements (\Cref{fig:subdiv}). The surface
of the environment within each grid element---the region normal to the grid
element---is modelled by the \ac{STM} map using a \ac{surfel}. The submap
division is performed until a desired grid element size is achieved. Depending
on the application, this could be simply chosen based on the dimensions and
physical capability of the robot. In general, it would be more sensible to
perform this partitioning in an adaptive manner, although we do not address this
in the current implementation, but will discuss this in future work (see
\Cref{sec:conclusion}).

In the remainder of this section, we discuss how each surfel in \iac{STM} map
models the surface of the environment. Subsequently, we discuss performing
incremental Bayesian inference on \iac{STM} map (\Cref{sec:model-inference}). It
should also be noted that we develop the \ac{STM} map within the relative
\ac{IRF} of a submap, and consequently, all variables are assumed to be defined
in the relative IRF. However, without loss of generality, this representation
could also be applied in a global IRF framework by defining the IRF accordingly,
which we demonstrate in \Cref{sec:global-irf-experiment}.

\subsection{Surface Element Model}
\label{sec:surfel-model}

\Iacf{STM} map consists of triangular \acfp{surfel}, which together form a
continuous representation of the surface of the environment within the submap.
We assume that the surface of the environment within the submap can be
represented adequately using a 2.5-D map. At any coordinate on the horizontal
plane, the environment normal to this plane need only be described using a
single height. Therefore, associated with each grid element is a triangular
surfel that models the surface of the environment within the grid element---the
region normal to the grid element.

\begin{figure*}[h!]
  \centering
  \includesvg[width=\textwidth]{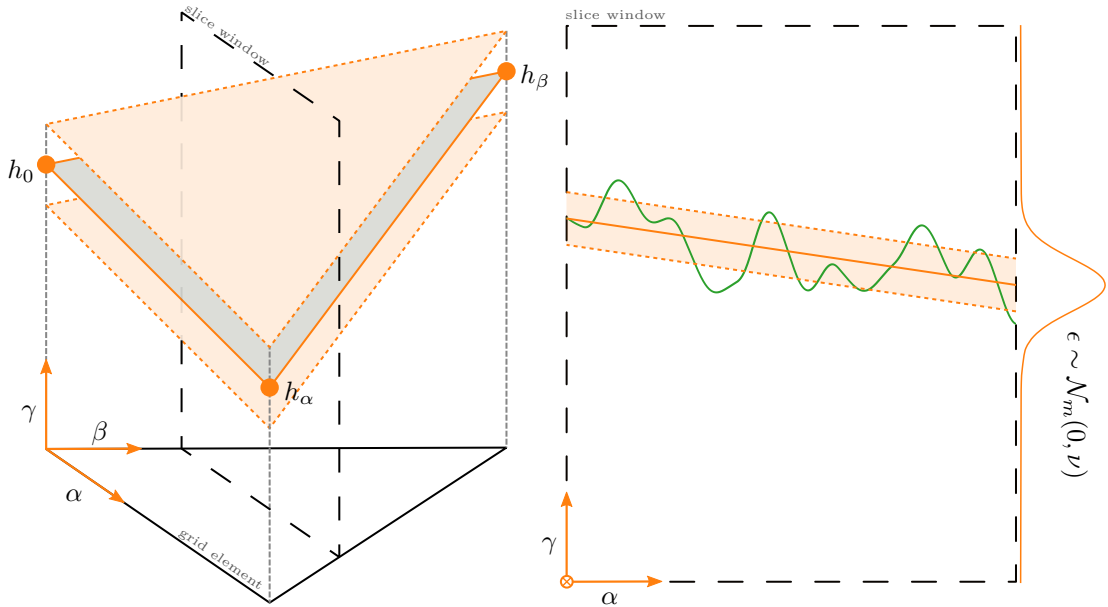}
  \caption{(left) The \acf*{surfel} model, consisting of a mean
    plane---parameterised by the vertex heights $\mvec{h} = \protect\bsmat{h_0,
      & h_\alpha, & h_\beta}^\T$---and a planar deviation $\nu$. (right) A 2-D
    slice of the surfel (orange) of an underlying surface (green). The shaded
    region indicates the one standard deviation confidence interval.
    $\GM(\cdot)$ denotes the moment form of a Gaussian distribution.}
  \label{fig:param_model}
\end{figure*}

If we initially consider a map with a single grid element, the maximum length
along both the $\alpha$- and $\beta$-axes within the grid element will be
unity---\Cref{eq:constraints}. The associated surfel models the surface of the
environment within the grid element as a stochastic process---a mean plane and a
homoscedastic\footnote{A stochastic process is \emph{homoscedastic} when the
  process variance is constant and finite.} deviation from it, normal to the
grid element (\Cref{fig:param_model}). This simple model does not try to exactly
represent the actual surface of the environment, but summarises the key aspects
of the surface instead. This is an efficient representation for the purpose of
robotic navigation, since the model captures the relevant aspects of the
environment, and avoids modelling unnecessary details that are not relevant for
the intended application. We consider this to be an advantage rather than a
disadvantage. More precisely, a point on the surface of the environment at some
given coordinates $(\alpha,\beta)$ is modelled by the stochastic process
\begin{equation}
  \label{eq:surffunc}
  \begin{split}
    \gamma
    &= f(\alpha, \beta, \mvec{h}) + \epsilon\\
    &=
    \begin{bmatrix}
      1 - \alpha - \beta & \alpha & \beta
    \end{bmatrix}
    \mvec{h} + \epsilon,
  \end{split}
\end{equation}
where $f(\alpha, \beta, \mvec{h})$ describes a point on the mean plane, and
$\epsilon \sim \GM(\epsilon; 0, \nu)$\footnote{For a Gaussian distribution, we
  denote the canonical form $\GC(\cdot)$, and the moment form $\GM(\cdot)$.}
models the deviation of the actual surface from the mean plane. We refer to the
variance of the stochastic process, $\nu$, as the planar deviation. The mean
plane is specified using the normal heights at each of the vertices of the grid
element
\begin{equation}
  \mvec{h} =
  \begin{bmatrix}[c]
    h_0     \\
    h_\alpha \\
    h_\beta
  \end{bmatrix},
\end{equation}
where $h_0$, $h_\alpha$, and $h_\beta$ are the heights at $(\alpha,\beta) =
(0,0)$, $(1,0)$ and $(0,1)$ respectively. Consequently, the surfel model is
parameterised by $\mvec{h}$ and $\nu$, which we combine into a single parameter
vector, $\mvec{\theta} = \bmat{\mvec{h} & \nu}^\T$. This simple stochastic
process allows us to account for a wide range of surfaces by essentially
summarising the underlying environment within the grid element. It should be
noted that we follow a Bayesian modelling approach, treating the map parameters
as random variables. We specifically model the surfels' heights as Gaussian
distributed, and the planar deviation as inverse-gamma distributed. We expand on
this when discussing performing inference on the model
(\Cref{sec:model-inference}).

In order to generalise this definition of the surfel model from a single to
multiple grid elements within a submap, we normalise each grid element---scaling
the maximum length of the $\alpha$- and $\beta$-axes to unity, and shifting to
the origin. This normalisation does not affect the values of the surfel
parameters, $\mvec{\theta}$, which are defined with respect to the
$\gamma$-axis. Additionally, as this is a deterministic linear operation, the
Gaussian measurement distributions in this normalised grid element remain
exactly Gaussian distributed, which will be important when performing inference
on the surfel model parameters (\Cref{sec:model-inference}).

\begin{figure*}[h!]
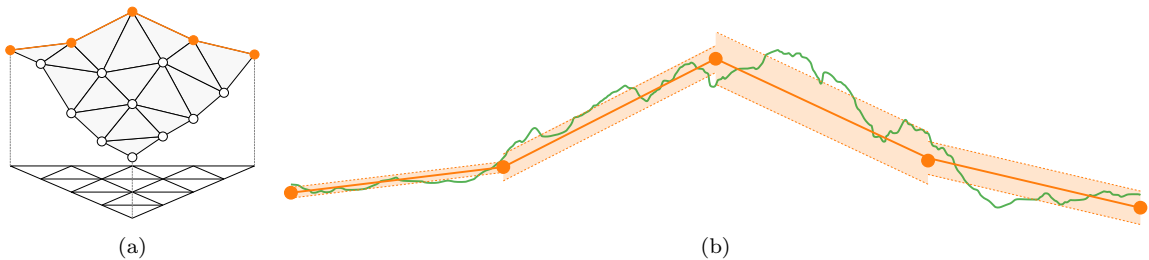

  \centering
  \subfloat[]{
    \includesvg[width=0.22\textwidth]{map_mean_mesh}
  }
  \subfloat[]{
    \includesvg[width=0.71\textwidth]{continuous_model_slice}
  }
  \caption{\Iacf{STM} forms a continuous representation of the surface of the
    environment. We visualise (a) the mean mesh, and (b) a slice through the STM
    map (orange) of an underlying surface (green). The shaded region indicates
    the one standard deviation confidence interval in the respective surfels.}
  \label{fig:cont_model}
\end{figure*}

To create a continuous representation, the mean planes of surfels in contiguous
grid elements are constrained to be continuous; that is, the heights at their
shared vertices should be equal (\Cref{fig:cont_model}). However, we do not
enforce any constraints on the planar deviations between surfels. The continuous
surface in a submap formed by the mean plane in each surfel describes the mean
mesh, which is parameterised by the ordered set of unique vertex heights,
$\mset{H}$. The stochastic deviation from this mean mesh is parameterised by the
ordered set of planar deviations in each surfel, $\mset{V}$. Both $\mset{H}$ and
$\mset{V}$ fully parameterise \iac{STM} map, which can be combined into a single
ordered set, $\mset{\Theta}$.

We have defined how the constituent surfels of \iac{STM} map model the
environment; we now seek to determine the parameter values for \iac{STM} map
based on measurements of the surface of the environment---in other words, to
perform inference.



\section{Model Inference}
\label{sec:model-inference}

Based on the definition of \iac{STM} map, we now wish to infer the values of the
map parameters, $\mset{\Theta}$, for some given noisy measurements of the
surface of the environment, $\mset{Z}=\given{Z}$. We follow a Bayesian modelling
approach, maintaining a belief distribution\footnote{A belief distribution is
  the posterior distribution over some states given all the relevant evidence.}
over the map parameters. We would therefore ideally like to calculate the joint
belief distribution over all the map parameters
\begin{equation}
  \belief(\mset{\Theta})
  =
  p(\mset{\Theta}|\mset{Z}=\given{Z}).
\end{equation}
However, for large maps, this would quickly become intractable to store, as the
storage grows quadratically with the number of surfels in the map (for our
choice of continuous parametric distributions). To circumvent this issue, we
instead directly calculate the marginal belief for each surfel's parameters
\begin{equation}
  \belief(\mvec{\theta})
  =
  p(\mvec{\theta}|\mset{Z}=\given{Z}),
\end{equation}
for which storage grows linearly with each added surfel.

Next we will look at calculating each surfel's belief independently of the rest
of the map (\Cref{sec:surfel-inference}); this is equivalent to neglecting the
continuity constraints between surfels. Building upon this, we expand to
performing inference on the full model (\Cref{sec:full-model-inference}).
Finally, we summarise the inference algorithm for \iac{STM} map, and discuss the
implementation thereof (\Cref{sec:algorithm}).

\subsection{Single Surfel Inference}
\label{sec:surfel-inference}

We initially consider each surfel in isolation, and wish to calculate the belief
distribution over a single surfel's parameters, $\mvec{\theta} = \bmat{\mvec{h}
  & \nu}^\T$, from noisy measurements associated with the respective grid
element\footnote{We associate a measurement with a grid element based solely on
  whether the measurement's mean lies within the grid element.}. We combine this
approach to handle multiple surfels in \Cref{sec:full-model-inference}, but it
is useful to first consider this perspective before performing inference on the
full model.

\begin{figure*}[h!]
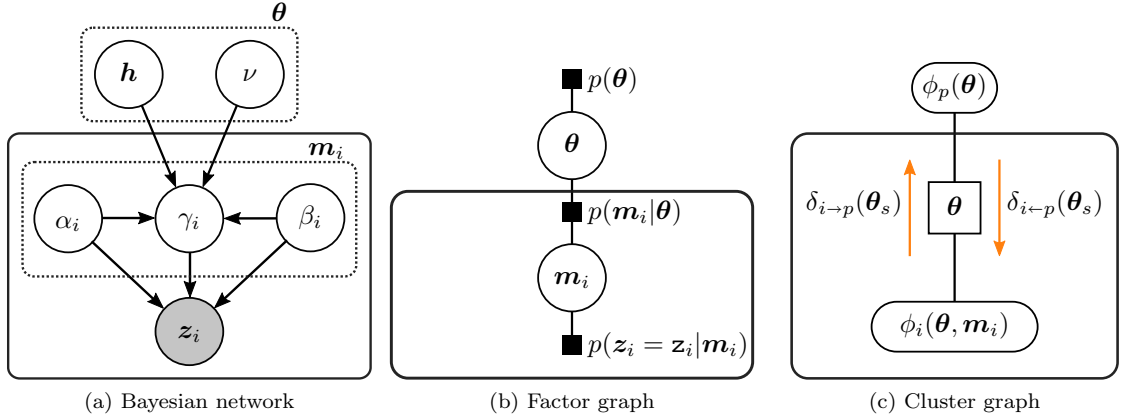

  \centering
  \subfloat[Bayesian network]{
  \includesvg[width=0.3\textwidth]{surfel_bayes_plate}
  \label{fig:bayes_net}
  }
  \subfloat[Factor graph]{
  \includesvg[width=0.3\textwidth]{surfel_fg_plate}
  \label{fig:fg}
  }
  \subfloat[Cluster graph]{
  \includesvg[width=0.3\textwidth]{surfel_cg_plate}
  \label{fig:cg}
  }
  \caption{Different types of probabilistic graphical models representing the
    inference problem for our surfel model. The surfel model parameters,
    $\mvec{\theta} = \protect\bsmat{\mvec{h} & \nu}^\T$, measurements,
    $\mvec{z}_i$, and corresponding points on the surface of the environment,
    $\mvec{m}_i = \protect\bsmat{\alpha_i & \beta_i & \gamma_i}^\T$, are all
    random variables. We use plate notation, where the solid rectangular plates
    indicate a repetition of the subgraph therein. (a) The dotted rectangle
    denotes an expanded vector, and the shaded circle indicates an observed (or
    given) variable. (b) Factors are indicated with $\blacksquare$. (c) Cluster
    potentials and messages are denoted $\cp(\cdot)$ and $\delta(\cdot)$
    respectively, and the sepset between clusters is indicated with $\boxdot$.}
\end{figure*}

Most probabilistic problems involve several random variables;
\acp{PGM}\footnote{For a comprehensive guide to PGMs we refer the reader to
  \citet{Koller2009}.} provide a compact graphical representation over this
complex high-dimensional space, which is useful in both modelling and performing
inference. There are many different PGMs for different types of problems (or
questions); in our case we focus on three types of PGMs: Bayesian networks,
factor graphs, and cluster graphs. A \emph{Bayesian network}, or \emph{belief
  network}, represents the conditional independence assumptions in the joint
distribution as a directed graph between random variable nodes. The Bayesian network representation of the surfel model we use is shown in
\Cref{fig:bayes_net}, which equivalently represents the factorisation over the
joint distribution
\begin{equation}
  \label{eq:joint_bayes}
  \begin{split}
    &p(\mvec{\theta}, \mset{M}, \mset{Z}=\given{Z})
    = p(\mvec{\theta})
    \underbrace
    {
      \prod_i
      p(\mvec{z}_i=\given{z}_i| \mvec{m}_i) \,
      \overbrace
      {
        p(\gamma_i | \alpha_i, \beta_i, \mvec{\theta}) \,
        p(\alpha_i) \,
        p(\beta_i)
      }^{p(\mvec{m}_i | \mvec{\theta})}
    }_{p(\mset{M}, \mset{Z}=\given{Z} | \mvec{\theta})}
    ,
  \end{split}
\end{equation}
where $\mset{Z} = (\mvec{z}_i)$ is the sequence of noisy measurements
corresponding to the points on the surface of the environment, $\mset{M} =
(\mvec{m}_i)$, as mentioned in \Cref{sec:decouple}.
This factorisation of the joint distribution models the generative process of
measurements---the model parameters describe where actual points on the surface
in the environment lie, and these points are observed through measurements of
the environment. In the measurement mode,
$p(\mvec{z}_i=\given{z}_i|\mvec{m}_i)$, we assume that a measurement,
$\mvec{z}_i$, is only dependent on the corresponding point on the actual surface
of the environment, $\mvec{m}_i$; this measurement distribution was derived in
\Cref{sec:decouple}. In $p(\gamma_i|\alpha_i,\beta_i,\mvec{\theta})$, we assume
that the height of the surface point, $\gamma_i$, is only dependent on the grid
element coordinates $(\alpha_i, \beta_i)$ and the surfel parameters,
$\mvec{\theta}$. This is based on the definition of the stochastic process in
the surfel model---\Cref{eq:surffunc}.

From this Bayesian network we can construct a \emph{factor graph}
(\Cref{fig:fg}). It represents the fully factorised functional form of the joint
distribution in an undirected graph, where the random variables are linked by
the factors they share. These factors can be grouped into clusters, resulting in
a \emph{cluster graph} representation (\Cref{fig:cg}). The combined factors in a
cluster are known as the cluster potential, and shared links between clusters
are defined by their sepset---a subset of the shared variables. Although there
is a unique factor graph for a Bayesian network, the grouping of factors in a
cluster graph is, in general, non-unique. We choose to group all the factors
inside the plate into a cluster, and the factor outside the plate into a
cluster. Both factor and cluster graphs can be used for inference, although we
will focus on inference in cluster graphs.

As a cluster graph is the result of grouping factors from the joint
distribution, we can equivalently represent the joint distribution as the
product over all the cluster potentials
\begin{equation}
  \label{eq:joint_cluster}
  p(\mvec{\theta}, \mset{M}, \mset{Z}=\given{Z})
  = \cp_p(\mvec{\theta})
  \prod_i
  \cp_i(\mvec{\theta}, \mvec{m}_i),
\end{equation}
where we refer to $\cp_p(\mvec{\theta})$ as the \emph{prior cluster potential}
and each $\cp_i(\mvec{\theta}, \mvec{m}_i)$ as a \emph{likelihood cluster
  potential}. We wish to infer the surfel belief distribution for some given
measurements, $\mset{Z}=\given{Z}$. Using Bayes' theorem, we can express this
belief in terms of the joint distribution
\begin{equation}
  \label{eq:bel_exact}
  \begin{split}
    \belief(\mvec{\theta})
    &= p(\mvec{\theta}|\mset{Z}=\given{Z})\\
    &\propto \int\displaylimits_{\mset{M}}
    p(\mvec{\theta},\mset{M}, \mset{Z}=\given{Z}) \, \text{d}\mset{M}\\
    &\propto \cp_p(\mvec{\theta})
    \prod_i \int\displaylimits_{\mvec{m}_i}
    \cp_i(\mvec{\theta},\mvec{m}_i)
    \, \text{d}\mvec{m}_i
    .
  \end{split}
\end{equation}
An equivalent method of calculating this belief is by using message passing. For
inference in cluster graphs, the outgoing and incoming messages for each cluster
needs to be calculated. According to the integral-product algorithm, the
outgoing message from the $i$-th likelihood cluster is defined as
\begin{equation}
 \label{eq:message_out}
 \outmsg{i}{p}(\mvec{\theta})
 =
 \int\displaylimits_{\mvec{m}_i}
 \cp_i(\mvec{\theta},\mvec{m}_i)
 \, \text{d}\mvec{m}_i.
\end{equation}
Similarly, the incoming message to the $i$-th likelihood cluster is defined as
\begin{equation}
 \label{eq:message_in}
 \begin{split}
   \inmsg{i}{p}(\mvec{\theta})
   &=
   \cp_p(\mvec{\theta}) \prod_{j \neq i}
   \outmsg{j}{p}(\mvec{\theta}).
 \end{split}
\end{equation}
By comparing \Cref{eq:bel_exact} with \Cref{eq:message_out,eq:message_in}, we can
see that the product of these messages equivalently specifies the surfel belief
\begin{equation}
  \label{eq:bel_messages}
  \belief(\mvec{\theta})
  \propto
  \outmsg{i}{p}(\mvec{\theta}) \inmsg{i}{p}(\mvec{\theta}).
\end{equation}
However, there is no convenient closed-form solution to either of the
messages---and consequently the surfel belief. This is due to nonlinear
relationships between the random variables in the likelihood cluster potentials.
In order to find a tractable solution to the surfel belief, we have to perform
approximate inference. We specifically focus on variational
inference\footnote{For a review on the principles of variational inference, see
  \citet{Blei2017}}, which projects a desired (intractable) probability
distribution onto an approximating family of distributions. This projection is
performed by minimising a chosen information divergence---a measure of
similarity---between the exact and approximate distributions. A commonly-used
information divergence is the \ac{KL} divergence
\begin{equation}
  \KL{q(x)}{p(x)} \triangleq \int\displaylimits_{x} q(x) \log\frac{q(x)}{p(x)} \,dx,
\end{equation}
which is a positive measure of relative entropy between two probability
distributions $p(x)$ and $q(x)$\footnote{The \acf{KL} divergence is
  asymmetric---$\KL{q(x)}{p(x)} \neq \KL{p(x)}{q(x)}$. If $p(x)$ is exact and
  $q(x)$ an approximation then $\KL{q(x)}{p(x)}$ is referred to as the exclusive
  KL divergence. The reverse of this, $\KL{p(x)}{q(x)}$, is referred to as the
  inclusive KL divergence.}. We minimise the exclusive \ac{KL} divergence to
calculate an approximate surfel belief
\begin{equation}
  \approxBel(\mvec{\theta})
  =
  \argmin_{ \approxBel(\mvec{\theta}) \in \mathcal{F}}
  \KL{\approxBel(\mvec{\theta})}{\belief(\mvec{\theta})},
\end{equation}
where $\mathcal{F}$ is a family of approximating distributions. This process of
minimising the exclusive KL divergence is commonly referred to as the
\emph{information projection}; alternatively minimising the inclusive KL
divergence is referred to as the moment projection. As opposed to minimising
this global divergence, a group of techniques known as \emph{approximate message
  passing}\footnote{For a review on message passing techniques, see
  \citet{Minka2005}.} distributes the problem of variational inference by
instead incrementally performing local projections with the aim of approximately
minimising the global divergence. In essence, message passing minimises the
global divergence using coordinate descent. This can be explained intuitively in
the context of performing inference in a cluster graph. Message passing
iteratively approximates each cluster potential; at each iteration, a certain
cluster potential is approximated, while keeping all other cluster potentials
constant. The constant cluster potentials determine the incoming message of the
cluster potential being approximated, and are used as context to update the
approximation (how the approximation is updated depends on the chosen divergence
measure). This process is repeated until convergence, upon which the global
divergence is assumed to be at a (local) minimum.

Performing message passing for our chosen divergence measure is known as
\ac{VMP}~\citep{Winn2005}. If we were to instead minimise the inclusive KL
divergence (moment projection), this would result in the \ac{EP} message passing
technique \citep{Minka2001}. A notable property of VMP is that a fixed point in
performing message passing is also a stationary point in the global
divergence---in other words, minimising local divergences is equivalent to
minimising the global divergence. VMP is also the only message passing technique
with this property, whereas other message passing techniques only approximate it
\citep{Minka2005}; this is as a result of VMP using the exclusive KL divergence.

In order to calculate the approximate surfel belief,
\begin{equation}
  \label{eq:approx_bel}
  \approxBel(\mvec{\theta})
  \propto
  \approxoutmsg{i}{p}(\mvec{\theta})
  \approxinmsg{i}{p}(\mvec{\theta})
  ,
\end{equation}
the approximate messages, $\approxoutmsg{i}{p}(\mvec{\theta})$ and
$\approxinmsg{i}{p}(\mvec{\theta})$, are calculated by replacing
$\cp_i(\mvec{\theta}, \mvec{m}_i)$ by an approximating likelihood cluster
potential, $\approxcp_i(\mvec{\theta}, \mvec{m}_i)$, in
\Cref{eq:message_in,eq:message_out}. To calculate each approximate likelihood
cluster potential, VMP performs a local information projection
\begin{equation}
  \label{eq:kl_local}
  \approxcp_i(\mvec{\theta},\mvec{m}_i)
  =
  \argmin_{\approxcp_i \in \mathcal{F}}
  \KL
  {\approxcp_i \approxinmsg{i}{p}}
  {\cp_i \approxinmsg{i}{p}}
  ,
\end{equation}
where $\mathcal{F}$ is a family of approximating distributions (we omit the
arguments of the functions to declutter the notation). It should be noted that
this is an iterative process, as we must perform this local projection for each
likelihood cluster potential. To make this local projection tractable, we also
make the simplifying assumption that each approximating likelihood cluster
potential is of the factorised form\footnote{From this point on, for notational
  brevity we loosely use the arguments to functions as unique identifiers,
  namely $f(x) \triangleq f_x(x)$ and $f(y) \not\triangleq f(x)$.}
\begin{equation}
  \label{eq:mean_field}
  \approxcp_i(\mvec{\theta}, \mvec{m}_i) =
  \approxcp_i(\mvec{h}, \mvec{m}_i)
  \approxcp_i(\nu).
\end{equation}
This factorisation over disjoint sets of the random variables is a commonly-used
approximation referred as the structured mean-field approximation
\citep{Saul1996}. We choose this structured factorisation in particular, as it
more accurately represents the exact distribution compared to a full
factorisation. In contrast to other structured factorisation combinations, it
groups variables from the same distribution family (as we will soon show). We
also assume the prior cluster potential to be of a similar factorised form,
\begin{equation}
  \cp_p(\mvec{\theta})=\cp_p(\mvec{h})\cp_p(\nu),
\end{equation}
therefore the approximate messages and surfel belief will be similarly
factorised.

In order to perform the minimisation in \Cref{eq:kl_local}, we first need to
specify the family of approximating distributions for the factors of
$\approxcp_i(\mvec{\theta}, \mvec{m}_i)$. For the factor over $\mvec{h}$ and
$\mvec{m}_i$, we choose the Gaussian distribution
\begin{equation}
  \approxcp_i(\mvec{h},\mvec{m}_i)
  =
  \GC(\bmat{\mvec{h}&\mvec{m}_i}^\T;
  \mvec{\xi}_{i}, \mmat{\Omega}_{i}),
\end{equation}
where $\mvec{\xi}_{i}$ is the information vector and $\mmat{\Omega}_{i}$ is the
information matrix of the Gaussian distribution in canonical form. For the
factor over $\nu$, we choose the inverse-gamma distribution
\begin{equation}
  \approxcp_i(\nu)
  =
  \Gamma^{-1}(\nu; a_i, b_i)
  ,
\end{equation}
where $a_i$ is the shape parameter and $b_i$ the scale parameter of the
inverse-gamma distribution. The resulting product of these
distributions---\Cref{eq:mean_field}---should not be confused with a
normal-inverse-gamma distribution. In our case, this is simply the product of
these two distributions without any conditional dependencies.

If we had already calculated the approximate likelihood cluster potential,
$\approxcp_i(\mvec{\theta}, \mvec{m}_i)$, we could calculate the outgoing
message using \Cref{eq:message_out}, which would result in the distribution
\begin{equation}
  \label{eq:dist_msg_out}
  \approxoutmsg{i}{p}(\mvec{\theta})
  =
  \underbrace{
    \GC(\mvec{h};
    \mvec{\xi}\outarrow{i}{p}, \mmat{\Omega}\outarrow{i}{p})}
  _{\approxoutmsg{i}{p}(\mvec{h})}
  \underbrace{
    \Gamma^{-1}(\nu; a\outarrow{i}{p}-1, b\outarrow{i}{p})}
  _{\approxoutmsg{i}{p}(\nu)}.
\end{equation}
We assume that the prior cluster potential is also similarly distributed, as
\begin{equation}
  \label{eq:prior}
  \begin{split}
    \cp_p(\mvec{\theta})
    &=
    \underbrace{\GC(\mvec{h}; \mvec{\xi}_p, \mmat{\Omega}_p)}
    _{\cp_p(\mvec{h})}
    \underbrace{\Gamma^{-1}(\nu; a_p, b_p)}
    _{\cp_p(\nu)}
    .
  \end{split}
\end{equation}

Both the Gaussian and inverse-gamma families of distributions are part of the
exponential family \citep{Bishop2006}. These have the convenient property that
the product and division of distributions in the family also results in a
distribution in the family \citep{Minka2005} (up to a proportionality constant),
which amounts to the addition and subtraction of their natural parameters
respectively. We therefore can calculate the approximate incoming message to a
cluster, which will be distributed similarly to its constituents,
\begin{equation}
  \approxinmsg{i}{p}(\mvec{\theta})
  =
  \underbrace{
    \GC(\mvec{h};
    \mvec{\xi}\inarrow{i}{p}, \mmat{\Omega}\inarrow{i}{p})}
  _{\approxinmsg{i}{p}(\mvec{h})}
  \underbrace{
    \Gamma^{-1}(\nu; a\inarrow{i}{p}, b\inarrow{i}{p})}
  _{\approxinmsg{i}{p}(\nu)}.
\end{equation}
According to \Cref{eq:message_in}, the natural parameters of
$\approxinmsg{i}{p}(\mvec{h})$ are calculated as
\begin{equation}
  \mmat{\Omega}\inarrow{i}{p} =
  \mmat{\Omega}_p^{}
  +
  \sum_{j \neq i} \mmat{\Omega}\outarrow{j}{p}
  \; \text{ and } \;
  \mmat{\xi}\inarrow{i}{p} =
  \mmat{\xi}_p^{}
  +
  \sum_{j \neq i} \mmat{\xi}\outarrow{j}{p},
\end{equation}
and the natural parameters of $\approxinmsg{i}{p}(\nu)$ are calculated as
\begin{equation}
  a\inarrow{i}{p} =
  a_p^{}
  +
  \sum_{j \neq i} a\outarrow{j}{p}
  \; \text{ and } \;
  b\inarrow{i}{p} =
  b_p^{}
  +
  \sum_{j \neq i} b\outarrow{j}{p}.
\end{equation}
The same is true for the approximate surfel belief,
\begin{equation}
  \label{eq:approx_bel_form}
  \begin{split}
    \approxBel(\mvec{\theta})
    &= \underbrace{\GC(\mvec{h}; \mvec{\xi}_h, \mmat{\Omega}_h)}
    _{\approxBel(\mvec{h})}
    \underbrace{\Gamma^{-1}(\nu; a_\nu, b_\nu)}
    _{\approxBel(\nu)}.
  \end{split}
\end{equation}
According to \Cref{eq:approx_bel}, the natural parameters of
$\approxBel(\mvec{h})$ are calculated as
\begin{equation}
  \mmat{\Omega}_h =
  \mmat{\Omega}\outarrow{i}{p}
  +
  \mmat{\Omega}\inarrow{i}{p}
  \; \text{ and } \;
  \mmat{\xi}_h =
  \mmat{\xi}\outarrow{i}{p}
  +
  \mmat{\xi}\inarrow{i}{p},
\end{equation}
and the natural parameters of $\approxBel(\nu)$ are calculated as
\begin{equation}
  a_\nu =
  a\outarrow{i}{p}
  +
  a\inarrow{i}{p}
  \; \text{ and } \;
  b_\nu =
  b\outarrow{i}{p}
  +
  b\inarrow{i}{p}.
\end{equation}

Performing VMP with the structured mean-field approximation therefore allows us
to perform tractable inference to obtain an approximation of the surfel belief.
Up to this point, we have assumed that we have the approximate likelihood
cluster potentials, $\approxcp_i(\mvec{\theta}, \mvec{m}_i)$. We now discuss
exactly how to calculate each of the factors of approximate likelihood cluster
potentials and, consequently, the outgoing message,
$\approxoutmsg{i}{p}(\mvec{\theta})$.

\subsubsection{Updating \texorpdfstring{$\approxcp_i(\mvec{h},
    \mvec{m}_i)$}{Mean Plane Factor}}
\label{sec:update_h_m}

The iterative update of $\approxcp_i(\mvec{h}, \mvec{m}_i)$ that would perform
the local information projection of \Cref{eq:kl_local} can be calculated
according to the VMP algorithm \citep{Winn2005} as
\begin{equation}
  \label{eq:approx_h_cluster_pot}
  \begin{split}
    &\log \approxcp_i(\mvec{h},\mvec{m}_i)
    + \log
    \approxinmsg{i}{p}(\mvec{h})
    \\
    &\propto \left\langle \log \cp_i(\mvec{\theta},\mvec{m}_i,
      \mvec{z}_i=\given{z}_i) \approxinmsg{i}{p}(\mvec{\theta})
    \right\rangle_{\approxBel(\nu)},
  \end{split}
\end{equation}
where $\langle {\cdot} \rangle$ denotes the expectation operation. In order to
arrive at a solution to $\approxcp_i(\mvec{h},\mvec{m}_i)$, it is convenient to
first describe the process of calculating the approximate belief,
\begin{equation}
  \label{eq:bel_h_m_defn}
  \approxBel(\mvec{h}, \mvec{m}_i)
  \propto
  \approxcp_i(\mvec{h}, \mvec{m}_i)
  \approxinmsg{i}{p}(\mvec{h})
  ,
\end{equation}
and then to divide out the message $\approxinmsg{i}{p}(\mvec{h})$ to recover the
desired approximate to the cluster potential $\approxcp_i(\mvec{h},
\mvec{m}_i)$. Rearranging \Cref{eq:approx_h_cluster_pot} in terms of
$\approxBel(\mvec{h}, \mvec{m}_i)$ we can calculate the expectation
\begin{equation}
  \label{eq:bel_h_m}
  \begin{split}
    \log
    \approxBel(\mvec{h}, \mvec{m}_i)
    &\propto
    \left\langle
      \log
      \left(
        \cp_i(\mvec{\theta},\mvec{m}_i)
        \approxinmsg{i}{p}(\mvec{\theta})
      \right)
    \right\rangle_{\approxBel(\nu)}\\
    &
    \begin{split}
      =&\log
      p(\gamma_i|\alpha_i,\beta_i,\mvec{h},
      \nu=\nu_{\belief})\\
      &+\log
      p(\mvec{z}_i=\given{z}_i|\mvec{m}_i)\\
      &+\log
      \left(
        p(\alpha_i)
        p(\beta_i)
        \approxinmsg{i}{p}(\mvec{h})
      \right)
      ,
    \end{split}
  \end{split}
\end{equation}
where
\begin{equation}
  \label{eq:expected_deviation}
  \nu_{\belief} = \left\langle \nu^{-1} \right\rangle_{\approxBel(\nu)}^{-1}= \frac{b_\nu}{a_\nu}.
\end{equation}
In order to find an analytical expression for $\approxBel(\mvec{h},\mvec{m}_i)$,
we consider each of the constituent distributions in \Cref{eq:bel_h_m}. As
mentioned in \Cref{sec:decouple}, the \emph{measurement distribution} is
Gaussian distributed, which can be rearranged such that the mean of the
distribution is the observation
\begin{equation}
  \begin{split}
    p(\mvec{z}_i=\given{z}_i|\mvec{m}_i)
    &=
    \GM(\mvec{z}_i=\given{z}_i; \mvec{m}_i, \mmat{\Sigma}_{z_i})\\
    &=
    \GM(\mvec{m}_i; \given{z}_i, \mmat{\Sigma}_{z_i}).
  \end{split}
\end{equation}
According to the surfel model in \Cref{sec:mapmodel}, the \emph{likelihood
  distribution} is Gaussian distributed
\begin{equation}
  \label{eq:like}
  p(\gamma_i|\alpha_i,\beta_i,\mvec{h}, \nu_{\belief})
  =
  \GM(\gamma_i; f(\mvec{h},\alpha_i,\beta_i), \nu_{\belief}),
\end{equation}
where $f(\cdot)$ is the mean of the stochastic process as defined in
\Cref{eq:surffunc}. We place an uninformative and independently distributed
Gaussian prior over $\alpha_i$ and $\beta_i$, and combine it with
$\approxinmsg{i}{p}(\mvec{h})$ to form the jointly Gaussian distributed
\emph{context distribution}
\begin{equation}
  p(\alpha_i) p(\beta_i)
  \approxinmsg{i}{p}(\mvec{h}) = \GM(\bmat{\mvec{h}& \alpha_i & \beta_i}^\T;
  \mvec{\mu}_c, \mmat{\Sigma}_c),
\end{equation}
with moments calculated as
\begin{equation}
  \mvec{\mu}_c =
  \begin{bmatrix}
    \mvec{\mu}\inarrow{i}{p}\\ \mu_{\alpha_i}\\ \mu_{\beta_i}
  \end{bmatrix}
  \; \text { and } \;
  \mmat{\Sigma}_c =
  \diag(\mmat{\Sigma}\inarrow{i}{p},
  \sigma^2_{\alpha_i}, \sigma^2_{\beta_i})
  ,
\end{equation}
where $\mmat{\Sigma}\inarrow{i}{p} = \mmat{\Omega}\inarrow{i}{p}^{-1}$,
$\mvec{\mu}\inarrow{i}{p} =
\mmat{\Sigma}\inarrow{i}{p}\mvec{\xi}\inarrow{i}{p}$, and the $\diag(\cdot)$
operator creates a block diagonal matrix of its arguments. Most notably, all
distributions except the likelihood distribution are exactly Gaussian
distributed over $\mvec{h}$ and $\mvec{m}_i$. This dissimilarity in likelihood
distribution is due to the nonlinear relationships introduced by $f(\cdot)$. To
ensure that the product of all the distributions is tractable, we also wish to
approximate the likelihood distribution as a Gaussian distribution. A common
approach to achieving this is by linearising; we therefore use a linear Taylor
series approximation,
\begin{equation}
  \label{eq:f_taylor}
  f(\mvec{x})
  \approx
  f(\bar{\mvec{x}})
  + \mmat{F}(\mvec{x}-\bar{\mvec{x}}),
\end{equation}
where $\mvec{x} = \bmat{\mvec{h} & \alpha_i & \beta_i}^\T$, and $\mmat{F}$ is
the Jacobian matrix of $f(\mvec{x})$ evaluated at the linearisation point
$\mvec{\mu}_c$:
\begin{equation}
  \label{eq:jacobian}
  \begin{split}
    \mmat{F} &= \left. \frac{\delta f}{\delta \mvec{x}}
    \right|_{\mvec{x}=\mvec{\mu}_c}
    =
    \left.
    \begin{bmatrix}
        1-\alpha_i-\beta_i & \alpha_i & \beta_i & h_\alpha-h_0 & h_\beta-h_0
    \end{bmatrix}
    \right|_{\mvec{\mu}_c}
  \end{split}.
\end{equation}
Using this we can calculate the product of the likelihood and context
distributions,
\begin{equation}
  \begin{split}
    &p(\gamma_i|\alpha_i,\beta_i,\mvec{h}, \nu_{\belief})
    p(\alpha_i)
    p(\beta_i)
    \approxinmsg{i}{p}(\mvec{h}) \\
    &\propto
    \GC(\bmat{\mvec{h}&\mvec{m}_i}^\T; \bar{\mvec{\xi}},
    \bar{\mmat{\Omega}})
  \end{split},
\end{equation}
with natural parameters calculated as
\begin{equation}
  \begin{split}
    \bar{\mvec{\xi}}
    &=
    \bar{\mmat{\Omega}}
    \begin{bmatrix}
      \mvec{\mu}_c\\
      f(\mvec{\mu}_c)
    \end{bmatrix}
  \end{split}
  \; \text { and } \;
  \begin{split}
    \bar{\mmat{\Omega}}
    &=
    \begin{bmatrix}
      \mmat{\Sigma}_c & (\mmat{F} \mmat{\Sigma}_c)^\T\\
      \mmat{F} \mmat{\Sigma}_c & \mmat{F} \mmat{\Sigma}_c \mmat{F}^\T+\nu_{\belief}\\
    \end{bmatrix}^{-1}
    .
  \end{split}
\end{equation}
The normalised product of the prediction distribution and the measurement
distribution results in the desired belief
\begin{equation}
  \label{eq:bel_h_m_gauss}
  \approxBel(\mvec{h}, \mvec{m}_i)
  =
  \GC(\bmat{\mvec{h}&\mvec{m}_i}^\T; \mvec{\xi}, \mmat{\Omega}),
\end{equation}
with natural parameters calculated as
\begin{equation}
  \begin{split}
    \mvec{\xi}
    &=
    \bar{\mvec{\xi}} + \mmat{S}^\T\mmat{\Sigma}_{z_i}^{-1}\given{z}_i\\
    &=
    \begin{bmatrix}
      \mvec{\xi}_{h} \\ \mvec{\xi}_{m}
    \end{bmatrix}
  \end{split}
  \quad \text { and } \quad
  \begin{split}
    \mmat{\Omega}
    &= \bar{\mmat{\Omega}} + \mmat{S}^\T\mmat{\Sigma}_{z_i}^{-1}\mmat{S}\\
    &=
    \begin{bmatrix}
     \mmat{\Omega}_{hh}^{} &\mmat{\Omega}_{hm}\\
     \mmat{\Omega}_{mh}^{}&\mmat{\Omega}_{mm}^{}\\
    \end{bmatrix}
  \end{split},
\end{equation}
where $ \mmat{S} =\bmat{\mmat{0}_{3\text{x}3} & \mmat{I}_{3\text{x}3}}$ is a
selection matrix. Notably, the above method is equivalent to an extended
information filter (EIF) with a nonlinear state transition model and a linear
measurement model \citep[ch.~3.5.4]{Thrun2005}. Finally, we can calculate
$\approxcp_i(\mvec{h},\mvec{m}_i)$ by dividing out
$\approxinmsg{i}{p}(\mvec{h})$ from $\approxBel(\mvec{h},\mvec{m}_i)$
(\Cref{eq:bel_h_m_defn}), which results in
\begin{equation}
    \approxcp_i(\mvec{h},\mvec{m}_i)
    =
    \GC(\bmat{\mvec{h}&\mvec{m}_i}^\T;
    \mvec{\xi}_{i}, \mmat{\Omega}_{i}),
\end{equation}
with natural parameters calculated as
\begin{equation}
  \begin{split}
    \mvec{\xi}_{i}
    &=
    \begin{bmatrix}
      \mvec{\xi}_{h}- \mvec{\xi}\inarrow{i}{p}
      \\ \mvec{\xi}_{m}
    \end{bmatrix}
  \end{split}
  \; \text{ and } \;
  \begin{split}
    \mmat{\Omega}_{i}
    &=
    \begin{bmatrix}
      \mmat{\Omega}_{hh}^{} - \mmat{\Omega}\inarrow{i}{p}&
      \mmat{\Omega}_{hm}\\
     \mmat{\Omega}_{mh}^{}&\mmat{\Omega}_{mm}^{}\\
    \end{bmatrix}.
  \end{split}
\end{equation}

From this result, we can now calculate a part of the outgoing message,
$\approxoutmsg{i}{p}(\mvec{h})$, by marginalising out $\mvec{m}_i$. Therefore,
using \Cref{eq:message_out}, we can calculate the natural parameters of
$\approxoutmsg{i}{p}(\mvec{h})$ (\Cref{eq:dist_msg_out}) as
\begin{equation}
  \begin{split}
    \mmat{\Omega}\outarrow{i}{p}
    &=
    \mmat{\Omega}_{hh}^{} -
    \mmat{\Omega}\inarrow{j}{p}
    - \mmat{\Omega}_{hm}^{}\mmat{\Omega}_{mm}^{-1} \mmat{\Omega}_{mh}
  \end{split}
\end{equation}
and
\begin{equation}
  \begin{split}
    \mmat{\xi}\outarrow{i}{p}
    &=
    \mvec{\xi}_h^{} - \mmat{\xi}\inarrow{j}{p} - \mmat{\Omega}_{hm}^{}
    \mmat{\Omega}_{mm}^{-1} \mvec{\xi}_m^{}.
  \end{split}
\end{equation}

We have now discussed how to calculate the likelihood cluster potential
$\approxcp_i(\mvec{h}, \mvec{m}_i)$ and the outgoing message
$\approxoutmsg{i}{p}(\mvec{h})$. Next, we focus on the likelihood cluster
potential and outgoing message over the planar deviation.

\subsubsection{Updating \texorpdfstring{$\approxcp_i(\nu)$}{Planar Deviation
    Factor}}
\label{sec:update_v}

Similarly to calculating $\approxcp_i(\mvec{h},\mvec{m}_i)$, to perform the
local information projection of \Cref{eq:kl_local} with respect to
$\approxcp_i(\nu)$, the VMP algorithm \cite{Winn2005} defines the iterative
update as
\begin{equation}
  \begin{split}
    &\log
    \approxcp_i(\nu)
    +
    \log
    \approxinmsg{i}{p}(\nu)
    \propto
    \left\langle
      \log \cp_i(\mvec{\theta},\mvec{m}_i)
      \approxinmsg{i}{p}(\mvec{\theta})
    \right\rangle
    _{\approxBel(\mvec{h},\mvec{m}_i)}
    .
  \end{split}
\end{equation}
The marginalisation to calculate the outgoing message,
$\approxoutmsg{i}{p}(\mvec{\theta})$---\Cref{eq:message_out}---will have no
effect on the outgoing message, $\approxoutmsg{i}{p}(\nu)$, due to our choice of
factorisation. Therefore, solving for the cluster potential is equivalent to
solving for the outgoing message---$\approxoutmsg{i}{p}(\nu)=
\approxcp_i(\nu)$---which we can directly calculate as
\begin{equation}
  \begin{split}
    \log
    \approxoutmsg{i}{p}(\nu)
    &\propto
    \left\langle
    \log
      p(\gamma_i|\alpha_i,\beta_i,\mvec{h}, \nu)
    \right\rangle_{\approxBel(\mvec{h},\mvec{m}_i)}\\
    &\propto
    -a\outarrow{i}{p}\log \nu - \frac{b\outarrow{i}{p}}{\nu}
    ,
    \end{split}
\end{equation}
where $\approxBel(\mvec{h}, \mvec{m}_i)$ is calculated from \Cref{eq:bel_h_m}.
The resulting approximate potential matches the form of an inverse-gamma
distribution, with shape and scale parameters
\begin{equation}
 \label{eq:b_update}
 a\outarrow{i}{p} = \frac{1}{2}
 \;\; \text{ and } \;\;
 b\outarrow{i}{p} =
 \frac{1}{2}
 \left\langle
   (\gamma_i - f(\alpha_i,\beta_i,\mvec{h}))^2
 \right\rangle_{\approxBel(\mvec{h},\mvec{m}_i)},
\end{equation}
where $f(\cdot)$ is defined by \Cref{eq:surffunc}. We again use the linear
Taylor series approximation to calculate the expectation\footnote{ This
  expectation could be calculated analytically, because $f(\cdot)$ is a
  multilinear function and $\approxBel(\mvec{h},\mvec{m}_i)$ is Gaussian
  distributed. However, we found that, in rare cases, it was problematic for the
  initially uncertain stages of message passing, and consequently hindered
  convergence. In contrast, using the Taylor series approximation was found to
  be a more robust approach. Additionally, the Taylor series is computationally
  cheaper to compute.}, where we linearise around the mean of
$\approxBel(\mvec{h}, \mvec{m}_i)$. This results in the expectation
\begin{equation}
  b\outarrow{i}{p} =
  \frac{1}{2}
  \mmat{F}_{\text{aug}}^{}\mmat{\Sigma}^{}\mmat{F}_{\text{aug}}^\T
  +
  \frac{1}{2}
  (\mu_\gamma - f(\mu_\alpha, \mu_\beta, \mvec{\mu}_h))^2,
\end{equation}
where $\mmat{F}_{\text{aug}} = \bmat{\mmat{F} & 1}$---with $\mmat{F}$ defined
by \Cref{eq:jacobian}---and $\mmat{\Sigma} = \mmat{\Omega}^{-1}$ and $\mvec{\mu}
= \mmat{\Sigma}\mvec{\xi}$ are the moments of the distribution in
\Cref{eq:bel_h_m_gauss}.

We have now discussed how to perform inference on each surfel in isolation.
Before discussing performing inference on the full model, we briefly look at the
performance of the proposed inference algorithm.

\subsubsection{Inference Performance}
\label{sec:infer-perf}

\begin{figure*}[h!]
  \centering
  \subfloat[]{
    \includegraphics[width=0.31\textwidth]{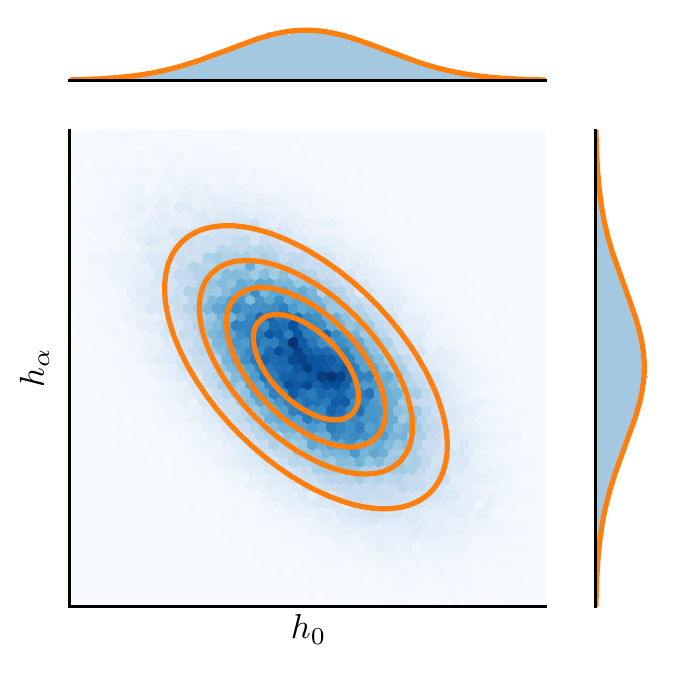}
  }
  \subfloat[]{
    \includegraphics[width=0.31\textwidth]{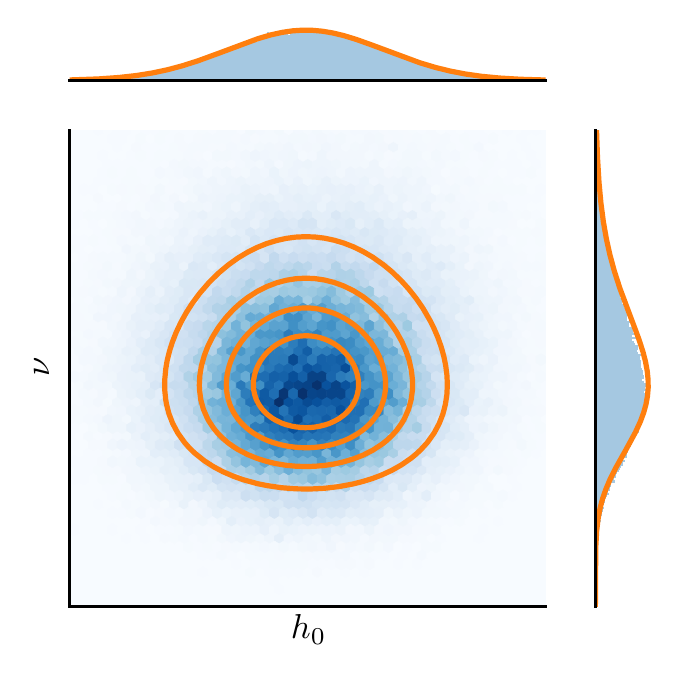}
  }
  \subfloat[]{
    \includegraphics[width=0.31\textwidth]{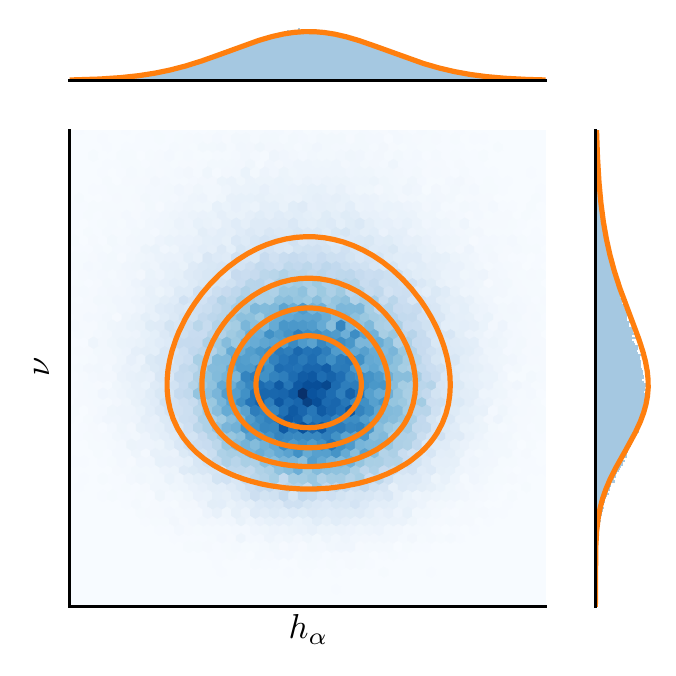}
  }

  \subfloat[]{
    \includegraphics[width=0.31\textwidth]{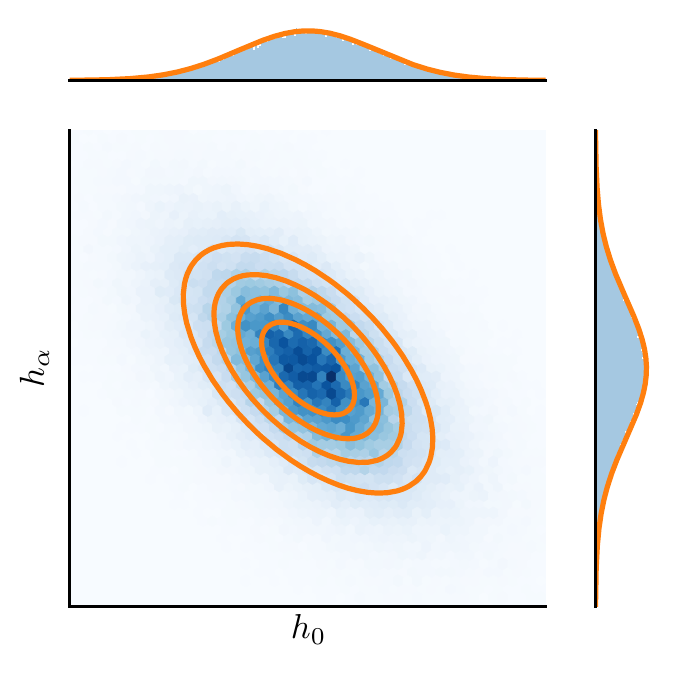}
  }
  \subfloat[]{
    \includegraphics[width=0.31\textwidth]{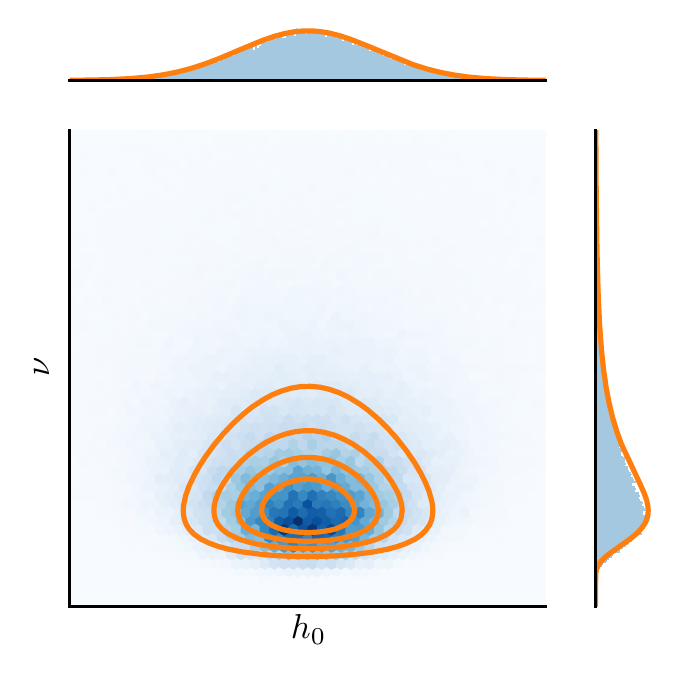}
  }
  \subfloat[]{
    \includegraphics[width=0.31\textwidth]{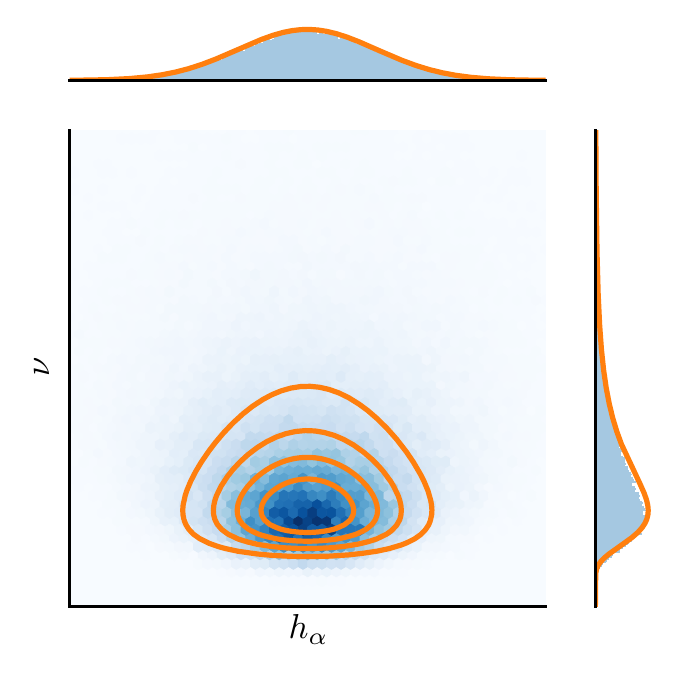}
  }
  \caption{The accuracy of our inference algorithm in calculating an approximate
    surfel model belief (orange), compared against \ac{MCMC} samples of the
    exact belief distribution (blue) for a 2-D simulation. We emulate two
    different sensor modalities: (a-c) a stereo camera by using 100 measurements
    with high range and low bearing uncertainty, and (d-f) a LiDAR sensor by
    using 10 measurements with low measurement uncertainty. Shown for these two
    cases are the three 2-D projections of the 3-D belief distribution over
    $h_0$, $h_\alpha$, and $\nu$, and the marginal distributions for each axis.
    Note that we do not show the axis ticks, as we are only concerned with the
    \emph{relative} difference between the exact belief and our approximation.}
  \label{fig:mcmc_comp}
\end{figure*}

To illustrate the performance of the proposed inference algorithm, we compare
our approximation of the surfel model belief against \ac{MCMC} samples of the
exact belief distribution for a 2-D simulation---using the Metropolis-Hastings
algorithm to generate samples. \Cref{fig:mcmc_comp} shows representative cases
for a stereo camera and a LiDAR sensor. We see that our algorithm accurately
matches the marginal distributions from \ac{MCMC}. Although we do not
demonstrate it, this is not always the case; when the measurement uncertainty is
unreasonably high\footnote{The measurement uncertainty relative to the size of
  the grid element increases with the grid division depth. Therefore, too fine a
  grid division could result in a measurement distribution that spans several
  grid elements.}, our algorithm results differ noticeably from the exact
belief; this is due to errors induced through linearisation. However, when more
measurements are added, this effect is mitigated. We also did not encounter any
issues with this when using either realistically simulated or practical
data---as we will see in \Cref{sec:results}.

We have shown that this method of approximation can accurately represent the
surfel belief; however, this was only for the case where each surfel is updated
in isolation. We will now consider inference on the full model, utilising the
approach we have developed here.


\subsection{Inference on the Full Model}
\label{sec:full-model-inference}

As opposed to calculating the joint belief distribution over all the surfel
parameters in \iac{STM} map, we instead wish to calculate the marginal belief
distributions over each surfel's parameters. However, as there are shared height
parameters between contiguous surfels, we cannot naively use exactly the same
approach as we did previously to perform inference on the full model. It should
be noted that, although we will consider the joint distribution over all the map
parameters, we do not calculate this expressly, as the marginal surfel beliefs
are our desired end result.

We can reformulate the inference problem over all the map parameters by
utilising our previous approach of performing independent surfel inference. For
a given surfel, $s$, we assume that measurements within the associated grid
element, $\mset{Z}_s$, and the points on the surface of the environment,
$\mset{M}_s$, are independent of all other measurements, $\mset{Z} \setminus
\mset{Z}_s$, and all other surface points, $\mset{M} \setminus \mset{M}_s$,
given the parameters of the surfel, $\mvec{\theta}_s$. Following this
assumption, and using Bayes' theorem, we can factorise the full joint
distribution
\begin{equation}
  \label{eq:full_joint_1}
  \begin{split}
    p(\mset{\Theta}, \mset{M}, \mset{Z}=\given{Z})
    &= p(\mset{\Theta}) p(\mset{M}, \mset{Z}=\given{Z} | \mset{\Theta})\\
    &= p(\mset{\Theta})
    \prod_{s \in \mset{S}} p(\mset{M}_s, \mset{Z}_s=\given{Z}_s | \mvec{\theta}_s)
    ,
  \end{split}
\end{equation}
where $\mset{S}$ is the set of all surfels in \iac{STM} map. Each conditional
distribution, $p(\mset{M}_s, \mset{Z}_s=\given{Z}_s | \mvec{\theta}_s)$, is
described by the same model we used previously, namely
\begin{equation}
  \label{eq:like_dist}
  p(\mset{M}_s, \mset{Z}_s=\given{Z}_s | \mvec{\theta}_s) =
  \prod_i\cp_i(\mvec{\theta}_s, \mvec{m}_i),
\end{equation}
where we again refer to $\cp_i(\mvec{\theta}_s, \mvec{m}_i)$ as the likelihood
cluster potential---as defined in \Cref{eq:joint_bayes,eq:joint_cluster}. We
factorise the prior distribution into a product of prior cluster potentials,
\begin{equation}
  p(\mset{\Theta}) \propto \prod_{s \in \mset{S}} \cp_p(\mvec{\theta}_s).
\end{equation}
As with our previous approach, each prior cluster potential is distributed
according to \Cref{eq:prior}---namely a Gaussian distribution over the vertex
heights and an inverse-gamma distribution over the planar deviation. We will
examine the effect that this factorisation has on the resulting surfel beliefs,
and how the parameter choices (specifically for the height prior distribution)
could encode additional structure in the environment
(\Cref{sec:experiment-priors}). Combining the prior factorisation with
\Cref{eq:like_dist}, we obtain
\begin{equation}
  \label{eq:joint_surfels}
  p(\mset{\Theta}, \mset{M}, \mset{Z}=\given{Z})
  =
  \prod_{s \in \mset{S}} \cp_s(\mvec{\theta}_s, \mset{M}_s)
  ,
\end{equation}
where the surfel cluster potential, $\cp_s(\mvec{\theta}_s, \mset{M}_s,
\mset{Z}_s)$, is calculated according to
\begin{equation}
  \begin{split}
    \cp_s(\mvec{\theta}_s, \mset{M}_s)
    &=
    \cp_p(\mvec{\theta}_s)
    p(\mset{M}_s, \mset{Z}_s=\given{Z}_s | \mvec{\theta}_s)\\
    &=
    \cp_p(\mvec{\theta}_s)
    \prod_i\cp_i(\mvec{\theta}_s, \mvec{m}_i)
    .
  \end{split}
\end{equation}

To calculate a surfel belief for some given measurements, $\mset{Z}=\given{Z}$,
we must marginalise out $\setminus \mvec{\theta}_s$---all variables except for
$\mvec{\theta}_s$---from the full joint distribution,
\begin{equation}
  \label{eq:surfel_belief_1}
  \begin{split}
    \belief(\mvec{\theta}_s)
    &\propto
    \int\displaylimits_{\setminus \mvec{\theta}_s}
    p(\mset{\Theta}, \mset{M}, \mset{Z}=\given{Z})
    \, \text{d}\setminus \mvec{\theta}_s\\
    &=
    \int\displaylimits_{\setminus \mvec{\theta}_s}
    \prod_{s \in \mset{S}} \cp_s(\mvec{\theta}_s, \mset{M}_s)
    \, \text{d}\setminus \mvec{\theta}_s
    .
  \end{split}
\end{equation}
The surfel belief can be separated into two terms: one containing the surfel
cluster potential of a surfel, $s \in \mset{S}$, and the other containing the
surfel cluster potentials of the rest of surfels, $\mset{R} = \mset{S} \setminus
s$, namely
\begin{equation}
  \label{eq:surfel_belief_2}
  \begin{split}
    &\belief(\mset{\theta}_s)
    \propto
      \underbrace{
      \int\displaylimits_{\vphantom{\setminus}\mset{M}_s}
      \cp_s(\mvec{\theta}_s, \mset{M}_s)
      \, \text{d}\mset{M}_s
      }_
      \mathlarger{
        \mathlarger{
          \cp_p(\mvec{\theta}_s)
          \prod\limits_{i}\outmsg{i}{p}(\mvec{\theta}_s)
        }
      }
      \underbrace{
      \int\displaylimits_{\setminus \mvec{\theta}_s}
      \prod_{r \in \mset{R}} \cp_r(\mvec{\theta}_r, \mset{M}_r)
      \, \text{d}\setminus \mvec{\theta}_s
    }_
    \mathlarger{
      \mathlarger{
        \inmsg{s}{\mset{R}}(\mvec{h}_s)
        \vphantom{\prod\limits_i}
      }
    }
    .
  \end{split}
\end{equation}
The first factor is identical to what was calculated previously when considering
each surfel in isolation, with the messages from each likelihood cluster
calculated according to \Cref{eq:message_out}. We therefore assume that we can
calculate this term. The second term is the message from the rest of the surfels
in the map, which is due to the shared heights between surfels. This message
essentially captures what the rest of the surfels in the map surmise about the
heights of the current surfel. Because of the highly coupled nature of the
problem, the marginalisation to calculate $\inmsg{s}{\mset{R}}(\mvec{h}_s)$ is
expensive to compute. To keep things tractable, we approximate this message
using a technique known as \ac{LBP} \citep{Frey1998}---a message-passing
technique that is particularly well suited to calculating approximate marginal
distributions without expressly calculating the full joint distribution. \ac{LBP} is
an iterative application of the integral-product algorithm applied to cyclic
graphs.

\begin{figure*}[h!]
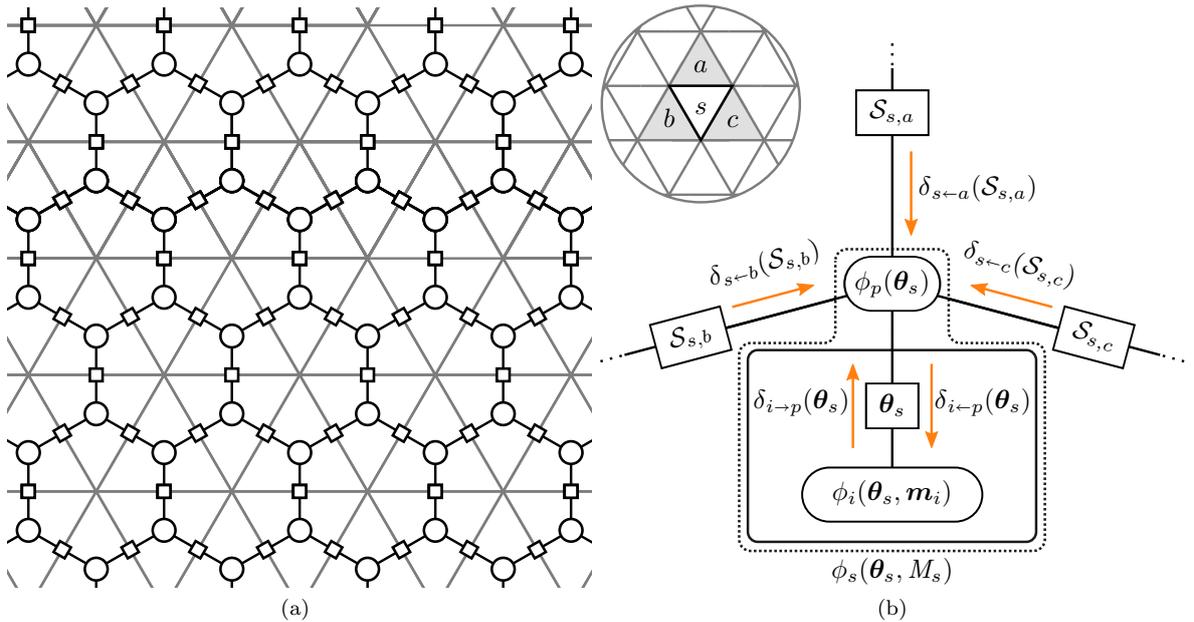

  \centering
  \subfloat[]{
    \includesvg[width=0.48\textwidth]{full_model_cg}%
  }
  \subfloat[]{
    \includesvg[width=0.48\textwidth]{full_model_cg_zoom}%
    \label{fig:full_cg_zoom}
  }
  \caption{(a) The cluster graph for the full inference problem shown for a
    zoomed-in section of \iacf{STM} map. We indicate the sepsets between
    clusters with $\boxempty$, and each surfel cluster potential with
    $\bigcirc$. The grey lines indicate the grid elements of the map. (b) We
    zoom in to the cluster graph, focusing on a specific surfel, $s$, which is
    contiguous with surfels $a$, $b$ and $c$. The resulting graph is similar to
    the cluster graph in \Cref{fig:cg}, but there are now connections between
    contiguous surfels. The expanded surfel cluster potential
    $\cp_s(\mvec{\theta}_s, \mset{M}_s)$ is shown using the dotted outline. The
    sepset between the surfels is denoted $\mathcal{S}$ (see text for more
    details on the scope of the sepsets).}
  \label{fig:full_cg}
\end{figure*}

In order to use LBP, we must first construct a cluster graph for the full
inference problem. We previously performed inference by only considering each
surfel cluster potential when calculating each surfel belief in isolation. We
therefore can construct the cluster graph of the full joint distribution,
$p(\mset{\Theta}, \mset{M}, \mset{Z}=\given{Z})$, by repeating the previous
cluster graph (\Cref{fig:cg}) for each surfel in the map, but also creating
connections between surfels that share an edge. The resulting cluster graph is
illustrated in \Cref{fig:full_cg}. The scope of the sepset between surfels is
generally over the two heights from the shared edge between the surfels---for
example $\sepset{s}{a} = \mvec{h}_s \cap \mvec{h}_{a}$. Although it is not shown
explicitly, to ensure a valid cluster graph of the joint distribution it is also
necessary to reduce the scope of some sepsets so that the graph obeys the
\emph{running intersection property}, which states that there can only be a
single path between the same variable in different parts of the graph. To ensure
this property is obeyed, we use the LTRIP algorithm of \citet{Streicher2017} to
construct a cluster graph for the fully-connected model.

From this cluster graph, we can instead calculate an approximation of the
message from the rest of the surfels in the map,
$\approxinmsg{s}{\mset{R}}(\mvec{h}_s)$, by only considering the messages from
the subset of surfels that are contiguous to the current surfel, $\mset{C}
\subseteq \mset{R}$. This results in a much simpler and tractable calculation,
\begin{equation}
  \label{eq:approx_rest_msg}
  \approxinmsg{s}{\mset{R}}(\mvec{h}_s)
  \propto
  \prod_{c \in \mset{C}}\inmsg{s}{c}(\sepset{s}{c}),
\end{equation}
where $\mathcal{S}$ denotes the sepset between surfels. It should be noted that
the scope of the resulting message might not always contain all the heights in
the surfel (as would be the case for surfels on the border of the submap, or if
the sepset scope has been reduced to satisfy the running intersection property).
Using the integral-product algorithm, \ac{LBP} defines the iterative update of a
message from one surfel, $s$, to another surfel, $a$, according to
\begin{equation}
  \begin{split}
    &\inmsg{a}{s}(\sepset{a}{s})
    \propto
    \int\displaylimits_{\setminus\sepset{a}{s}}
    \cp_p(\mvec{\theta}_s)
    \prod\limits_{i}\outmsg{i}{p}(\mvec{\theta}_s)
    \prod_{c \in \mset{C} \setminus a}\inmsg{s}{c}(\sepset{s}{c})
    \, \text{d}\setminus\sepset{a}{s}.
  \end{split}
\end{equation}
Using \Cref{eq:surfel_belief_2}, this can be simplified to
\begin{equation}
  \label{eq:out_msg_neigh}
  \begin{split}
    &\inmsg{a}{s}(\sepset{a}{s})
    \propto
    \int\displaylimits_{\setminus\sepset{a}{s}}
    \frac{\belief(\mvec{\theta}_s)}{\inmsg{s}{a}(\sepset{a}{s})}
    \, \text{d}\setminus\sepset{a}{s}
    .
  \end{split}
\end{equation}
By iteratively passing these messages between surfels, \ac{LBP} approximately
calculates the marginal belief of each surfel. In order to perform this
marginalisation and calculate each surfel belief, the likelihood cluster
potentials in each surfel potential need to have a closed-form solution, but
this is not the case due to the nonlinearities between the variables in the
likelihood cluster potentials. In order to find a closed-form solution to each
surfel belief (and consequently the map belief), we again turn to \acf{VMP} to
approximate each likelihood cluster potential; that is, using the local
information projection according to \Cref{eq:kl_local}, repeated here for
clarity:
\begin{equation*}
  \approxcp_{i}^*(\mvec{\theta}_s, \mvec{m}_i)
  =
  \argmin_{\approxcp_{i} \in \mathcal{F}}
  \KL
  {\approxcp_{i} \approxinmsg{i}{p}}
  {\cp_i \approxinmsg{i}{p}}
  .
\end{equation*}
However, the approximate incoming message,
$\approxinmsg{i}{p}(\mvec{\theta}_s)$, is now calculated according to
\begin{equation}
  \label{eq:approx_inmsg_nodiv}
  \approxinmsg{i}{p}(\mvec{\theta}_s)
  =
  \cp_p(\mvec{\theta}_s)
  \approxinmsg{s}{\mset{r}}(\mvec{h}_s)
  \prod_{j \neq i}\approxoutmsg{j}{p}(\mvec{\theta}_s)
  .
\end{equation}
Similarly to previously---\Cref{eq:message_in}---this message is calculated by
combining the prior cluster potential, $\cp_p(\mvec{\theta}_s)$, with the
messages from all other likelihood clusters, $\prod_{j \neq
  i}\approxoutmsg{j}{p}(\mvec{\theta}_s)$, but now it additionally combines the
message from neighbouring surfels, $\approxinmsg{s}{\mset{r}}(\mvec{h}_s)$.
Using \Cref{eq:surfel_belief_2}, the approximate incoming message can be
rewritten in terms of the surfel belief,
\begin{equation}
  \label{eq:approx_inmsg_div}
  \approxinmsg{i}{p}(\mvec{\theta}_s)
  \propto
  \frac
  {\approxBel(\mvec{\theta}_s)}
  {\approxoutmsg{i}{p}(\mvec{\theta}_s)}
  ,
\end{equation}
which, in comparison to \Cref{eq:approx_inmsg_nodiv}, is cheaper to calculate
when iterating over all the measurements. If we combine the prior potential,
$\cp_p(\mvec{\theta}_s)$, with the incoming message,
$\approxinmsg{s}{\mset{r}}(\mvec{h}_s)$, into a pseudo-prior, the same approach
to what we described previously can be used to perform the local information
projection. This specifically includes using the Gaussian family of
distributions to model the height distributions, which results in \ac{LBP} performing
approximate inference on only Gaussian distributions---a variant of \ac{LBP} known as
Gaussian \ac{LBP}.
Gaussian \ac{LBP} is a standard problem formulation for which implementations
are available, and we therefore use \acs{EMDW}---an in-house PGM
library\footnote{This is currently in the process of being made open source.}.
\ac{LBP} is a tractable approximate inference technique; however, due to its
iterative nature, its convergence has not yet been proven. In Gaussian \ac{LBP},
however, it is known that, if convergence is reached, the posterior means are
correct \citep{Rusmevichientong2001,Weiss2001}. Although the surfel belief
distributions in our problem are not guaranteed to be exactly Gaussian this is a
fair assumption as the number of measurements becomes large enough.
Additionally, \citet{Weiss2001} showed that convergence is guaranteed when
performing Gaussian \ac{LBP} on problems with diagonally dominant information
matrices. In our problem we expect local regions of the environment to be
correlated. Under this assumption, the underlying model should be diagonally
dominant. To support this, in our implementation we have not empirically
experienced any issues with convergence.

The approach presented here performs inference on \iac{STM} map using a hybrid
message-passing technique using \acf{LBP} and \acf{VMP}. Next we provide an
overview of the proposed inference algorithm, and discuss some aspects of its
implementation.


\subsection{Algorithm Overview and Implementation}
\label{sec:algorithm}

In \Cref{alg:inference} we describe the process of performing inference given a
single batch of measurements. Note that, this could be a batch of measurements
accumulated over time, or from a single time step. Notably,
\cref{alg:inf_lbp1,alg:inf_lbp2,alg:inf_lbp3,alg:inf_lbp4,alg:inf_lbp5,alg:inf_lbp6}
correspond to the \acf{LBP} update steps, and
\cref{alg:inf_vmp1,alg:inf_vmp2,alg:inf_vmp3,alg:inf_vmp4,alg:inf_vmp5,alg:inf_vmp6}
to the \acf{VMP} update steps. We also adjust the method of updating each surfel
belief to be more efficient: if only the incoming message from neighbouring
surfels has changed during the iterative update process, then it is unnecessary
to calculate the updated surfel belief using \Cref{eq:surfel_belief_2}. We
instead (equivalently) simplify the update to
\begin{equation}
  \label{eq:surfel_belief_3}
  \approxBel(\mvec{\theta}_s)
  \triangleq
  \approxBel(\mvec{\theta}_s)
  \frac
  {
    \approxinmsg*{s}{\mset{R}}(\mvec{h}_s)
  }
  {
    \approxinmsg{s}{\mset{R}}(\mvec{h}_s)
  }
  ,
\end{equation}
where we denote the updated and previous incoming messages
$\approxinmsg*{s}{\mset{R}}$ and $\approxinmsg{s}{\mset{R}}$ respectively. We
next discuss some of the aspects of the algorithm.

\begin{algorithm*}[!h]
  \caption{\acs{STM} Map Inference}
  \label{alg:inference}
  \hspace*{1.5em}\textbf{Inputs}:\\
  \hspace*{2.5em}Prior cluster potentials for each surfel:
  $\cp_p(\mvec{\theta}_s)$ $\forall\,\, s \in \mset{S}$\\
  \hspace*{2.5em}Measurements in each surfel: $\given{Z}_s$ $\forall\,\, s \in \mset{S}$\\
  \hspace*{1.5em}\textbf{Output}:\\
  \hspace*{2.5em}Surfel beliefs: $\approxBel(\mvec{\theta}_s)$ $\forall\, s \in \mset{S}$\\
  \begin{algorithmic}[1]
    \State Initialise all messages

    \Statex{~}

    \LineComment{-2}{Initialise all surfel beliefs\dotfill\Cref{eq:surfel_belief_2}}

    \ForAll{$s \in \mset{S}$}

    \State
    $\approxBel(\mvec{\theta}_s)
    \gets
    \phi_p(\mvec{\theta}_s)
    \approxinmsg{s}{\mset{R}}(\mvec{h}_s)
    \prod\limits_{i}\approxoutmsg{i}{p}(\mvec{\theta}_s)$

    \EndFor

    \Statex{~}

    \Repeat

    \LineComment{-0.5}{Cycle through all the surfels in the map}

    \ForAll{$s \in \mset{S}$}

    \LineComment{1}{Calculate the updated incoming message from neighbouring
      surfels\dotfill\Cref{eq:approx_rest_msg}}

    \State
    $\approxinmsg*{s}{R}(\mvec{h}_s)
    \gets
    \prod\limits_{c \in \mset{C}}\inmsg{s}{c}(\sepset{s}{c})$
    \label{alg:inf_lbp1}

    \Statex{~}

    \LineComment{1}{Update the surfel belief\dotfill\Cref{eq:surfel_belief_3}}
    \State
    $\approxBel(\mvec{\theta}_s)
    \gets
    \approxBel(\mvec{\theta}_s)
    \frac
    {
      \approxinmsg*{s}{\mset{R}}(\mvec{h}_s)
    }
    {
      \approxinmsg{s}{\mset{R}}(\mvec{h}_s)
    }
    $
    \label{alg:inf_lbp2}

    \State
    $\approxinmsg{s}{R}(\mvec{h}_s)
    \gets
    \approxinmsg*{s}{\mset{R}}(\mvec{h}_s)$
    \label{alg:inf_lbp3}

    \Statex{~}

    \LineComment{1}{Update the likelihood cluster potential for each measurement}

    \ForAll{$\given{z}_i \in \given{Z}_s$}
    \label{alg:inf_vmp1}

    \LineComment{2.5}{Calculate incoming message to the likelihood
      cluster\dotfill\Cref{eq:approx_inmsg_div}}

    \State
    $\approxinmsg{i}{p}(\mvec{\theta}_s)
    \gets
    \frac
    {\approxBel(\mvec{\theta}_s)}
    {\approxoutmsg{i}{p}(\mvec{\theta}_s)} $
    \label{alg:inf_vmp2}

    \Statex{~}

    \LineComment{2.5}{Update the likelihood cluster potential\dotfill\Cref{sec:update_h_m,sec:update_v}}

    \State
    $\approxcp_{i}(\mvec{h}, \mvec{m}_i)
    \approxcp_{i}(\nu)
    \gets
    \argmin_{\approxcp_{i}}
    \KL
    {\approxcp_{i} \approxinmsg{i}{p}}
    {\cp_{i} \approxinmsg{i}{p}}$
    \label{alg:inf_vmp3}

    \Statex{~}

    \LineComment{2.5}{Calculate the outgoing message from the likelihood
      cluster\dotfill\Cref{eq:message_out}}

    \State
    $\approxoutmsg{i}{p}(\mvec{\theta}_s)
    \gets
    \approxcp_{i}(\nu)
    \int \approxcp_{i}(\mvec{h}, \mvec{m}_i) \, \text{d}\mvec{m}_i$
    \label{alg:inf_vmp4}

    \Statex{~}

    \LineComment{2.5}{Update the surfel belief\dotfill\Cref{eq:approx_bel}}

    \State
    $\approxBel(\mvec{\theta}_s)
    \gets
    \approxoutmsg{i}{p}(\mvec{\theta}_s)
    \approxinmsg{i}{p}(\mvec{\theta}_s)$
    \label{alg:inf_vmp5}

    \EndFor
    \label{alg:inf_vmp6}

    \Statex{~}

    \LineComment{1}{Calculate the outgoing messages to neighbouring surfels\dotfill\Cref{eq:out_msg_neigh}}

    \ForAll{$c \in C$}
    \label{alg:inf_lbp4}

    \State
    $\inmsg{c}{s}(\sepset{c}{s})
    =
    \int
    \frac{\belief(\mvec{\theta}_s)}{\inmsg{s}{c}(\sepset{c}{s})}
    \, \text{d}\setminus\sepset{c}{s}$
    \label{alg:inf_lbp5}

    \EndFor
    \label{alg:inf_lbp6}

    \EndFor
    \Until{all messages converge}
  \end{algorithmic}
\end{algorithm*}

\paragraph{Message initialisation}
We first need to initialise all of the messages before we can begin message
passing. In general, the initialisation for iterative approximate inference
algorithms can influence the performance, affecting the local optimum to which
the algorithm converges, and the convergence speed. We initialise the messages
between surfels to be vacuous. For the outgoing messages from each likelihood
cluster, we use a simple empirical method of initialisation. Specifically, we
initialise each outgoing message, $\approxoutmsg{i}{p}(\mvec{h}_s)$, to have a
mean at the $\gamma$ component of the measurement $\given{z}_i$, and a
(practically) uninformative covariance. This ensures that $\approxBel(\mvec{h})$
is initialised to have a mean at the average $\gamma$ component of the
measurements. We initialise each outgoing message, $\approxoutmsg{i}{p}(\nu_s)$,
such that the expected value calculated in \Cref{eq:expected_deviation} is the
variance of the $\gamma$ component of the measurements.

\paragraph{Incremental updating}
During online operation, measurements are received incrementally, and we wish to
calculate the surfel beliefs based on all currently available measurements.
Ideally, the surfel beliefs should be recalculated as new measurement batches
arrive, while also considering all previous measurements. However, as more
measurement batches are received, this approach would quickly become
intractable. A viable approximation is to only update the outgoing messages from
a window of the $W$ latest measurements \citep{Qi2007}. Fixing the messages from
clusters containing measurements outside this window ensures that updating the
surfel belief is linear in the window size $W$. In practice, we combine these
constant messages with the prior, discarding the old measurements so that the
storage will not grow unbounded. This new prior now summarises all the previous
measurements. Following this approach, we can then use the same procedure
outlined in \Cref{alg:inference}. It should also be noted that previously
calculated messages between surfels should not be discarded, but now used as the
initialisation for the next batch update.

\paragraph{Convergence}
To monitor the state of convergence, we use the fact that all messages will be
constant at convergence---that is, convergence has been reached if all messages
do not change between iterations. In practice, a message is deemed to be
constant if the difference between the successive iterations of the message is
negligible. To evaluate this difference, we use the exclusive \acf{KL}
divergence\footnote{However, other divergences could be used: the symmetric
  \ac{KL} divergence, inclusive \ac{KL} divergence, the Mahalanobis distance, or
  the Bhattacharyya distance, to name a few.} between a message at the current
and previous iterations---$\KL{\text{current iteration}}{\text{previous
    iteration}}$---and the message has converged if the resulting divergence is
below a threshold. If all messages related to a surfel have converged, then it
is no longer necessary to iterate over it. It should also be noted that,
although convergence is not theoretically guaranteed (due to the use of
\acf{LBP}), we have not empirically experienced any issues with convergence in
our implementation.

\paragraph{Computational complexity}
We can see that the asymptotic computational complexity of our algorithm is
$\bigO{KN}$, where $K$ is the number of iterations in the outermost loop before
convergence, and $N$ is the total number of measurements being incorporated into
the map. In practice, $K$ is constant with respect to $N$, and therefore the
algorithm can be considered approximately linear in $N$. We support this claim
with experimental results (\Cref{sec:converge-experiment}).


We have now described performing inference on \iac{STM} map. Next we look at a
series of experiments evaluating the use \iac{STM} map to represent environments
with both simulated and practical data.


\section{Experimental Results}
\label{sec:results}

Up to this point we have discussed how \iac{STM} map models the environment, as
well as the process of performing inference in the map. We now investigate how
the \ac{STM} mapping technique performs under different experimental conditions.
Note that, although we only test \ac{STM} mapping using data from ground
vehicles, it is applicable to all types of autonomous robots: terrestrial,
underwater, and aerial.

For the first set of experiments, we use simulated measurement data
(\Cref{sec:experiment-priors,sec:converge-experiment,sec:model-compare-experiment}).
We investigate how the prior distribution parameter choices affect the resulting
map (\Cref{sec:experiment-priors}), analyse the cost of performing inference
(\Cref{sec:converge-experiment}), and compare the accuracy of \iac{STM} mapping
to elevation mapping (\Cref{sec:model-compare-experiment}).

We then consider practical datasets collected using either stereo cameras or a
\ac{LiDAR} sensor
(\Cref{sec:relative-irf-experiment,sec:global-irf-experiment}). We demonstrate
the modelling performance of the \ac{STM} mapping technique in relative and
global \acfp{IRF}, and perform a qualitative analysis of the resulting \ac{STM}
maps (\Cref{sec:relative-irf-experiment,sec:global-irf-experiment}
respectively).

\subsection{Parameters of the Prior Distribution}
\label{sec:experiment-priors}

In \Cref{sec:surfel-inference}, \Cref{eq:prior}, we gave the form of the prior
distribution over the surfel parameters, but did not specify the parameters of
the prior distribution. For surfels in the map that contain many measurements,
the choice of prior parameters has a negligible effect on the resulting map
belief. However, if there are clusters of neighbouring surfels with no
measurements, then these surfels' beliefs are largely determined by the prior
distributions.

We want to capture knowledge of the structure of the environment in the prior
distribution. The structure we expect is that neighbouring regions tend to have
similar heights, which means that we expect neighbouring heights to be
positively correlated. If a completely uninformative height prior distribution
is used, then only the heights of shared vertices will be affected when
performing inference in unobserved regions. In order to achieve this, we use
a height prior distribution with covariance,
\begin{equation}
  \mmat{\Sigma}_p =
  \sigma^2
  \begin{bmatrix}
    1    & \rho & \rho\\
    \rho & 1    & \rho\\
    \rho & \rho & 1\\
  \end{bmatrix}
  ,
\end{equation}
where $\rho$ is the Pearson correlation coefficient and $\sigma^2$ the variance.
Without any other information, we expect the heights in a submap to be
distributed around the submap plane. We therefore choose the mean of the height
prior distribution to represent this initial belief; that is, we set the mean to
the zero vector. We choose $\sigma^2$ to be uninformative and focus on $\rho$,
which should be chosen in the range $\rho \in [0, 1)$ to ensure a positively
correlated prior distribution. It should be noted that, as the planar deviations
in neighbouring surfels are modelled as being statically independent, we model
the prior of the planar deviation in a surfel as an uninformative inverse-gamma
distribution.

\begin{figure*}[h!]
  \centering
  \includegraphics[]{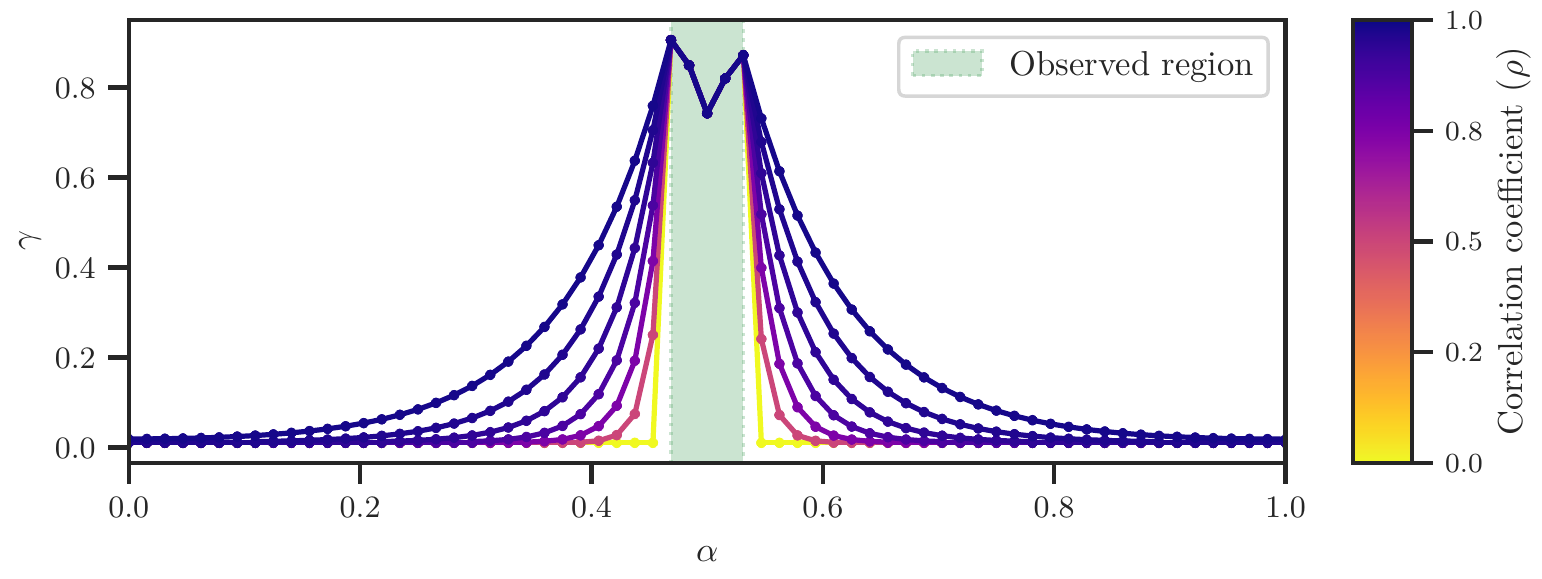}
  \caption{The effect of using a correlated prior distribution when a section of
    the environment is observed. This is shown for a series of correlation
    coefficients, spaced using a geometric series: $\{0, 2^{-1}, 2^{-1}+2^{-2},
    \ldots\}$. Only the mean of the mean mesh belief is shown, and it is coloured
    according to the correlation coefficient (the darker, the more correlated).}
  \label{fig:h_prior}
\end{figure*}

In \Cref{fig:h_prior}, we show the result of varying the correlation coefficient
on the map belief for a 2-D simulation. From this we can see that, by increasing
the correlation coefficient, the mean heights of unobserved surfels are affected
at an increasing distance from the observed region, which is consistent with
smoothly varying environments. We empirically chose $\rho=0.5$, although this
choice may vary depending on the nature of the environment. This parameter could
conceivably be learnt from data.

\subsection{Cost of Inference}
\label{sec:converge-experiment}

From a tractability perspective, for the \ac{STM} mapping technique to be viable
for online dense mapping, the cost of incorporating one measurement into the map
should be constant, irrespective of the size of the map or the number of
measurements in the map. This computational requirement is equivalent to the
computational complexity of updating the map being linear in the number of
measurements, which we previously stated to be true for \iac{STM} map
(\Cref{sec:algorithm}). To motivate this statement, we analyse the cost of
inference by simulating two scenarios we expect a robot to encounter frequently
in practice: incrementally observing new regions of the environment, and
re-observing previously observed regions. It should be noted that, to quantify the
cost of inference, we count the number of outgoing messages calculated when
updating the map (\Cref{alg:inference}, \cref{alg:inf_vmp4,alg:inf_lbp4}). This
is an appropriate metric, as it measures the number of iterations for the
innermost loops of our inference algorithm.

For both scenarios, we represented the surface of the environment using a 2.5-D
surface generated using Perlin noise\footnote{A type of gradient noise
  originally developed by \citet{Perlin1985}, which is used to procedurally
  generate virtual environments and textures in computer graphics.}---a
different surface was generated for each scenario. A single submap was used
containing \num[round-mode=off,scientific-notation=false]{1024} surfels.
Simulated measurements were generated within the relative IRF of the submap by
randomly sampling the surface in the observed region (we discuss the sampling
strategies further shortly), and then adding zero-mean non-identically
distributed Gaussian noise to each sample. The noise covariance was determined
by randomly rotating a diagonal covariance matrix. We also ensured that the
measurement density in the observed region was approximately uniform and
constant across all time steps, at 10 measurements per surfel. Finally, after
updating the map at each time step, all the new measurements are windowed.

\begin{figure*}[h!]
  \centering
  \subfloat[]{
    \includesvg[]{results/incremental/push-broom/setup}%
    \label{fig:incr_push_setup}
  }
  \subfloat[]{
    \includegraphics[]{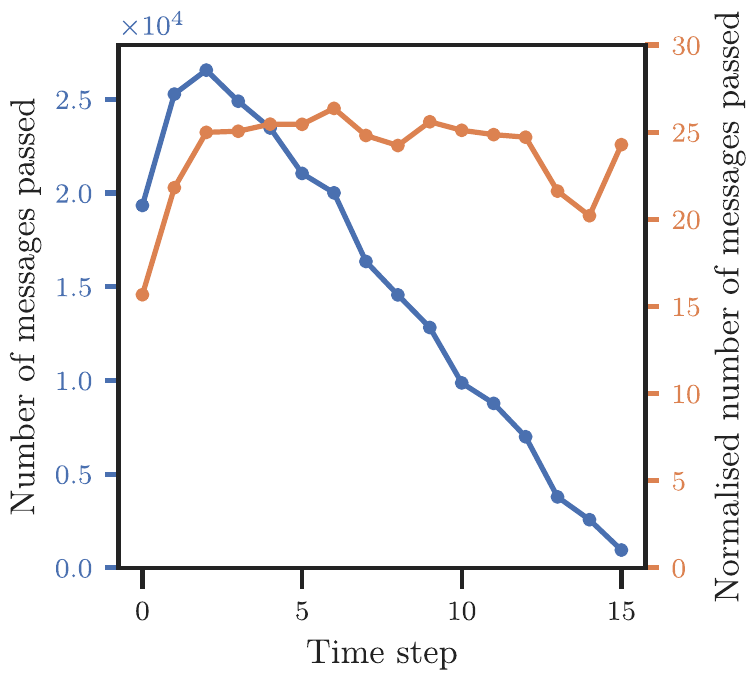}%
    \label{fig:incr_push_converge}
  }

  \subfloat[]{
    \includegraphics[]{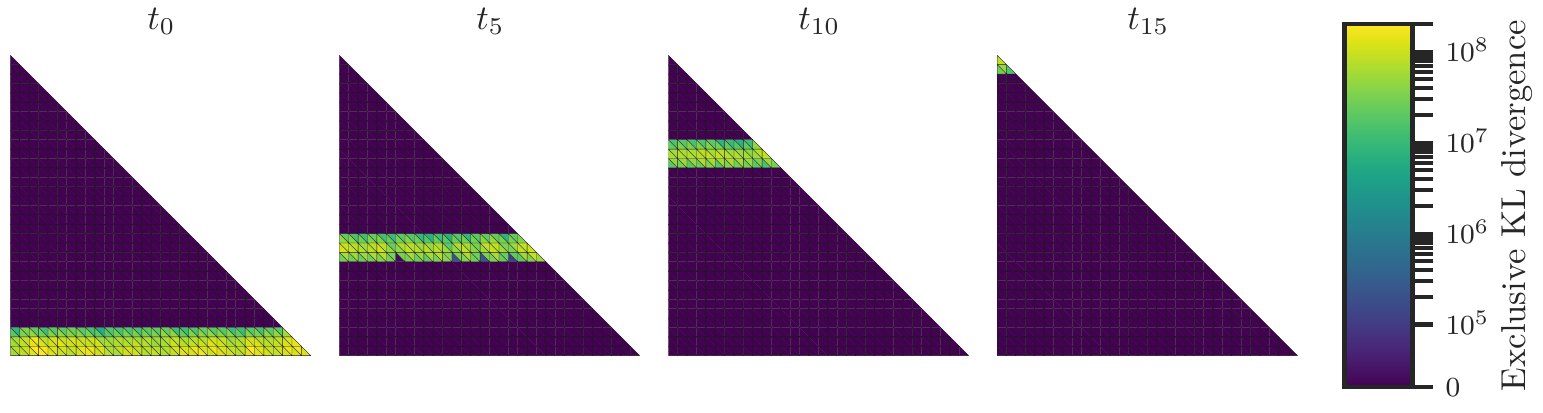}%
    \label{fig:incr_push_distances}
  }
  \caption{An experiment showing the cost of inference for the scenario in which
    new regions of the environment are incrementally observed. (a) The
    experimental setup showing the sampling regions at selected time steps
    (green, shaded). Note that the size of the sampling region at each time step
    decreases linearly. (b) The cost of inference is quantified by the number of
    messages passed (blue, left axis). This message count is also normalised by
    the number of new measurements in each update (orange, right axis). (c) To
    visualise the incremental changes in the surfel beliefs, we calculate the
    exclusive \acf{KL} divergence between the surfel beliefs before and after
    the measurements at a specific time step were incorporated. Note, the
    colour mapping is scaled using a symmetrical logarithm, which is linear on
    $[0,10^5]$ and logarithmic everywhere else.}
  \label{fig:incr_push}
\end{figure*}

In the first scenario, we simulated a newly observed region of the environment
that shifts incrementally forward--this is typically the case for a robot
driving with a 2-D LiDAR sensor mounted in a push-broom configuration. At each
time step, simulated measurements are generated from a sampling region
(\Cref{fig:incr_push_setup}), and the map is updated with the new measurements.
The cost of inference for this experiment is shown in
\Cref{fig:incr_push_converge}. We see that the total number of messages passed
at each time step has a linearly decreasing trend, which is expected due to the
linearly decreasing sampling region. When we normalise the message counts in
each update by the number of new measurements in the update, then this
normalised message count is relatively constant. Note that with each incremental
update, only local regions of the map were affected significantly
(\Cref{fig:incr_push_distances}).

\begin{figure*}[h!]
  \centering
  \includegraphics[width=0.98\textwidth]{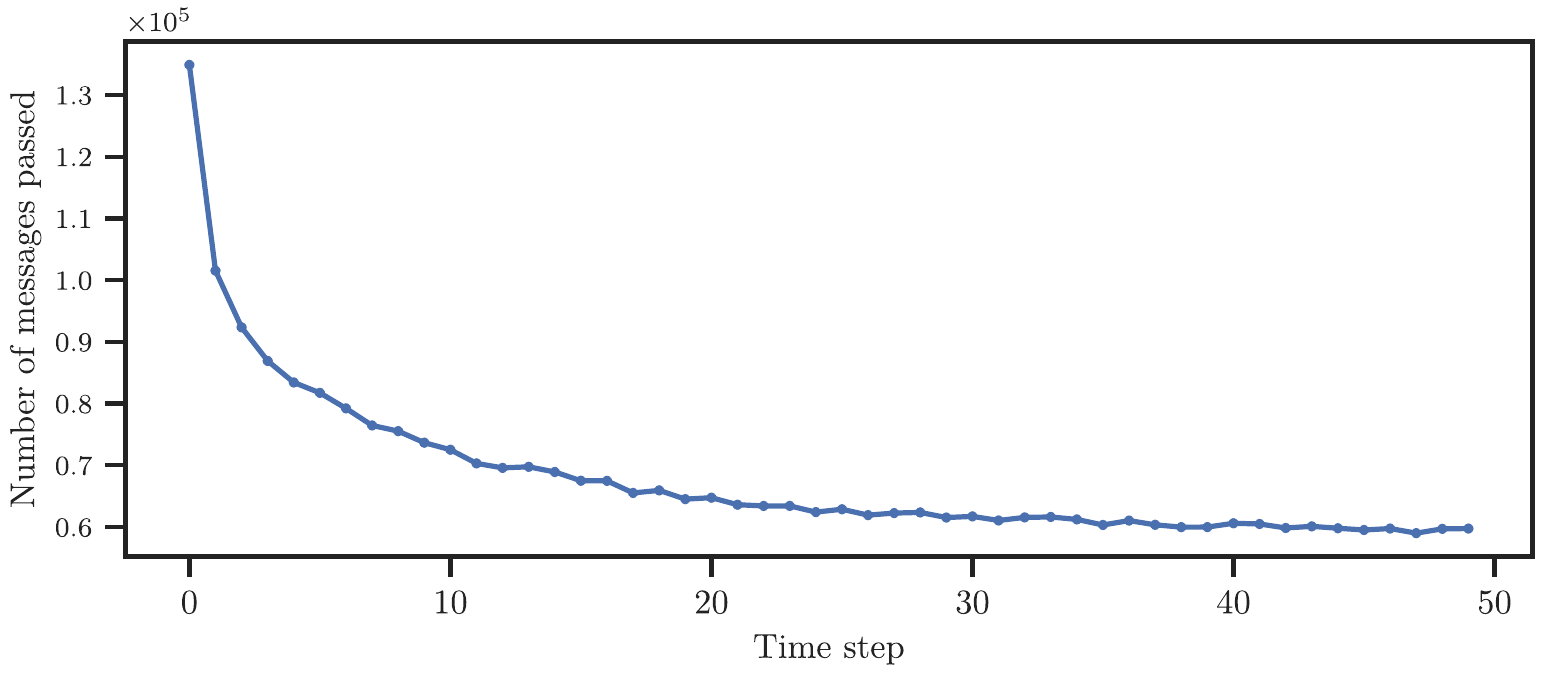}
  \caption{An experiment showing the cost of inference for the scenario in which
    the same region of the environment is observed repeatedly.}
  \label{fig:incr_reobserve}
\end{figure*}

In the second scenario, we simulated a region of the environment that is
observed repeatedly. This emulates a robot revisiting a previously explored
region, or a static robot repeatedly observing the same region. At each
time step, simulated measurements are generated by sampling from the entire
mapping region. The number of messages passed follows an exponentially
decreasing trend (\Cref{fig:incr_reobserve}); that is, subsequent observations
of a previously observed region become computationally cheaper. This is because
fewer iterations are required.

From these two scenarios, we see that updating \iac{STM} map does indeed have a
linear computational cost in the number of measurements, which makes \ac{STM}
mapping a viable online dense mapping technique for the scenarios we expect a
robot to encounter in a practical application. We have, however, only analysed
the relative computational cost of performing inference on \iac{STM} map. When
considering the absolute computational cost, one of most important aspects is
the message and factor dimensionality. In our algorithm, a message passed
between neighbouring surfels has at most two dimensions, a message passed within
a surfel has three dimensions, and the largest factor has six dimensions. These
dimensions are all fixed and relatively low, therefore dimensionality is not a
problem for our algorithm. Additionally, although our unoptimised Python
implementation takes \SI{\pm 3}{\milli\second} per measurement to update
\iac{STM} map\footnote{This is based on a later experiment
  (\Cref{sec:global-irf-experiment}) using 19 million measurements.}, we expect
significantly faster results with an efficient implementation.


\subsection{Comparisons of Model Accuracy}
\label{sec:model-compare-experiment}

\Iac{STM} map is an explicit surface representation that also captures model
uncertainty. Therefore, to evaluate the modelling accuracy of \iac{STM} map, we
compare it against a similar class of mapping techniques, namely ones that also
use an explicit surface representation and capture model uncertainty. From our
summary of the key attributes of the existing mapping techniques
(\Cref{tab:mapcompare}), the suitable candidate mapping techniques are \acf{NDT}
mapping, \acf{GP} mapping, and elevation mapping. NDT mapping, however, does not
model any measurement uncertainty, and GP mapping is intractable as an online
dense mapping technique due to the cubic computational complexity. We therefore
only compare \ac{STM} mapping against standard elevation mapping---that is, each
height is updated using a one-dimensional Kalman filter \citep{Triebel2006,
  Fankhauser2018}.

We perform this comparison using simulated measurement data by quantifying each
model's accuracy against the ground truth using the \ac{MSE} of each model, and
the log-likelihood ratio between both models. Using the same set of
measurements, models were built for increasingly finer grid divisions. We first
perform this comparison on a 2-D environment\footnote{The 2-D simulation
  environment was created using the height profile of a mountain extracted using
  manual photogrammetry.}, as it is more intuitive to visualise, and we then
extend this to a 3-D environment, which is generated using Perlin noise. We also
ensured that there were $\pm10$ measurements per surfel at the finest grid
division (for both the 2-D and 3-D experiments).

The results of the 2-D experiment are shown in \Cref{fig:elev_comp_2d}. The
\ac{MSE} of the \ac{STM} map is better than that of the elevation map for all
grid divisions greater than zero. When there is no subdivision (grid division
depth = 0), the mean plane of the only surfel in the \ac{STM} map is almost zero
and at the same height as the elevation map (\Cref{fig:elev_comp_2d_combined},
grid division depth = 0), because the average gradient of the surface of the
environment is close to zero. As the grid divisions increase past a certain
point, the accuracy of both models begins to degrade (for the \ac{MSE}, this
occurs for grid divisions $> 6$ for the \ac{STM} map and $> 7$ for the elevation
map). This effect is attributed to the measurement noise (relative to the size
of the grid element) becoming too large, and the number of measurements in each
grid element decreasing with each grid division. The effect is apparent when
looking at the grid division depth of 8 in \Cref{fig:elev_comp_2d_combined}.
From the \ac{MSE} alone, this result might not appear significant; looking at
the log-likelihood ratio shows that, for this environment, the \ac{STM} map is a
more likely model by orders of magnitude. This is as a consequence of the
elevation map estimating only the mean height of the surface, whereas the
\ac{STM} map represents the surface as a stochastic process, which is a better
representation of the surface of the environment.

\begin{figure*}[h!]
  \centering
  \subfloat[]{
    \includegraphics[width=0.48\textwidth]{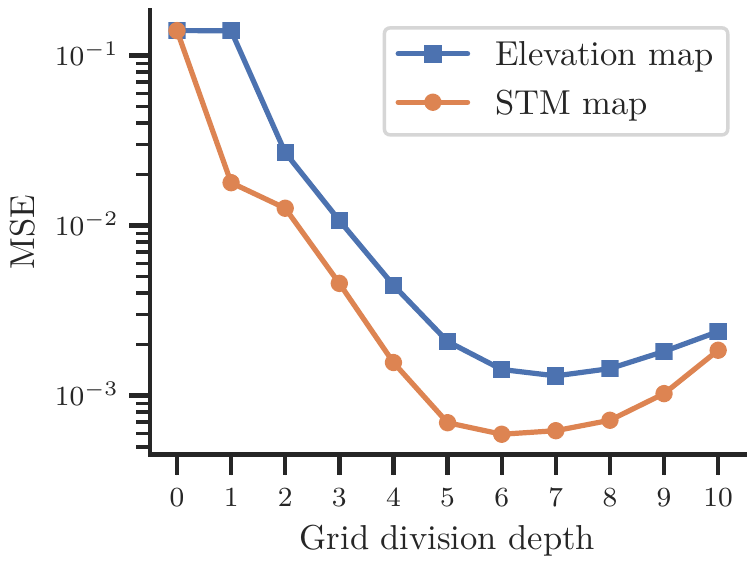}
  }
  \subfloat[]{
    \includegraphics[width=0.48\textwidth]{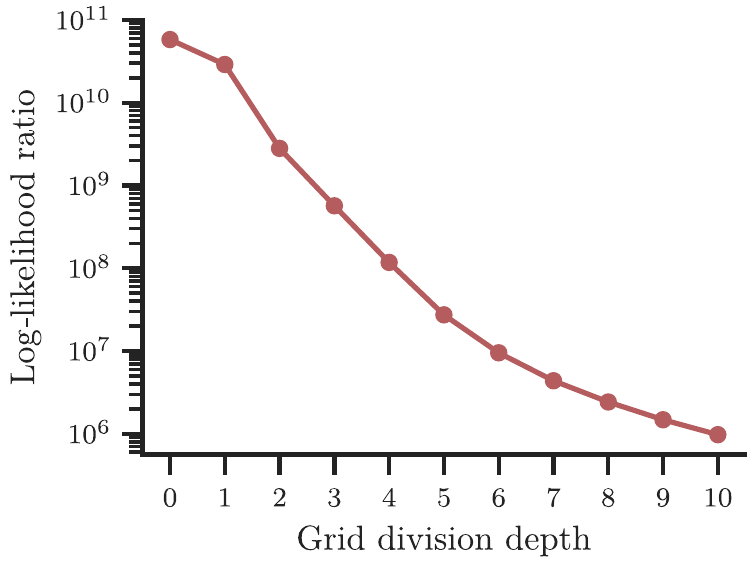}
  }
  \phantomcaption
\end{figure*}
\begin{figure*}[h!]
  \centering
  \ContinuedFloat
  \subfloat[]{
    \includesvg[width=\textwidth]{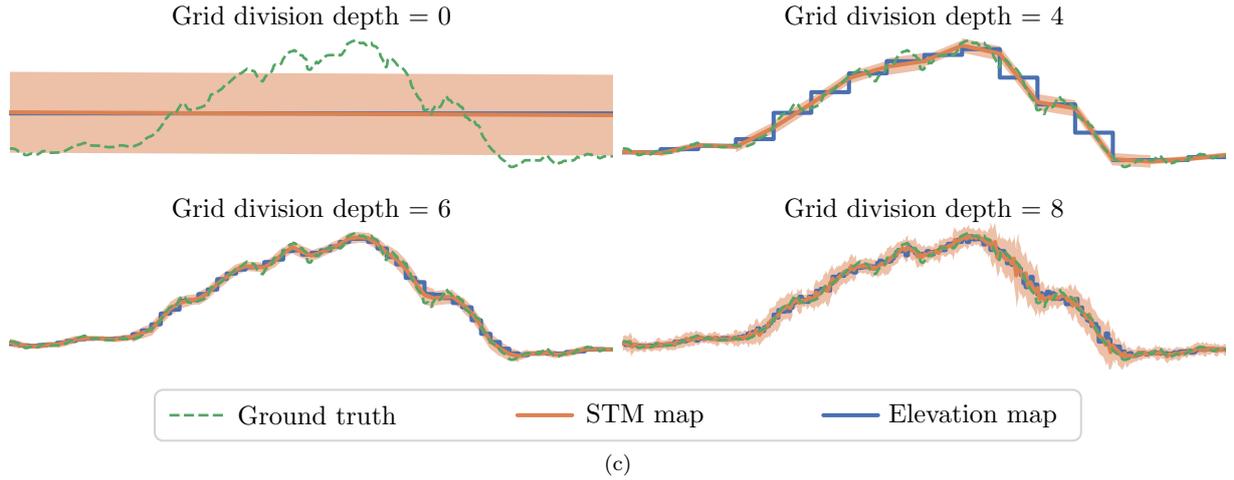}
    \label{fig:elev_comp_2d_combined}
  }
  \caption{An experiment showing a 2-D simulation comparing \iacf{STM} map with
    the standard elevation map \citep{Triebel2006, Fankhauser2018}. For
    increasing grid divisions, we calculate (a) the \acf{MSE} of each mapping
    technique compared to the ground-truth surface (lower is better), and (b)
    the log-likelihood ratio between the \ac{STM} map and the elevation map is
    evaluated along the ground-truth surface (lower is better); that is,
    $\log\text{(\ac{STM} map likelihood})-\log(\text{elevation map
      likelihood})$. Note that the vertical axis for both comparisons uses a log
    scale. (c) We also show the ground truth and resulting models for selected
    grid depths. The shaded regions indicate the one standard deviation
    confidence intervals in both models (the standard deviations of the
    elevation map are too small to be seen).}
  \label{fig:elev_comp_2d}
\end{figure*}

\begin{figure*}[h!]
  \centering
  \subfloat[]{
    \includegraphics[width=0.48\textwidth]{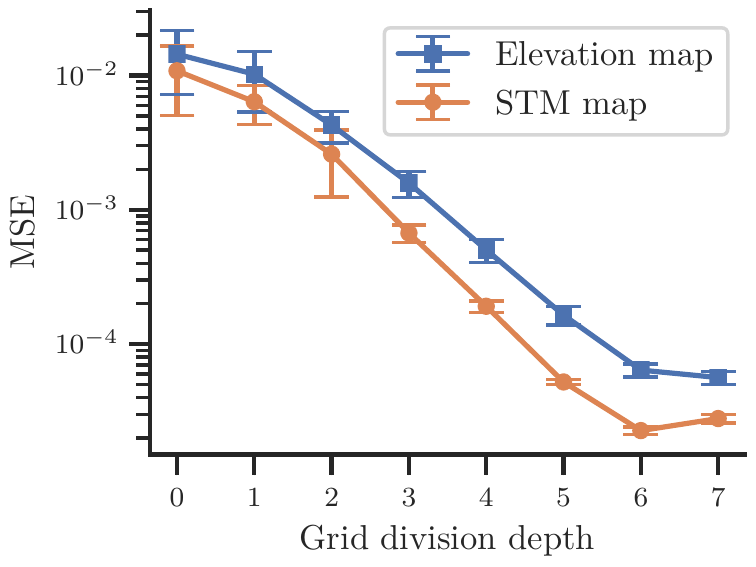}
  }
  \subfloat[]{
    \includegraphics[width=0.48\textwidth]{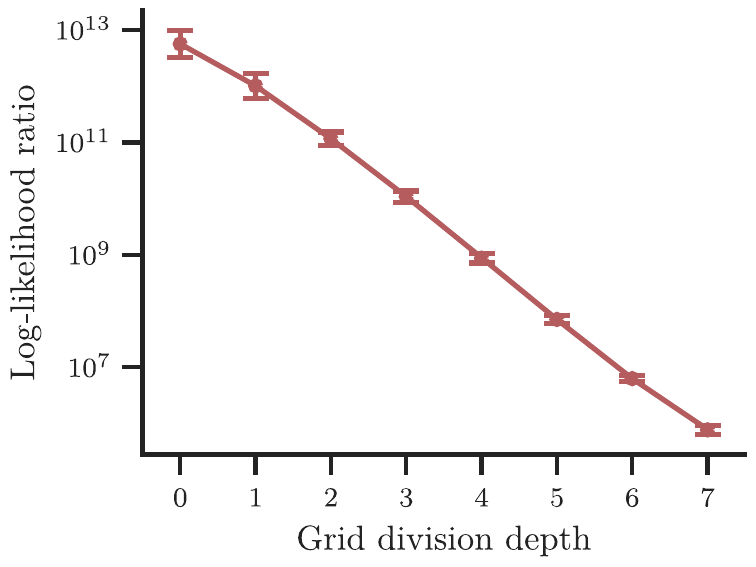}
  }
  \caption{An experiment showing a 3-D batch simulation ($n=10$) comparing
    \iacf{STM} map with the standard elevation map \citep{Triebel2006,
      Fankhauser2018} for various generated environments. The mean and first
    standard deviation are shown for (a) the \acf{MSE} of each mapping technique
    compared to the ground truth surface, and (b) the log-likelihood ratio
    between the \ac{STM} map and the elevation map is evaluated along the
    ground-truth surface; that is, $\log\text{(\ac{STM} map
      likelihood})-\log(\text{elevation map likelihood})$.}
  \label{fig:elev_comp_3d}
\end{figure*}

For the 3-D experiment, we ran a batch simulation for 10 synthetically generated
environments (\Cref{fig:elev_comp_3d}). Compared to the 2-D experiment, we see a
similar trend in the \ac{MSE}, namely that the \ac{STM} map is clearly and
consistently better than the elevation map for grid division greater than 2.
Interestingly, as the grid divisions decrease, there is also a decrease in the
batch \ac{MSE} standard deviation for both models ($> 2$ grid divisions). This
is because the lower grid divisions are too coarse to effectively represent the
environment, and therefore the model accuracy is largely dependent on the
particular generated environment. Similarly to the 2-D experiment, the
log-likelihood ratio between both models shows that \iac{STM} map is a
significantly better representation than an elevation map.

From these results, we can conclude that, in comparison to a standard elevation
map, \iac{STM} map is a more accurate and better representation of the
environment for appropriately chosen grid divisions.


\subsection{STM Mapping in a Relative IRF}
\label{sec:relative-irf-experiment}

\begin{figure*}[p!]
  \centering
  \subfloat[]{
    \includegraphics[]{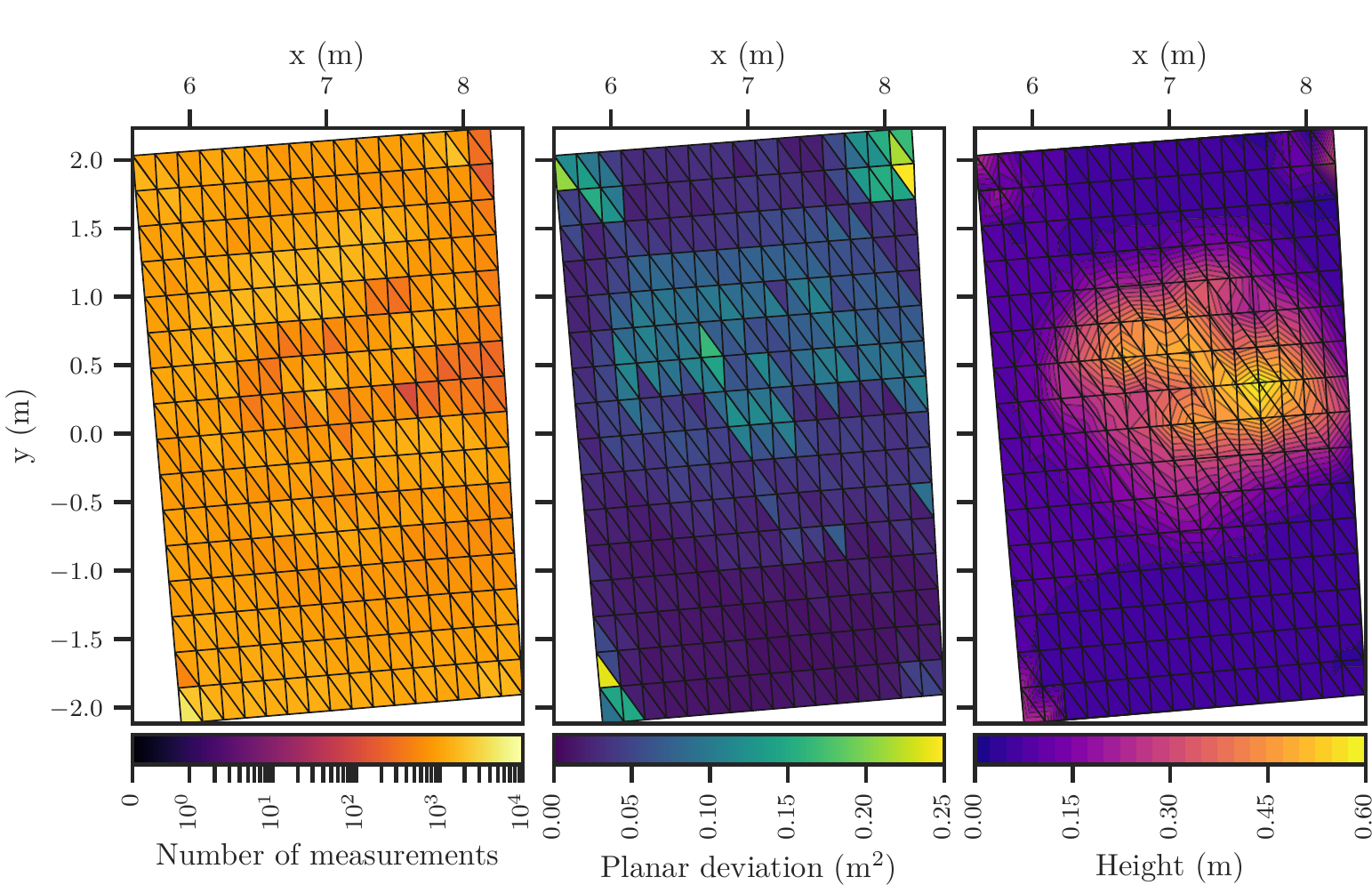}
    \label{fig:stationary_fused}
  }

  \subfloat[]{
    \includegraphics[width=0.2\textwidth, height=3cm, keepaspectratio]{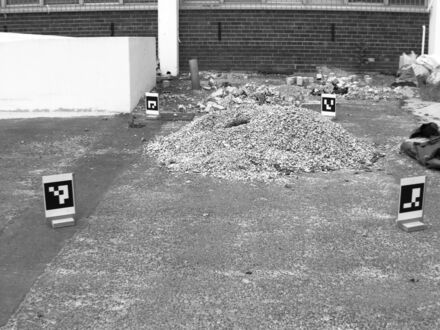}
    \label{fig:stationary_f0}
  }
  \subfloat[]{
    \includegraphics[width=0.22\textwidth, height=3cm, keepaspectratio]{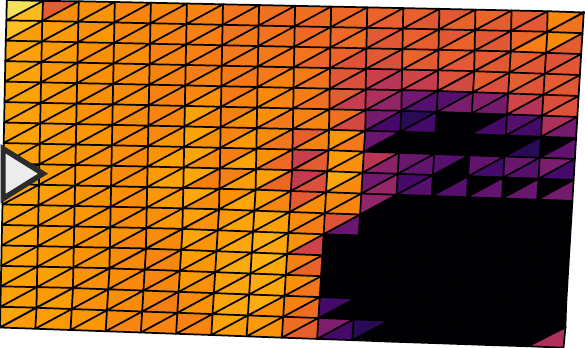}
    \label{fig:stationary_f1}
  }
  \subfloat[]{
    \includegraphics[width=0.22\textwidth, height=3cm, keepaspectratio]{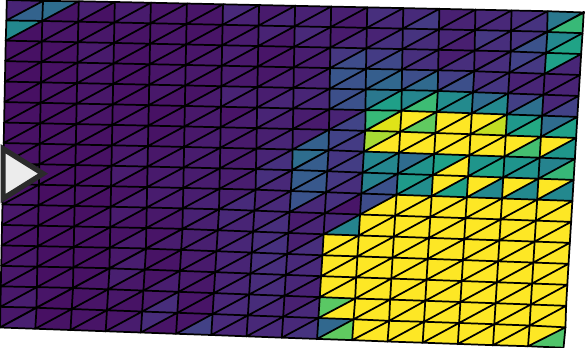}
    \label{fig:stationary_f2}
  }
  \subfloat[]{
    \includegraphics[width=0.22\textwidth, height=3cm, keepaspectratio]{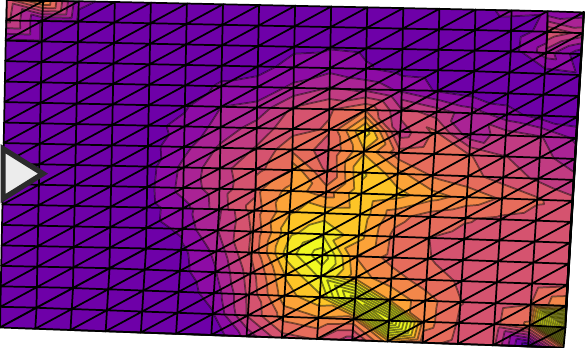}
    \label{fig:stationary_f3}
  }

  \subfloat[]{
    \includegraphics[width=0.2\textwidth, height=3cm, keepaspectratio]{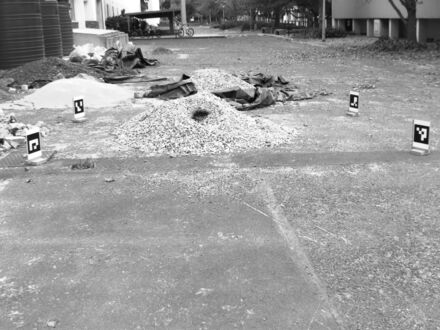}
    \label{fig:stationary_s0}
  }
  \subfloat[]{
    \includegraphics[width=0.22\textwidth, height=3cm, keepaspectratio]{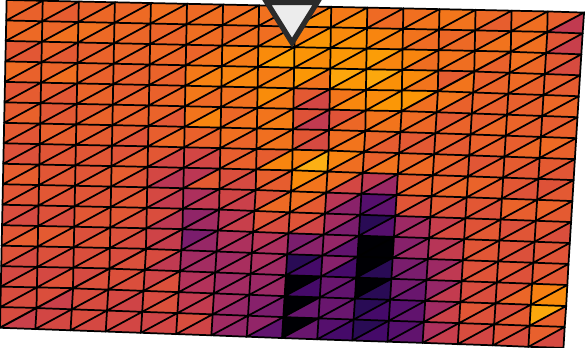}
    \label{fig:stationary_s1}
  }
  \subfloat[]{
    \includegraphics[width=0.22\textwidth, height=3cm, keepaspectratio]{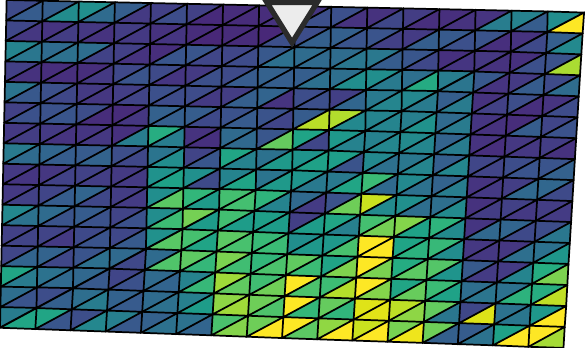}
    \label{fig:stationary_s2}
  }
  \subfloat[]{
    \includegraphics[width=0.22\textwidth, height=3cm, keepaspectratio]{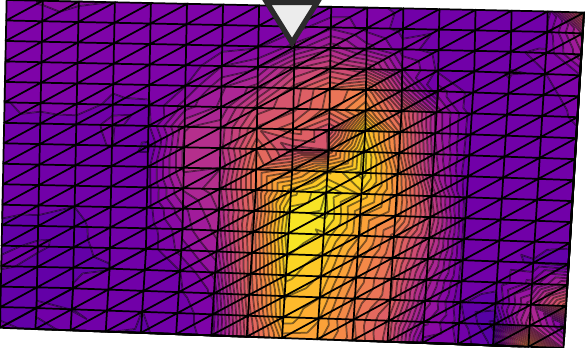}
    \label{fig:stationary_s3}
  }

  \subfloat[]{
    \includegraphics[width=0.2\textwidth, height=3cm, keepaspectratio]{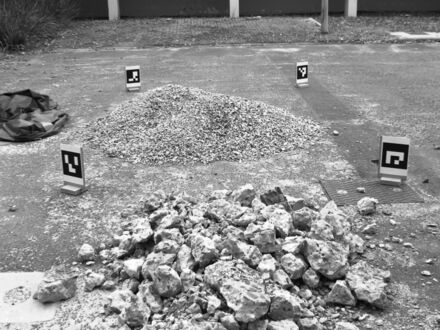}
    \label{fig:stationary_b0}
  }
  \subfloat[]{
    \includegraphics[width=0.22\textwidth, height=3cm, keepaspectratio]{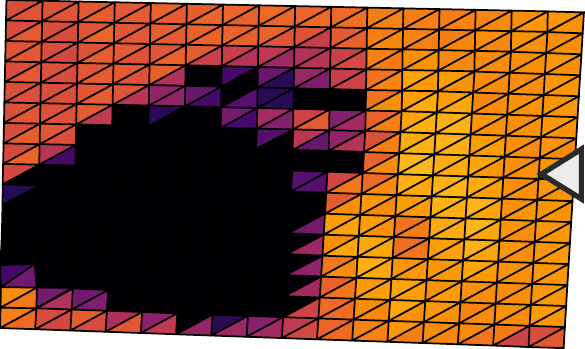}
    \label{fig:stationary_b1}
  }
  \subfloat[]{
    \includegraphics[width=0.22\textwidth, height=3cm, keepaspectratio]{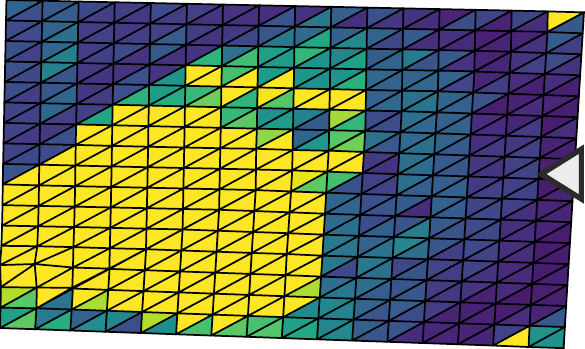}
    \label{fig:stationary_b2}
  }
  \subfloat[]{
    \includegraphics[width=0.22\textwidth, height=3cm, keepaspectratio]{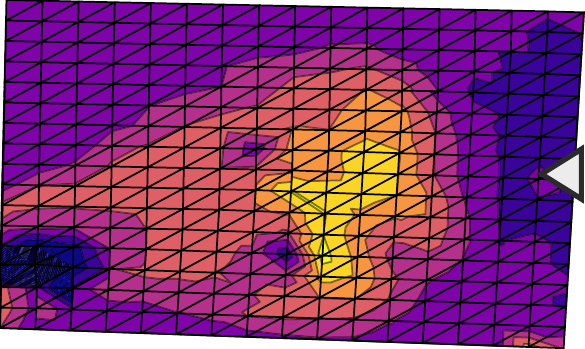}
    \label{fig:stationary_b3}
  }
  \caption{An experiment showing \iac{STM} map created using only three pairs of
    stereo images of an outdoor environment. This was made possible by using
    relative IRFs to construct two submaps, with artificial ArUco markers
    \citep{Garrido2014} as landmarks. Each submap contains 256 surfels. (a) The
    resulting map is shown for the mean of the mean mesh belief and the planar
    deviation belief---calculated according to \Cref{eq:expected_deviation}. We
    also show the number of measurements incorporated into each surfel on a
    symmetrical logarithmic scale, with a linear region on $[0, 1]$. (b-m) The
    \ac{STM} maps built for each stereo image pair---namely (b-e) a front view,
    (f-i) a side view, and (j-m) a rear view---are on the same scale as (a). The
    viewpoints are indicated using arrows. Notably, the occlusions are evident
    due to a lack of measurements (c, g, k). The orientation of (a) is visually
    consistent with the front view (b)---that is, a \SI{90}{\degree} rotation of
    (c-e).}
  \label{fig:stationary}
\end{figure*}

Although robustly and persistently identifiable landmarks are not currently
possible for general environments, there are still viable practical applications
using landmark-based relative \acp{IRF}. In order to demonstrate this, we place
artificial landmarks in the environment in the form of ArUco markers
\citep{Garrido2014}---a popular open-source library enabling the creation and
detection of binary square fiducial markers. Using these landmarks to define the
submapping region, we are able to create \iac{STM} map using only three pairs of
stereo images (\Cref{fig:stationary}). The image pairs were synchronously
acquired using a stereo setup consisting of two
\SI[round-mode=off]{1.3}{\mega\px} FLIR Flea3 GigE cameras at baseline of
\SI{18}{\centi\meter}. We calculated disparity maps (depth images) from image
pairs using the \ac{LIBELAS} of \citet{Geiger2011}. The disparity maps were then
transformed to noisy 3-D point measurements using the unscented transform
\citep{Julier2002}. Following this process, \num{525331} point measurements were
extracted from the three image pairs and incorporated into \iac{STM} map.

When considering the resulting \ac{STM} maps, we see that the height maps are
visually consistent with the environment. In the rougher regions of the
environment (the gravel heap and landmarks in the corners) we see elevated
planar deviations, whereas in the smooth regions (the asphalt path) the planar
deviations are low. This illustrates the usefulness of the planar deviation in
\iac{STM} map quantifying the surface roughness, which could be used as a
measure of terrain drivability. Additionally, despite there being occlusions in
the viewpoints (\Cref{fig:stationary_f1,fig:stationary_s1,fig:stationary_b1}),
the resulting height maps in these occluded regions are affected by the
inference process, and are consistent with a smoothly varying environment. This
is due to the correlated prior over the surfel vertex heights
(\Cref{sec:experiment-priors}). From this experiment, the resulting map beliefs
show that \iac{STM} map that is visually consistent with the environment can be
constructed from only a few viewpoints. This is also performed using stereo
camera measurements, which have large range uncertainty.

This test simultaneously demonstrates the success of performing dense mapping in
a relative IRF for two practical scenarios: the worst-case SLAM scenario, where
there is no information about the robot motion (also commonly known as the
kidnapped robot problem \citep[chap.~7.1]{Thrun2005}); as well as the scenario
where multiple robots are mapping the same region of the environment. From the
perspective of the first scenario, this experiment shows how using a relative
IRF fits into the SLAM framework, and how it can perform dense mapping even
without any pose information (except that which can be calculated from the
landmarks). From the perspective of the second scenario, we illustrate the
simplicity of using a relative IRF to perform multi-robot mapping. The maps
created from each stereo image pair represent a robot's perspective
(\Crefrange{fig:stationary_f0}{fig:stationary_b3}). As each submap is created
relative to the landmarks in the environment, the submaps are already aligned
and can be fused together to form a single belief over the environment
(\Cref{fig:stationary_fused}).

\subsection{STM Mapping in a Global IRF}
\label{sec:global-irf-experiment}

In order to demonstrate the ability to use \ac{STM} mapping in a global
\ac{IRF}, we use the Canadian planetary emulation terrain 3-D mapping dataset by
\citet{Utoronto2013}. Specifically, we use the \verb|box_met|
dataset\footnote{\url{http://asrl.utias.utoronto.ca/datasets/3dmap/box_met.html}}
taken in the Mars emulation terrain---a \SI[product-units =
brackets]{60x120}{\metre} outdoor area. The \verb|box_met| dataset consists of
\num{19.044655e6} LiDAR measurements, taken from 112 poses, and aligned using a
\ac{DGPS}.

In order to construct \iac{STM} map, a global \ac{IRF} was created for the
rectangular mapping region. The region was subdivided into two submaps, each
containing \num{65536} surfels. From the resulting \ac{STM} map
(\Cref{fig:utoronto}), we see that the mean mesh belief (\Cref{fig:utoronto_b})
is visually consistent with satellite imagery. From the zoomed-in region
(\Cref{fig:utoronto_d}), we see that all the major features visible in the
satellite imagery of the environment, even small rocks, are also visible in the
map belief. We again see that the planar deviation beliefs
(\Cref{fig:utoronto_c}) are visually consistent with the terrain roughness. The
steeper regions in the environment, which have a higher planar deviation, also
form an outline of the obstacles in the environment. This experiment
demonstrates that \iac{STM} map can be created in a global \ac{IRF} framework.
Although this lacks the benefits of a relative \ac{IRF}, the added complexity of
a relative \ac{IRF} may not be necessary in situations where the robot
localisation is guaranteed to perform accurately; however, this is generally not
the case. Additionally, this experiment shows that \ac{STM} maps can handle
datasets with a large number of measurements.

\begin{figure*}[p!]
  \centering
  \subfloat[]{
    \includesvg[width=0.85\textwidth]{results/utoronto/google_overlay}%
    \label{fig:utoronto_a}
  }

  \subfloat[]{
    \includesvg[width=0.85\textwidth]{results/utoronto/h}%
    \label{fig:utoronto_b}
  }

  \subfloat[]{
    \includesvg[width=0.85\textwidth]{results/utoronto/v}%
    \label{fig:utoronto_c}
  }
  \phantomcaption
\end{figure*}
\begin{figure*}[p!]
  \centering
  \ContinuedFloat
  \subfloat[]{
    \includesvg[width=0.85\textwidth]{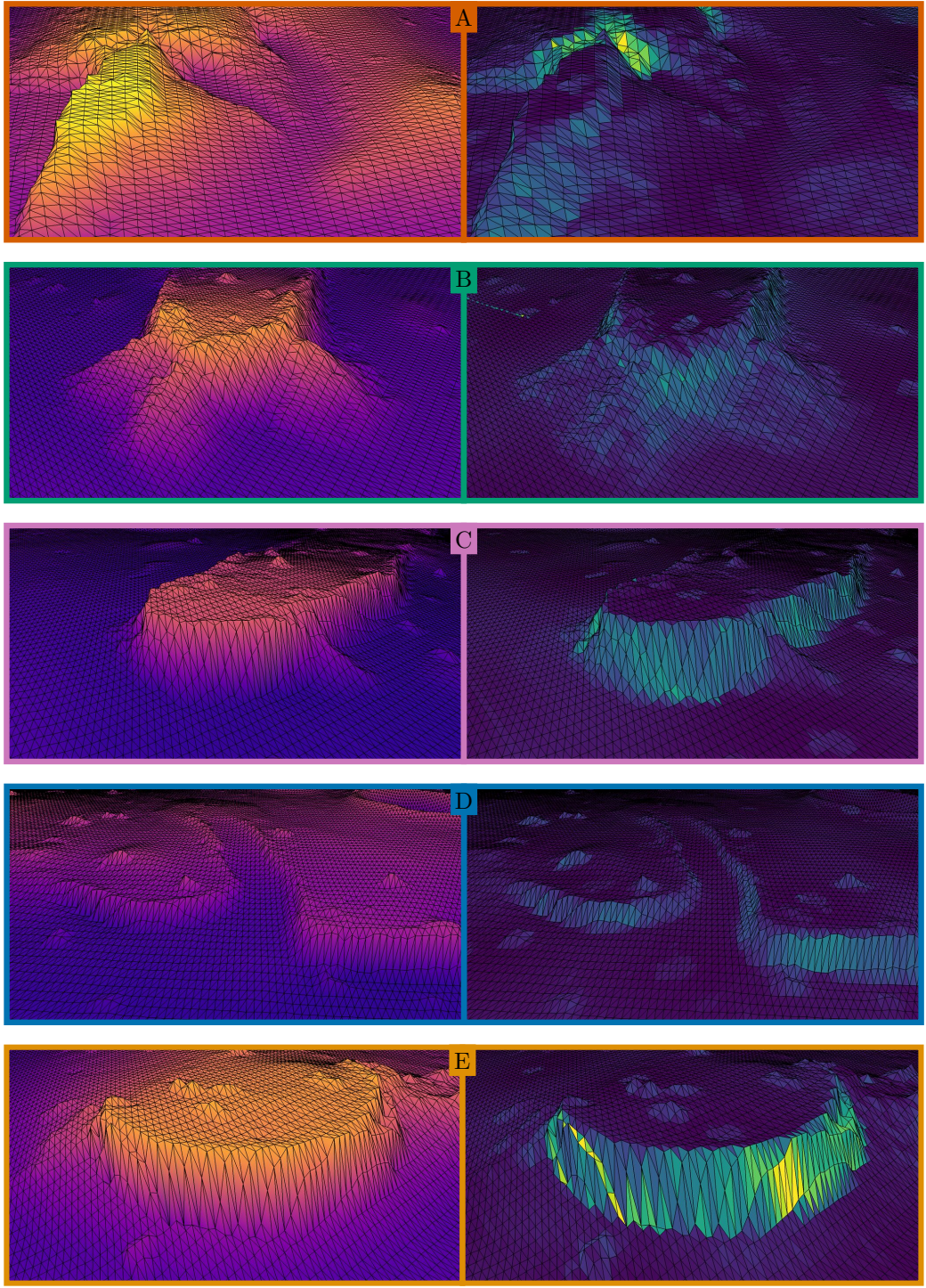}%
    \label{fig:utoronto_d}
  }
  \caption{An experiment showing \iac{STM} map created in a global \ac{IRF}
    using the Canadian planetary emulation terrain 3-D mapping dataset
    \citep{Utoronto2013}. (a) A satellite view of the mapping region (Google
    Earth). Note that five zoom regions are demarcated using coloured rectangles
    and labelled A to E. The resulting map is shown for (b) the mean of the mean
    mesh belief, and (c) the planar deviation beliefs---calculated according to
    \Cref{eq:expected_deviation}. (d) The map belief for the zoomed regions;
    they share the same colour maps as in (b, c).}
  \label{fig:utoronto}
\end{figure*}


\section{Conclusions and Future Work}
\label{sec:conclusion}

To address the drawbacks in existing dense mapping techniques, we presented the
\acf{STM} mapping technique, which: represents the structure in the environment
using a collection of stochastic processes, forming a continuous representation;
is efficient to update, with approximately linear computation time in the number
of measurements; is able to handle uncertainty in both measurements and the
robot pose; and allows online operation by incrementally updating the map.
Although existing techniques have some of these attributes, \ac{STM} mapping is
the only technique that is able to achieve them all. We have shown that
\iac{STM} map is a more accurate and a better representation of the environment
in comparison to the standard elevation map \citep{Triebel2006, Fankhauser2018},
when evaluated in terms of \acl{MSE} and model likelihood. Qualitative results
on practical datasets showed that \ac{STM} maps provide an accurate and
descriptive model of the surface of the environment. Additionally, we showed
that \ac{STM} maps are agnostic to the sensor modality, and can handle both
accurate (LiDAR) and inaccurate (stereo camera) sensors. Therefore, if a sensor
can produce point measurements, it can be used with \iac{STM} map.

In addition to presenting the \ac{STM} mapping technique, we have also
demonstrated dense mapping using relative \acp{IRF} in the \ac{HYMM} submapping
framework \citep{Guivant2004, Nieto2006}, and extended the framework to 3-D.
Using this submapping approach largely decouples the process of localisation
from that of dense mapping. This negates some of the issues in performing
localisation using \ac{SLAM}---specifically, it seamlessly allows the
incorporation of measurements from multiple robots into a single, consistent
representation, mapping between robot poses when the absolute position of the
robot is unknown, or mapping when loop closure is performed. The HYMM submapping
framework also complements the \ac{STM} mapping technique, because it uses
triangular submapping regions that seamlessly incorporate the triangular surfels
in \iac{STM} map.

\subsection{Future Work}
\label{sec:future-work}

The environment often consists of large, homogeneous regions, and naively
dividing a dense map into fixed-sized elements is an inefficient use of storage.
Therefore, in order for \ac{STM} maps to better model the environment, future
work will look at adaptively subdividing the grid. Surfels with very low planar
deviations could be grouped into a coarser resolution and, conversely, surfels
with high planar deviations could be subdivided into a finer resolution. This,
however, would add complexity to the inference process.

As most environments with practical significance are not static, it would be
desirable to relax the static environment assumption. Future work will look at
incorporating temporal information into the map to facilitate this.

Although we propose the terrain roughness as a possible metric for drivability
analysis, we do not demonstrate this. As a motivation for \ac{STM} maps to be
used in the context of autonomous navigation, future work will look at
specialised planning algorithms to exploit the rich environment representation
of \iac{STM} map.

We currently treat each submap independently. However, as each submap lies on a
plane defined by the landmarks at its vertices, the partitioning of 3-D space
between neighbouring submaps will contain overlapping- or dead-zones. Future
work could look at a handling the boundaries between submaps accordingly.

\Iac{STM} map is a 2.5-D representation of the surface of the environment;
however, to handle 3-D general environments, future work will look at extending
the representation accordingly. As each \ac{STM} submap is associated to the
plane defined by the landmarks vertices, a solution could be to use a more
sophisticated strategy of submapping, and then handling the boundaries as
mentioned previously. Alternatively, a 3-D triangular meshing strategy could be
investigated; however, this will most likely be intractable for online dense
mapping if done in a fully probabilistic manner---as with \ac{STM} mapping.

Our algorithm performs inference on \iac{STM} map using a combination of both
\ac{VMP} and \ac{LBP}, and although \ac{VMP} is guaranteed to converge, \ac{LBP}
on the other hand, does not have any convergence guarantees. We have provided
empirical evidence supporting the convergence of our algorithm; however, future
work will investigate a principled and in-depth analysis on the convergence of
our algorithm.



\section*{Acknowledgements}
The authors would like to thank Johan du Preez for the valuable discussions, and
allowing the use of the in-house probabilistic graphical modelling toolbox.

The financial assistance of the National Research Foundation (NRF) towards this
research is hereby acknowledged. Opinions expressed and conclusions arrived at,
are those of the authors and are not necessarily to be attributed to the NRF.
The financial assistance of Armscor towards this research is hereby
acknowledged.


\bibliographystyle{./elsarticle-num-names}
\bibliography{main}

\end{document}